# An Explainable Collaborative Dialogue System using a Theory of Mind

Philip R. Cohen, Lucian Galescu, Maayan Shvo

Openstream.ai

{phil.cohen, lucian, maayan}@openstream.com

## Abstract

Eva is a neuro-symbolic domain-independent multimodal collaborative dialogue system that takes seriously that the purpose of task-oriented dialogue is to assist the user. To do this, the system collaborates by inferring their intentions and plans, detects obstacles to success, finds plans to overcome them or to achieve higher-level goals, and plans its actions, including speech acts, to help users accomplish those goals. In doing so, the system maintains and reasons with its own declaratively-specified beliefs, goals and intentions, and explicitly reasons about those of its user. Because Eva can track different users' mental states, it can engage multiple agents in multi-party dialogues. Reasoning is accomplished with a modal Horn-clause meta-interpreter that enables computable inference within the subset of logic implemented. The system employs both hierarchical and backward-chaining planning, operating over a rich modal logic-based knowledge and action representation. The planning and reasoning subsystems obey the principles of persistent goals and intentions including: 1) The formation and decomposition of intentions to perform complex actions, 2) the conditions under which persistent goals and intentions can be given up, and 3) persistent goal and intention revision using the relativizing formulas that are created during the planning process. The system treats its speech acts just like its other actions. This general approach enables Eva to plan a variety of speech acts, including requests, informs, questions, confirmations, offers, acceptances, and emotive expressions. Because the dialogue engine is a planner, as the dialogue proceeds, the system can flexibly generate, execute, and potentially repair its plans using physical, digital, and speech actions. Importantly, Eva can explain its utterances because it has created a plan that caused it to utter them.[1]

## 1 Introduction

In this paper we describe Eva, a fully-functional neuro-symbolic domain-independent collaborative dialogue system that takes seriously the tenet that the purpose of task-oriented dialogue is to *assist* the user. Eva attempts to collaborate with its users by inferring and debugging their plans, then planning to overcome obstacles to achieving their higher-level goals. In order to do so, Eva represents and reasons with beliefs, goals and intentions ("BDI")[2] of the user and the system itself. Because the dialogue engine is a planner, as the dialogue proceeds, the system is able to go beyond scripted, slot-filling, or finite state dialogue behavior to flexibly generate, execute, and potentially repair its plans using both non-communicative actions and speech-acts. As part of its reasoning, Eva performs plan/goal recognition on the user's mental state. Importantly, the system itself decides what to say, not the developer, by obeying the well-studied principles of persistent goals and intentions (see Cohen and Levesque [1]). Importantly, thanks to the BDI underlying machinery, Eva is able to explain its actions and its plans, thus achieving more trustworthy interactions.

A useful and meaningful dialog system in a rich task-oriented natural language conversational setting must be collaborative. Indeed, collaboration is so essential to society that we teach our children to be collaborative at a very early age [2]. True collaboration is more than just being "helpful", in that one could help someone else by setting up the "environment" such that the other agent succeeds. For example, we might be helpful with children in such a way that they do not know what we have done to help them. However, most conversational systems, even those dubbed as "assistants," do not know how to be helpful, much less to collaborate. At the dialogue level, they are generally incapable of inferring and responding to the intention that motivated the utterance. We and others have argued that deep collaboration involves agents' (mutual) beliefs and joint intentions to ensure that the joint goals are achieved [3–6]. Whereas physical actions are planned to alter the physical world, communicative acts are planned to alter the (joint) mental and social states of the interlocutors. Dialogue is a special case of collaboration that has properties of its own. A collaborative dialogue system is able to combine information from a representation of domain and communicative actions, a representation of the world and of its interlocutors' mental states, and a set of planning and plan recognition algorithms, in order to achieve its communicative goals. Among the actions that are planned are speech acts, some of whose definitions we have given in various papers of ours (e.g., [7–9]). The system thus plans its speech acts (e.g., to ask for the user's age) just like it plans its other actions (e.g., to make an appointment for the user). The approach dates back to work done at Bolt Beranek and Newman [10–12], at the University of Toronto [9, 13, 14], and at the University of Rochester (e.g., [15–19]). Such systems attempt to infer their conversants' plan that resulted in the communication, and then to ensure that the plans succeed. Recent works in this vein include [5, 17, 20, 21].

---

[1] ACM Class: I.2.7; I.2.8; I.2.4; I.2.3; I.2.11. Additional Keywords: Dialogue, planning, reasoning, multiagent systems, modal logic, intention, plan recognition, theory of mind

[2] We will use the expression "BDI" even though the system deals with (persistent) goals rather than desires.

We claim this expectation for dialogue and task collaboration derives from implicit joint commitments or shared plans [1, 5, 6, 22] among the conversants, here the task-oriented dialogue system and its user, towards the achievement of the user's goals. Such a joint commitment implies that the parties will help each other to achieve the jointly committed-to goal, and will inform one another if that goal becomes impossible.[3] Whereas it is possible to build dialogue systems that reason *with* the formalization of joint intention/commitment [23], the present system attempts to behave according to JI Theory principles (cf. [18, 19, 24, 25]).[4] A central feature of this approach is that the system will attempt to infer as much of the user's plan as it can, will try to identify obstacles to its success, and plan to overcome those obstacles in order to help the user achieve his/her higher-level goals. Thus, plan recognition and planning are essential to Eva's architecture and processing.

Though the collaborative plan-based approach to dialogue is an attractive theory that has received many years of research (e.g., [1, 8–10, 13, 15, 16, 30–45]), few full-scale implementations have taken place that incorporate all the components needed to create a system that can engage in useful collaborative dialogues.[5] We have built a dialogue system whose internal state is declaratively specified as logical expressions, which provides a basis for a reasoning system, as well as a formal semantics for the system's internal states and its inferences. As a result, we *and the system itself* are able to explain its inner workings. In other words, the system's state is not merely latent [46], but transparent.

The planning-based approach detailed here is quite different from approaches in present research or commercial dialogue systems, and even plan-based approaches of the past. Essentially, rather than just populate a system "belief state" made up of so-called "intents" with "slots" and "values" as present conversational systems do [47], the Eva system maintains a far more expressive representation of beliefs, goals, and intentions that drive the collaborative planning and plan recognition system. Furthermore, the same reasoning mechanism supports multi-agent (person or bot) dialogues. In doing so, the system uses a rich knowledge representation that describes its user and its domain of discourse.[6]

Whereas much of the basic theory has been developed for some time, among the novelties here is how the pieces can be combined in an *operational* dialogue system that provides collaborative dialogue capabilities. For example, current generation "intent+slot" systems have great difficulty with the user's answering a slot-filling question ("what time do you want the reservation?") with a statement of a constraint ("not before 11am"). Furthermore, such systems assume that it is the *user* who fills slots. But systems can propose potential slot fillers ("would you like 9 am?"), to which a user may then accept, reject, or negotiate [48]. The parties have *together* filled the slot, not just one of them.[7] We accomplish the reasoning with a Horn-clause interpreter (realized in Prolog), which enables computable inference within its subset of logic. Many researchers have thought the logic of intention to be too complex to implement and to support a dialogue system. However, the Eva system operates in real-time using spoken language and driving a complex avatar. This paper shows what components are needed in order to create such a functioning system. With such a system in place, we can now go back and revisit some of the research of the past three decades that can help to support even more expressive dialogues.

---

[3] There is still an asymmetry between the system and user's commitments. Namely, the joint commitment between the system and user is relative to the user's goals. Thus, if the user changes his/her goal, s/he can drop the joint commitment (e.g., simply hang up). However, the user is committed to helping the system help the user. Thus, the system cannot simply hang up by changing its commitment; it needs to close the conversation properly.

[4] We argue that the central tenet of a collaborative dialogue is a joint commitment to understanding what each other means [26], i.e., what the parties' locutionary and illocutionary [27] intentions are. Such a joint commitment underlies backchannels and grounding [28]. Of course, there are instances of non-cooperative dialogue in which even that joint commitment may not apply (e.g., a cross-examination) [29].

[5] The work of Allen et al. at the University of Rochester (see references in the text, most recently Galescu et al. [19]) is a notable exception.

[6] Of course, the hope of the present machined learned approaches to dialogue is that such knowledge representations can be avoided, and the needed inferences will somehow emerge from the latent representations (of text, knowledge graphs, or some other large-scale resource). Given the complexity of the inferences that Eva performs, we believe that remains to be seen.

### 1.1 Sample Dialogue

To give the reader an idea of the kinds of dialogues that Eva can support, we present one example set in a vaccination appointment scheduling setting. Another sample dialogue in a completely different domain can be found in the video ([2,3]) shown in Section 2.2.

| ***Nr.*** | ***Utterance*** |
|---|---|
| *1.* | U: *Are there any covid vaccination centers nearby* |
| *2.* | S: *Yes, the CVS Covid vaccination center is located at 130 Main St and this location is 6 miles away from you via the following route: turn right, go straight, then turn left.* |
| *3.* | S: *Would you like to be vaccinated at a Covid vaccination center?* |
| *4.* | U: *Yes* |
| *5.* | S to CVS: *Do you have covid vaccine available?* |
| *6.* | CVS to S: *Yes* |
| *7.* | S to U: *OK, I believe that the CVS covid vaccination center has Covid vaccine.* |
| *8.* | S: *How old are you?* |
| *9.* | U: *why do you ask* |
| *10.* | S: *The reason is that I need to determine whether you are eligible for the vaccine.* |
| *11.* | S: *Sorry to repeat myself… How old are you?* |
| *12.* | U: *45 years old* |
| *13.* | S: *What is your occupation?* |
| *14.* | U: *why* |
| *15.* | S: *The reason is I need to determine if you are an essential worker* |
| *16.* | S: *Sorry to ask you again, but what is your occupation?* |
| *17.* | U: *I am a teacher* |
| *18.* | S: *You are eligible for the covid vaccine and appointments are available at CVS on June 26 at 9am.* |
| *19.* | S: *Would you like me to make an appointment at a covid vaccination center?* |
| *20.* | U: *Yes, please* |
| *21.* | S: *What time would you like the appointment?* |
| *22.* | U: *Monday the earliest time available* |
| *23.* | S: *the earliest time available is 9 am.* |
| *24.* | S: *Would you like me to make an appointment for you at 9 am?* |
| *25.* | U: *Yes, please* |
| *26.* | S: *Ok I made an appointment for you at CVS on June 26 2023 at 8am.* |
| *27.* | S: *Is there anything else I can help you with?* |

[7] See [49] for further discussion.

| | |
|---|---|
| *28.* | U: *yes, i actually cannot make it on monday at 8* |
| *29.* | S: *Would you like me to reschedule your appointment at CVS that is currently scheduled on June 26 2023 at 8?* |
| *30.* | U: *Yes* |
| *31.* | S: *What date would you like me to reschedule your appointment at CVS to?* |
| *32.* | U: *Monday at 10* |
| *33.* | S: *Ok, I've rescheduled your appointment at CVS to a new time on June 26 2023 at 10am.* |
| *34.* | S: *Is there anything else I can help you with?* |
| *35.* | U: *No, thanks.* |
| *36.* | S: *Glad to be of service. Goodbye!* |

In this sample dialogue (see more details in Appendix B), we see that the system can:

- Plan its actions to achieve its goals,
- Infer the goals and intentions that the user likely has,
- Plan to assist the user by proactively informing what it believes the user needs to know (e.g., the route),
- Handle the user's over-answering
- Handle constraints (e.g., "the earliest time available")
- Reason with domain rules (e.g., a person who is less than 50 years old is eligible for the covid vaccine if that person is an essential worker),
- Explain its actions (e.g., its reasons for asking about the user's age and occupation)
- Generate new knowledge acquisition goals.
- Reason about who knows the answers to its knowledge acquisition goals (e.g., CVS),
- Ask that third party a question,
- Develop and execute a plan to accommodate the user's changing her mind by achieving the user's revised goal. In the process, the system undoes what has already been done in the process of achieving the user's original goal (i.e., it reschedules the user's appointment from Monday at 9 to Monday at 10).

The rest of the paper describes how our planning-based dialogue system can support such behavior. Moreover, Appendix B will revisit this example in greater detail.

## 1.2 Map of the Paper

In Section 2 we describe the systems architecture, including its essential representations, its basic operating loop, its embodiment as an avatar, and its logical form meaning representations. Section 3 presents the formalism, mostly drawn from Cohen and Levesque [1], including the base Horn clause logic, the modal operators, and the action representation. Section 4 shows how Eva's Horn clause modal logic meta-interpreters can reason about and maintain the system's rational balance among its mental states. Given this machinery, Section 5 discusses how speech acts are represented as planning operators, illustrated with three types of questioning speech acts along with requests. Section 6 presents our approach to collaboration, especially its reliance on planning and plan recognition.

Eva is driven by its beliefs about its own and its user's goals and intentions, so it is important to know where goals come from. In particular, in Section 6 we show how goals arise during the course of planning, including how a model of the user's mental states underlies the system "slot-filling" capabilities. Section 8 describes our BDI architecture, which reasons with these mental state formulas to form plans of action incorporating both domain and communicative actions. This architecture is shown to have an operational semantics based on the specification of rational interaction provided by Cohen and Levesque [1]. Section 9 discusses how the system chooses which of its many intentions to execute next, and Section 10 presents Eva's approach to maintaining and using context. Among the important features of the example and system is how Eva handles requests for explanation during the dialogue. We discuss how explanation is accomplished in Section 11.

Because this research digs deeply into the past four decades of a number of branches of AI research, we cannot possibly do justice to all the relevant work, but refer the reader to surveys of some of the relevant literature [50, 51]. We provide some of the major references in the text, and discuss in Section 12 some of the more recent work that is treading the same ground. The appendices provide the details of the formalism, mostly taken from Cohen and Levesque [1], and some of the speech act definitions.

# 2 Overall System Architecture

Referring to **Figure 1**, Eva takes as input speech and other modalities (e.g., gesture, sketch, touch, vision, etc.), parses them into logical form meaning representations and fuses their meanings into logical forms (LFs) that incorporate one or more speech acts (sometimes referred to as dialogue acts). Using the same representation of speech acts for planning and plan recognition [8, 9, 14, 17, 20, 52–54], the LF is input to a plan recognition process that attempts to infer why the user said/did what was observed. Once a user's plan is derived, Eva adopts the user's goals as its own if they do not conflict with its other goals and obligations. The system then collaborates by attempting to find obstacles to the plan [8, 9, 13, 17, 20, 52, 53], which it plans to overcome in order to help the user achieve their higher-level goals, resulting in intended actions. Afterwards, it executes (some of) those intended actions, which may well involve communicating with the user, generating linguistic and multimodal output, including text-to-speech, graphics, and avatar behavior. In the course of this processing, the system may access backend databases and commercial systems to gather and/or update required information and take needed actions. We only discuss the dialogue manager in this paper, but first we mention the system's inputs and outputs.

## 2.1 Natural Language Parsing and Generation

For many current task-oriented dialogue systems, the meaning representation is simply an "intent+slot" representation of an action and its arguments that, it is assumed, the system is being requested to perform [46, 55–57]. However, this is too simplistic a meaning representation to support logically expressive dialogues. Eva's LF meaning representation involves more complex formulas that express both the speech actions that the parties are doing as well as the content of their utterances, which not only includes domain actions to perform[8], but also complex operator combinations (e.g., comparatives, superlatives, Boolean combinations, temporal constraints, etc.), and operators over actions (e.g., "quitting smoking," "permitting to be interviewed"). We provide in Section 12.2 a detailed comparison of the slot-filling approach to dialogue with our plan-based approach that incorporates true logical forms (see also Cohen [49])). Eva maps utterances into "surface speech acts" [9, 14], from which it can infer the intended meaning of indirect speech acts.

Although this paper will not discuss natural language processing *per se*, we briefly mention in reference to Figure 1 that Eva employs a deeply learned neural sequence-to-sequence semantic parser whose encoder is a pre-trained CodeT5 language model that we have fine-tuned on pairs of utterances and logical form (LF) meaning representations. The parser returns an LF representation that incorporates both the surface speech act as well as its propositional content. These LFs are then input to the plan recognition component that starts reasoning about the user's speech act(s) and what s/he was trying to achieve in performing them. The training of the parser is begun with "canonical" (affectionately known as "clunky") utterances generated from logical forms that are derived from the backend application, as in the "Overnight" approach [58]. These canonical utterances are then paraphrased into natural language, using both machine resources and crowd-sourcing. Because the system makes use of a large-scale multi-lingual language model during the parsing process, when building the system to be used in a language other than English, a relatively small number of human-generated paraphrases of the canonical utterances can be gathered in that language and added to the training data [59].

Natural language generation uses a hierarchical generator driven off the logical form structure that creates "canonical" utterances (the "clunky form"), which are then post-processed with a small set of rules to produce reasonable English output. This output is further passed on to translation services to produce output in other languages supported by our system. We have also successfully explored using LLM-based paraphrasing to go directly from clunky form to human-like surface output. Thus, our symbolic dialogue system determines *what* to say while the LLM decides *how* to say it. We do not use large language models by themselves as generators because they are not sensitive to what the system is intending to convey, potentially resulting in inappropriate or untruthful utterances at that stage of the dialogue. In addition, a generator needs to decide such issues as when to issue a pronoun, how to refer to entities, when to issue discourse markers, etc. So far large-scale language models do not provide general solutions to such problems.

[8] They can include descriptions of actions to *not* do (e.g., "*please don't cancel my insurance*").

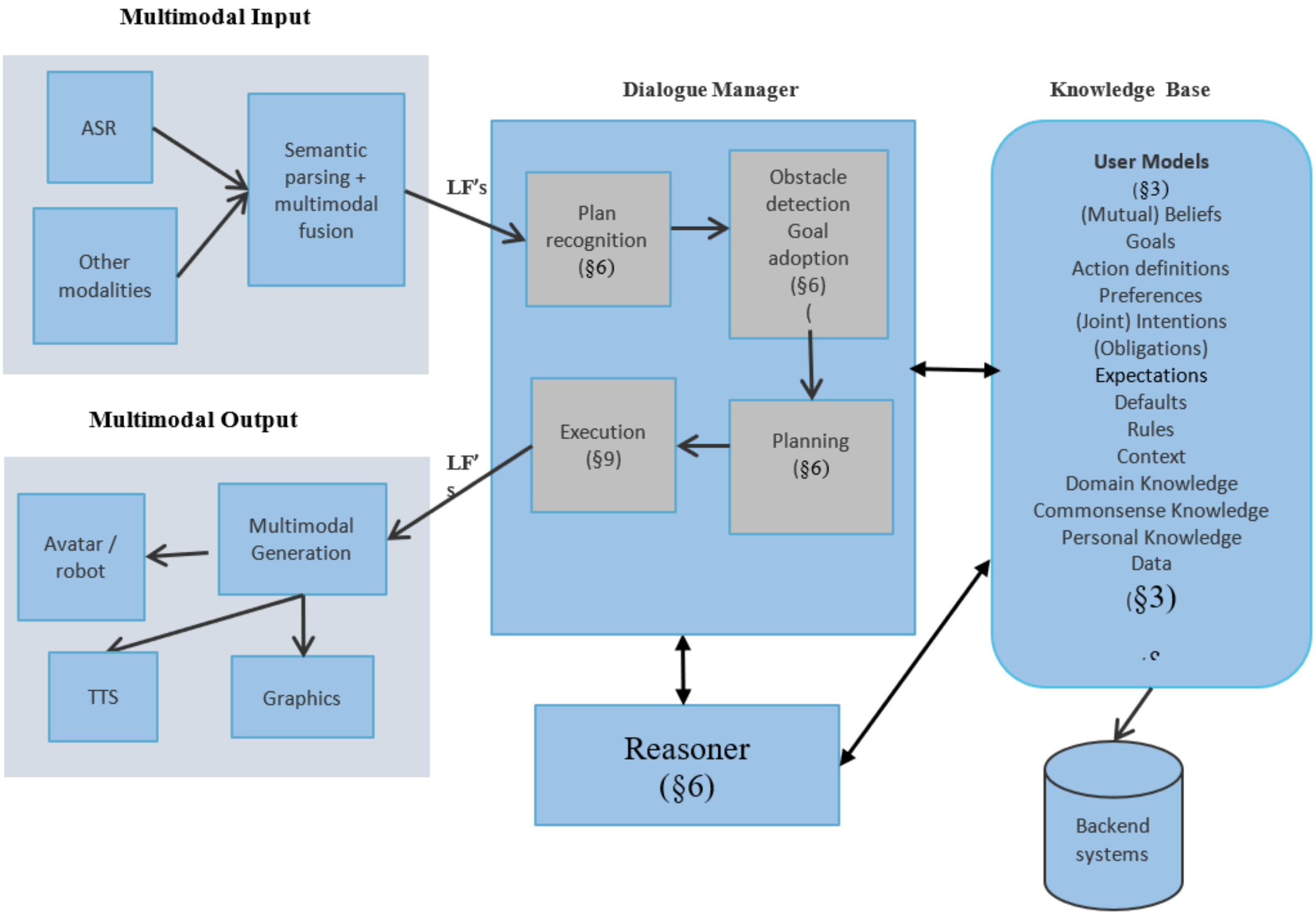

**Figure 1. Eva's architecture of a collaborative, planning-based dialogue system**

Because this paper is about collaborative dialogue, not the natural language processing itself, we will not delve further into the details of the NLP in the rest of this paper. Let us assume the parser can provide a proper logical form, and a generator can produce natural language from a logical form.

## 2.2 Multimodal Input/Output

Eva has been given a multimodal avatar embodiment (see **Figure 2**) that accepts spoken language, camera-based input, and text, and produces spoken, textual, GUI, and face/head gestures as output. Below is a screen shot of a recent dialogue. The system tracks various emotional states and engagement in real-time, enabling it to generate responses to the user's signaled states, using its model of the user's beliefs and intentions [60]. Conversely, the system can generate facial gestures and rapport-building utterances to signal various emotional states, such as sadness. The cues for emotions (visual and via text-to-speech) are based on the words being generated, the logical forms generated in context, and discourse-level features (e.g., topic shifts). As an example, consider if the system asked the user "*Was there a reaction to that vaccination?*" and the user said, "*Yes*." The system can generate an empathic utterance (e.g., "*That's awful!*"), even though the user issued a "positive" or "neutral" utterance. There is more to say about Eva's multimodal capabilities, which will be covered in another paper.

# 3 Knowledge Representation and Inference

Below we present the knowledge representation that Eva uses, which is encoded as modal logic formulas describing the system's and the users' beliefs, goals, and intentions. The reason for encoding the system's knowledge in a logic is that it affords the more expressive representations required for engaging in substantive dialogues about tasks. In the sections that follow, we will provide extensive examples of the kinds of representations the system maintains and its associated reasoning. Our purpose then is to develop the tools sufficient to engage in expressive dialogues, not to focus on logic and reasoning *per se*. Rather, we believe any system sophisticated enough to engage in such dialogues will need to make the distinctions that are encoded herein.

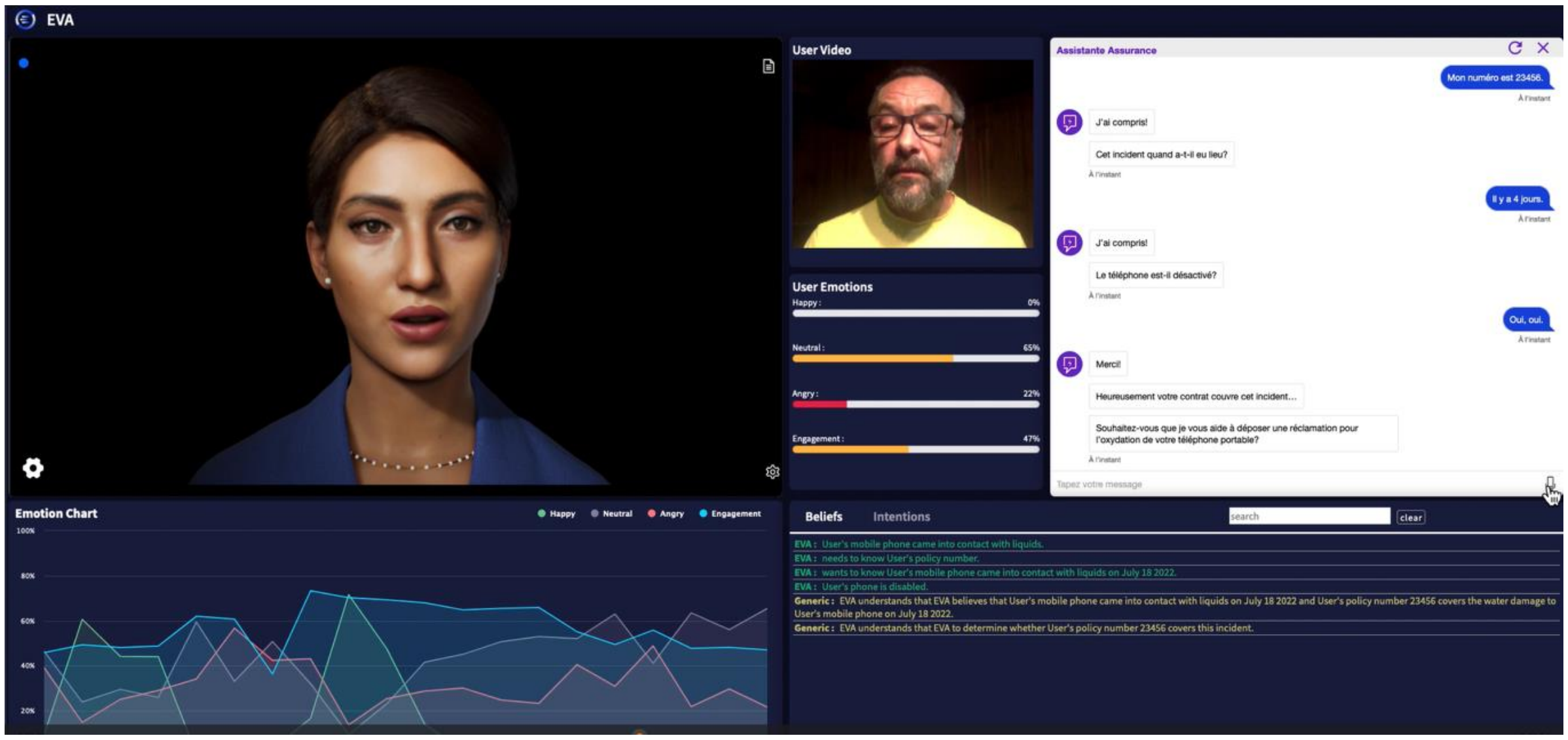

**Figure 2. Eva avatar, vision-based emotion recognition, dialogue, and a snapshot of the system's beliefs, goals, and intentions**.

In this section and in Appendix A, we describe the formalism and its semantics, drawn from Cohen and Levesque [1]. The Eva system uses a Horn clause-encoded first-order modal logic, with constants, variables, typed terms, n-ary predicates, conjunctions, disjunctions, negation of literals, and existential quantification. In addition, we include second-order predicates, such as superlatives, set operators, etc. Predicates and actions take a list of arguments, with each argument specified by a role name, a variable, and a type drawn from an ontology. Syntactically, we adopt the Prolog convention of signifying variables with capital letters. Thus, an argument list will be of the form: [*Role:Variable#Term …*]. Note that this use of the term 'role' is derived from natural language processing, as opposed to the planning literature. From here on, assume all arguments presented have this form, though we will only typically specify the Variable. Importantly, Eva's knowledge representation incorporates a belief operator (*bel*), along with two operators defined in terms of *bel*, for knowing whether a formula holds (*knowif*) and knowing the referent of a description (*knowref*). Likewise, it incorporates persistent goal (*pgoal*) and *intend* as modal operators, with *pgoal* taking a formula as an argument, and *intend* taking an action expression (see 3.5). Attribution of mental states, as well as plan recognition and planning, involves uncertainty. Each mental state has a probability argument, which will not be discussed in this paper. The probability of a pgoal or intention declines as more rules are applied, and as fan-outs (disjunctions) are created during planning and plan recognition. Where available, prior probabilities on intended actions are multiplied, lowering the score. A separate utility calculation could be performed for each intended action, with some domain predicates having negative utility (e.g., *dead(user)*), and some positive utility (e.g., *vaccinated(user)*). We make no claims to having a theory of what utilities various actions and states should have.

For understanding the formulas and reasoning that Eva uses, we note briefly here:[9] Predicates over action expressions include: *do*, *doing*, and *done.* We also allow specific times to be arguments and have special "before" and "after" operators that apply to them. Eva takes advantage of basic Horn clause reasoning (as in Prolog), and incorporates two meta-interpreters for forward ('→') and backward ('istrue') modal logic reasoning, which are used to assert and to prove formulas, respectively. Prolog incorporates the not-provable operator '\+', and does-not-unify operator '\=', both of which we also use in the meta-interpreter. Thus, system tries its best to prove a proposition *P* given the rest of the system's mental states and inference rules, but non-provability of P is different than proving ~P.

[9] The full syntax of our language is given in Appendix A.

## 3.1 Mental State Modal Operators

The attitude modalities we employ are Belief (*bel*), Persistent Goal (*pgoal*), and Intending to Perform Actions (*intend*),[10] along with two defined operators, *knowif* and *knowref.*

### 3.1.1 Belief

Modal logics of knowledge and belief have been well-studied, dating back to Kripke [61] and Hintikka [62]. We will use a Kripke-style possible worlds semantics of belief, such that propositions that an agent believes are true in all possible worlds that are related to the given one (the world the agent thinks s/he might be in). Thus, for a well-formed formula *P* we say:

***Syntax**: bel(X, P*) — agent *X believes* formula *P*, if P is true in all of X's belief-related worlds. (See Appendix A).

If *P* is true in some of *X*'s belief-related worlds, but not in all of them, then the agent neither believes *P* nor believes ~*P*.

The *bel* modal operator also has a positive introspection property — if the agent has a belief that *P*, it has a belief that it believes that *P*, and conversely. However, we do not adopt a negative introspection property -- the agent does not believe that *P*, it does not have to believe that it does not believe that *P*.

For example, we will want to be able to state such formulas such as (after utterance 21 in the example from section 1.1) that the user believes that the system wants to find out the date that user wants the system to make an appointment for the user. One can see that there are multiple embeddings of modal operators in such a formula. By the end of this section, the reader will be able to see how such formulas can easily be expressed. The syntax and formal semantics of our modal operators is given in Appendix A.

### 3.1.2 Goal

The model of goals adopted in Cohen and Levesque [1] is that goals encode agents' choices. This is modeled in a possible worlds semantics by having goal formulas *P* be true in all of the agent's chosen worlds. These are not simple desires because desires can be inconsistent. However, the goals *P* can be false at some time and true at others, as would be the case for achievement goals. Among the worlds compatible with the agent's beliefs are all the ones compatible with the agent's goals (choices). In other words, by assumption in Cohen and Levesque [1], the chosen worlds are a subset of the worlds consistent with the agent's beliefs. [11] That does not mean that if the agent has *P* as a goal, it believes *P* is true. Instead, we require the converse (see Appendix A) — if the agent believes formula *P* is true, it is true in all belief-compatible worlds, so it is true in all chosen worlds, which are a subset of the belief-compatible ones. The agent must choose what it believes to be true "now."

***Syntax**: goal(X, P) — X has a goal that P is true, if P* is true in all of *X*'s chosen worlds. Because of the way formulas are evaluated in terms of a world and time, *P* is true in the chosen world at the time of evaluation. Most often, however, *P* will incorporate a temporal predication.

Agents are not unsure of their goals – if an agent has the goal that *P*, it believes it has the goal that *P* and vice-versa. Given the realism constraint discussed above, the goal modal operator also has a positive introspection property — if the agent has a goal that *P*, it has a goal that it has the goal that *P*, and conversely. However, as with belief, we do not have a negative introspection property of goal itself. Finally, based on the semantics given in Appendix A, an agent cannot have as a goal a formula that it believes to be impossible (the agent believes it will never be true.)

## 3.2 Basic Axioms of the Modal Logic

Given these operators, we provide axioms that they need to satisfy. We assume all axioms of first-order logic. The system supports belief and goal reasoning with a KD4 semantics and axiom schema [64, 65] (see below and Appendix

[10] We are also starting to experiment with deontic concepts as well, including *permit* [41].

[11] In Cohen and Levesque [1], we called this the *realism constraint*. Some authors (e.g., [63]) have suggested that the possible worlds underlying beliefs should merely intersect with those that underlie goals. That means that among one's goal worlds are worlds the agent may think are impossible. Cohen and Levesque [1] viewed that mental state as incoherent, and opted for the agent's accepting what it cannot change (after all *pgoal*s are achievement goals).

A for details). Specifically, *bel(X,P)* and *goal(X,P)* means P "follows X's beliefs/goals". '|=' means 'is satisfied in all worlds':

K: *If P is a theorem*[12], $\models$ *bel(X,P)* and $\models$*goal(X,P)* – theorems are true in all worlds

$\models bel(X,P \supset Q) \supset (bel(X,P) \supset bel(X,Q))$ – agents can reason with their beliefs

D: $\models bel(X,P) \supset \sim bel(X,\sim P)$ – agents' beliefs are consistent

$\models goal(X,P) \supset \sim goal(X,\sim P)$ – agents' goals are consistent[13]

4: $\models bel(X,P) \equiv bel(X, bel(X,P))$ – positive belief introspection

$\models goal(X,P) \equiv goal(X, goal(X,P))$ – positive goal introspection

*Realism:* $\models bel(X,P) \supset goal(X,P)$ – agents' chosen worlds include what they believe to be currently true.

For this system, material implication '$\supset$' is approximated using Prolog's Horn clause reasoning.

Whereas the axioms above license the system's belief reasoning as sound relative to the semantic model, the system is incomplete and does not derive all the logical consequences of its beliefs.

## 3.3 Defined Modal Operators

A critical aspect of a system that reasons about its users' mental states is that it can only have an incomplete model of them. Specifically, we need the system to be able to represent that an agent "knows whether or not" a proposition is true, without its knowing what the agent believes to be the case. For example, we might model that the user knows whether or not her car is driveable. Likewise, the system needs to represent that an agent knows the value of a function or description, without knowing what the agent thinks that value is. For example, the user knows his/her birthdate, social security number, etc. Because Eva does not use a modal operator for knowledge, we define the concepts in terms of belief below.[14] Whereas a number of recent epistemic planning systems incorporate a *knowif* operator, as applied to propositional formulas, none of them incorporate a modal logic *knowref* operator as applied to first-order formulas and their arguments.

### 3.3.1 Knowing whether (*knowif*)

Given the modal operator for belief, we can define operators that capture incomplete knowledge about another agent's beliefs. Specifically, we define an agent's "knows-whether" or "knows if" a formula P as follows [8, 9, 13, 14, 39, 66, 67] .

$$knowif(X, P) =_{def} bel(X, P) \vee bel(X, \sim P)$$

In words, an agent *X*'s 'knowing whether or not' *P* is true is defined as either *X* believes *P* is true, or *X* believes *P* is false. Notice that this is different between *X*'s believing (*P* or *~P)* holds, which is a tautology. One agent *X* can believe that another agent *Y* knows whether or not *P* holds without *X*'s knowing what *Y* in fact believes. However, *X* believes *Y* is not undecided about the truth of *P*. For this reason, in our prior example S can ask CVS a yes-no question as to whether CVS has covid vaccine because S believes CVS knows whether or not it has vaccine.

### 3.3.2 Knowing the referent of a description (*knowref*)

Of critical importance to dialogue systems is the ability to represent and reason with incomplete information about functions, values, etc. Specifically, a system needs to represent *that* an agent knows what the value of the function or term is without knowing what that agent thinks that value is (or else the system would not need to ask). For example, we should be able to represent that John knows Mary's social security number without knowing what John thinks it is. This is weaker than believing that the agent thinks the value is a constant, but stronger than believing that the agent thinks there merely is a value (i.e., Mary has a social security number). The classical way to represent such expressions in the philosophy of mind is via a quantified-in belief formula [68–70]. Accordingly, we define an agent X's knowing the referent of a description as[15]:

$$knowref(X, Var\text{^}Pred) =_{def} \exists Var\ bel(X, Pred), \text{ where } Var \text{ occurs free in } Pred$$

---

[12] See definition of Satisfaction ("|=") given in Appendix A.

[13] This is different from having inconsistent desires. Goals represent choices, and those choices are taken to be consistent.

[14] Unfortunately, English does not have a nice verbalization for "believes if", so please refer to the definition when "knows if" is used. Similarly, when we say "the user knows the referent of a description", we will be using a definition in terms of belief.

The semantics for quantifying into beliefs and goals is given in Appendix A; in short, this expresses that there is some entity of which the agent believes has the property *Pred*. Semantically, a value is assigned to the variable *Var*, and it remains the same value in all of the agent's belief-compatible worlds. For example, the system may believe that agent *X* knows the referent of *X*'s social security number, without the *system's* knowing what that number is. Notice that this is different than the system's believing *X* believes *X* has a social security number, for which the existential quantifier is *within* the scope of *X*'s belief. We argue below that the overriding concern of the present task-oriented dialogue literature involving "slot-filling" is better modeled with goals to *knowref.*

Building systems that could represent and reason with this formula was a major accomplishment of the early work in plan-based dialogue systems [8, 9, 14, 34, 52, 71, 72], and see also Moore [73]. Yet, apart from the work from Allen's group in Rochester, no recent systems have undertaken to plan speech acts (or any acts) using such representations. Indeed, the many epistemic planning systems that use the PDDL language or other propositional logic base cannot do so. Cohen [49] shows that the task-oriented dialogue paradigm can be recast in our modal logic incorporating these expressions for so-called "slot-filling" goals. Moreover, we show there and here how one needs to be able to quantify into modalities other than just belief, such as persistent goals.

### 3.3.3 Intention as a species of persistent goal (*pgoal*)

Intention is a concept critical to virtually all analyses of human cognition, including common sense, legal, philosophical, psychological, practical, linguistic, pragmatic, and social analyses. The philosophies of mind and language include subspecialities on theories of intention, including notable works by Anscombe [74], Bratman [75], Grice [26], Searle [76], and many others. Following Grice, intention is a critical concept to enable an agent to reason about what a *person meant* in saying something, which led the Toronto group to focus on plans and plan recognition. However, the concepts that their plans used in that early work were only initial approximations to mental states. Cohen and Levesque's analysis [28], which was further developed in Cohen and Levesque [1], provided a formalization of intention in a modal logic that shared with Bratman's the notion of commitment. However, unlike Bratman, Cohen and Levesque [1] defined intention in terms of goals (choices) that persist subject to certain conditions driven by the semantics and by the relativization of intentions on other mental states. Such internal commitments were shown to satisfy Bratman's desiderata for a theory of intention.

The representation of intention as a species of persistent goal (*pgoal*) is the basis for Eva's operation.[16] The idea is that simple goals can be ephemeral, so one cannot count on a simple goal or choice as having any consequences on an agent's future actions. Rather, agents need to be committed to their choices [28, 75]. We will typically assume in this paper that the phrases "want" and "need" will be translated into *pgoal* in the logical forms.

We represent by *pgoal(X, P, Q)* the fact that agent *X* has a persistent goal to bring about the state of affairs satisfying formula *P,* relative to formula *Q*, and define it as:

$$pgoal(X, P, Q) =_{def}$$

$$goal(X, \Diamond P) \land \sim bel(X, P) \land before((bel(X, P) \lor bel(X, \Box\sim P) \lor \sim bel(X, Q)), \sim goal(X, \Diamond P))$$ [17]

That is, *X* does not currently believe *P* to be true and has a goal that *P* eventually be true (*X* is thus an achievement goal); *X* will not give up *P* as an achievement goal at least until it believes *P* to be true, impossible, or irrelevant. By

[15] We read the '^' operator as "such that".

[16] We show here how to represent and reason with intention and persistent goal, and how to build a system that behaves according to the principles in Cohen and Levesque [1]. One could perhaps say that the system "has" approximations of intentions, plans, beliefs, etc., up to a level of delicacy. We do not stake a claim on whether taking this "intentional stance" [77, 78] as seriously as we do means that the system in fact "has" intentions, only that it behaves according to the principles of intention.

[17] The definition below refers to linear time temporal operators, including ◊ for eventually, and □ for always. See Appendix A.

'irrelevant' we mean that some other condition *Q*, typically encoding the agent's reasons for adopting *P*, is no longer believed to be true.[18] We frequently will relativize one persistent goal to another as part of planning. If the agent does not believe Q, it can drop the persistent goal. However, if none of these conditions hold, the agent will keep the persistent goal. Cohen and Levesque [1] showed conditions under which having a persistent goal would lead to planning and intention formation. See the formal semantics in Appendix A.

For example, in utterance 19 of the sample dialogue, the system asks the user if the user wants (has as a *pgoal*) that the system make an appointment for the user. The *pgoal* that led it to plan this question is represented as[19]:

*pgoal(sys, knowif(sys, pgoal(usr, P, Q)), R),* where

*P = done(sys, make_appointment(usr, Business, Date, Time))*[20]

That is, the system has a persistent goal relative to *R* to come to know whether the user has a persistent goal to achieve *P*, relative to *Q*. When the user says "yes", the system adopts the persistent goal to make an appointment, which is relative to the user's persistent goal that the system do so (*R*). If the system learns that the user no longer wants the system to make the appointment, the system can drop its persistent goal and any parts of the plan that depend on it. Otherwise, the system will keep that persistent goal and plan to achieve it.

As a second example, in utterance 21, the system asks what date the user wants the appointment. This question is planned because the system has created a *pgoal*, relative to its wanting to make an appointment for the user, to know the *Date* such that the user has a *pgoal* (relative to some other proposition Q) that the system make an appointment on that Date.

*pgoal(sys, knowref(sys,Date^pgoal(usr, P, Q), R),* where

*P = done(sys, make_appointment(usr, Business, Date, Time))*

### 3.3.4 Intention to do (*intend*)

Agent *X*'s intending to perform action *Act* relative to formula *Q* is defined to be a persistent goal to achieve *X*'s eventually having done *Act*:

$$intend(X, Action, Q) =_{def} pgoal(X, done(X, Action), Q).$$

In other words, *X* is committed to eventually having done the action *Action* relative to *Q*. An intention is thus a commitment to act in the future.[21] Notice that intentions take (potentially complex) actions as arguments and are relativized to other formulas, such as pgoals, other intentions, and beliefs. If a relativization condition is not believed to be true, the agent can drop the intention. For example, in the sample dialogue at utterance 21, we can see the system's forming the intention to ask a wh-question relative to the persistent goal to *knowref*, and then executing the intention.

This notion of intention is different than the current (mis)use of the term "intent" in the "intent+slot" dialogue literature as it incorporates an internal commitment to act as part of a plan and relative to having other mental states. "Intent" in the dialogue literature is basically taken to refer to an action that the user supposedly wants the system to perform, or occasionally, to refer to an ordinary predicate. Here, we treat *intend* as a modal operator, with a formal logic and semantics (see Cohen and Levesque [1]).

---

[18] Cohen and Levesque [1] had the agent's believing the relativization condition was false. In this system, we weaken this constraint somewhat by having the agent not believe the relativization formula is true.

[19] Hereafter we use *sys* and *usr* as constant entities (of type agent) to identify the system and the user, respectively.

[20] For ease of readability, we omit the existential quantifiers within the scope of *pgoal* binding Date and Time here and elsewhere in this paper.

[21] Cohen and Levesque [1] also had the agent believing it was about to do the action. We omit this condition for simplicity.

### 3.4 Defaults

Eva operates with a collection of defaults [79], many of which are targeted at its user model and domain facts. The general default schema is, using Prolog operators ':-' (is true if'), '\= '(not unifiable) and '\+ '(not provable):

*istrue(P) :- default(P), P \= ~Q, \+ ~P.*

That is, *P* is true if *P* is stated as a default predicate, *P* is not the negation of a formula *Q*, and it cannot be proven that *~P*. If *P* is a default formula, but its negation can be proven, then *istrue(P)* does not hold. These are among the normal defaults (Reiter, 1980).

For example, we might have the following default schemas:

*default(driveable(car_of(usr)))* — by default, the user's car is driveable.

*default(knowif(usr, damaged(mobile_phone(usr))))* — by default, the user knows whether the user's phone is damaged.

*default(knowif(usr, pgoal(usr, Pred, Q)))* — by default, the user knows whether the user has a *pgoal* that *Pred* be true.

*default(knowif(usr, knowref(usr, Var^Pred)))* — by default, the user knows whether the user knows the referent of *Var^Pred*.

The previous two defaults schemas have schematic variables for the interesting formulas and variables (*Pred* and *Var*, and of course the relativizer *Q*). When those are provided with actual formulas and variables, a specific default would be queried.

### 3.5 Action Expressions and their Description

Of particular importance to a planning system is the representation of actions. In this section, we provide *action expressions* that incorporate both primitive and composite *actions*. Using an extended STRIPS-like description of actions, primitive actions are described using an *action signature*, which consists of the action name and a list of arguments, where each argument is of the form *Role:Value#Type*.[22] The only requirement for an action is that there be an *agent* role and filler of type *agent* (which is a concept in the ontology). **Figure *3*** shows an English gloss of the description of the action of a vaccine center's vaccinating a patient at a location, date and time. Composite action expressions will be detailed below. We will use the term "action" to mean both primitive and composite action expressions.

Action descriptions state, in addition to the signature, the action's *Precondition*, *Effect*, *Applicability Condition*, and *Constraint*, which are well-formed formulas. The *Precondition* is that the Patient has an appointment at a certain *Date* and *Time*, and that the *Patient* is located at location *Loc*, which the *Constraint* stipulates to be the location of the vaccination center. The *Effect* states conditions that the agent wants (i.e., has as a *pgoal*) to be true at the end of the action. The *Effect* stated here is that the *Patient* is vaccinated by the *Center*.

We define four functions that operate over action expressions *Action*, providing the well-formed formulas *P* that are an action's precondition, effect, constraint, applicability condition.

*precondition(Action, P)*

*effect(Action , E),*

*constraint(Action, C),*

*applicability_condition(Action, A)*

---

[22] For terms in the form *Role:Value#Type*, the *Role* can be any atom. We intend the meaning of "role" to be similar to the notion of (thematic) role used in the natural language literature, namely as a labeled argument in a predicate or action. That is different from the meaning often found in the planning literature in which multiple agents may play different roles (e.g., pilot vs. wingman). Sometimes, one can encode the latter with expressions using roles, fillers, and the types of those fillers, but perhaps not always.

*Center* ***vaccinates*** *Patient against Disease with Vaccine*

*Precondition: Patient has an appointment at Center at Date and Time*

*Patient is located at Loc at Date and Time*

*Constraint: Center is located at Loc*

*Effect: Patient is vaccinated by Center with Vaccine*

*Applicability Condition: Center has Vaccine and Patient is eligible for Vaccine*

**Figure 3. The action description of a Vaccination Center's vaccinating a Patient**

Because the action description is just that, a description, rather than a modal operator (as in dynamic logic), the planner and reasoner cannot conclude that after an action is performed, the effect *E* in fact holds, or that it holds conditioned on the precondition. Furthermore, there is no attempt here to prove that after a (potentially complex) action has been performed, the complex action expression or plan is a *valid* way to achieve the effect (cf. [53]). Rather, as a description of an action, the stated effect *E* is described as the *desired* or *intended effect*, which is realized by an agent's having a *pgoal* to achieve effect *E*.

Given a *pgoal* to achieve *P*, Eva's planning subsystem will attempt to find an action expression whose effect *E* as stated in the action description *unifies* with *P*. If *P* is a complex formula, planning rules will decompose it, and attempt to find more primitive actions that achieve the simpler components. For example, it may decompose effects that are conjunctions into individual plans to achieve the conjuncts. But, as is well-known, this may well be problematic (cf. [80]), and thus Eva's planning is only a heuristic approximation to finding a plan that truly achieves the overall goal *P*. We will discuss below a special case in which we do in fact engage in such goal decomposition. Still, this limitation may not matter for a dialogue system, as we are not attempting to reason in advance about a long sequence of utterances during an interactive dialogue. Rather, the system attempts to take a step towards its goal and react to the inputs it receives, much as other BDI architectures do [81]. However, unlike such architectures, Eva engages in backward-chaining and plan recognition, and reasons about other agents' mental states. Moreover, other BDI architectures are oriented towards communication among artificial agents using KQML communicative actions, which we have criticized elsewhere [82]. Instead, we use a well-founded set of speech acts inspired by natural language communication.

If the agent has the *pgoal* that a formula *E* holds, and that formula unifies with the effect of some action expression *A*, it would add a *pgoal* to perform *A* (i.e., an intention to do *A*). It would then add *A*'s precondition as a *pgoal* to the plan, if it does not believe the precondition holds. This backward-chaining may continue via effect-act-precondition reasoning. Finally, the *Applicability Condition* (*AC*) states formulas that must be true, but cannot be made true by the agent. For example, Figure 3 shows the description of the vaccination action. The *AC* here is that the vaccine center has the vaccine and the Patient is eligible for the vaccine. If such *ACs* are believed to be false (vs. not believed to be true), the agent believes the action *A* is impossible to perform.[23] Thus, the agent would not create a persistent goal or intention to perform action *A* because this violates the definition of persistent goal. If the system does not know whether the *AC* holds, it creates a *pgoal* to *knowif(sys,AC)* and blocks the intention from further inference. If it learns the *AC* is false, and the *pgoal* to do the action has been created, it would remove that *pgoal* and any *pgoals* that depend on it. Further discussion can be found in Section 6. Finally, the system may represent actions hierarchically with actions optionally having a *Body,* which would decompose into a complex action described with the dynamic logic operators below.

Notice that this action expression language is an extension of the hierarchical variants of the Planning Domain Definition Language (PDDL) [83] such as HDDL [84]. In particular, we provide descriptions of composite actions and add preconditions and effects on higher-level actions in addition to the primitives [85]. Also of importance, Eva allows action expressions as arguments to other actions, supporting directive and commissive speech acts like requesting, recommending, etc. We will see such a speech action definition in Section 5.8.

---

[23] We could specialize action definitions such that different agents have different action representations. For example, the vaccine center's representation of the act of vaccinating could have different preconditions, effects, and applicability conditions than the user's representation of that action. Because the user cannot do anything about the availability of vaccine, it is an applicability condition. However, the vaccine center can order more vaccine, so vaccine availability is a normal preconditionthatcan be planned by the Center (e.g., by ordering more vaccine).

As part of the action's signature, the system keeps track of the agent doing it. The action description *action(Agent, Action, Constraint)* indicates that agent *Agent* is the agent of the *Action* expression such that *Constraint* holds. However, *Action* itself has an agent role (call its value *Agent1*). In most cases, the two agents are the same, but they need not be. By allowing for them to be different, we can represent that *Agent* does *something* to *help* or *get Agent1* to perform the *Action.* An example might be submitting an insurance claim on behalf of someone else. Doing actions on behalf of (to the benefit of) someone else may require explicit agreement or permission to do it.

### 3.5.1 Predicates over Action Expressions

These predicates allow us to say that an action will be done in the future (*do*), is being executed (*doing*), or has occurred in the past (*done*). Unlike in Cohen and Levesque [1], we also provide explicit time arguments for these predicates, for which it is simple to provide semantics using the logical tools in Cohen and Levesque [1]:[24]

- *do(action(Agent, Action, Constraint), Location, Time) — Action* will be done at location *Location* and *Time* in the future.[25]
- *done(action(Agent, Action, Constraint), Location, Time) — Action* has been done at *Location* and a past *Time*. [26]
- *doing(action(Agent, Action, Constraint), Location, Time) — Action* is ongoing at *Location* and *Time*.

One additional predicate that we adopt that was not in Cohen and Levesque [1] is *failed(Action, Reason).* Eva uses this predicate when it sends an action invocation request to a backend system (e.g., asking a credit card validation system to validate a particular card number). If the backend system returns a failure, the *Reason* parameter encodes the cause of the failure, such as the card is invalid, overlimit, etc. It can be text sent via the backend or a logical form. Eva assumes that a failure means the action is impossible, so it drops the intention to perform it, though it may try to find another way to achieve the higher-level goal.

### 3.5.2 Complex actions combinators

We provide complex action expressions with combinators drawn from dynamic logic [86], in addition to a combinator for hierarchical decomposition.[27] As we develop the action expressions below, the predicates will become more realistic.

Specifically, we have:

| <u>**Compound actions:**</u> | <u>**Action expression**</u> | <u>**Example**</u> |
|---|---|---|
| Sequential actions: | *seq(A, B)* | *seq(find_identification_number(X,P), informref(X,Y,Q))* |
| Conditional Actions: | *condit(P, A)* | *condit(eligible(X),make_appointment(X,B,D,T))* |
| Non-deterministic Mutually Exclusive Or: | *disj(A, B)* | *disj(inform(S,U,P), inform(S,U,~P))* |

In addition to the four action description functions discussed above, we now add two predicates for defining hierarchical action expressions.

*body(Action, ActionBody)*

---

[24] To improve readability, in this paper we sometimes simplify these expressions and use, e.g., *done(X, Action)* rather than *done(action(X, Action, Constraint), Location, Time)* if the additional arguments are not relevant to the discussion.

[25] Note that *do* is actually redundant given ◊ and *done*. We are not enforcing future and past in the syntax, but in the formal and operational semantics.

[26] This is different than the definition of *done* in Cohen and Levesque [1], which is satisfied immediately before now.

[27] Whereas we use some of the dynamic logic combinators (except iteration), we do not describe actions as a modal logic. That is, we do not provide a semantics for "after action A happens, P is true."

*in_body(ActionBodyElement, Action)*

The first predicate (*body*) maps an action into an *ActionBody* compound action expression. There may be more than one *Action* that decomposes into the *ActionBody*, but each *Action* only has one *Body*. The *in_body* predicate relates an element of an *ActionBody* with the higher level *Action* of which it is a part. It could be one of the sequential actions, or a conditional or disjunctive action. There could be more than one *Action* that contains *ActionBodyElement*. The predicate *in_body* looks for any action within a named complex action expression, searching through the entire complex action library.

As an example of action decomposition relationships, we have the following:

*body(informif(S,H,P), disj(inform(S,H,P),inform(S,H,~P)))* and *in_body(inform(S,H,P), informif(S, H, P)*

This example shows that an *informif* speech act from *S* to *H* that *P* (i.e., informing whether *P* is true) can be decomposed into the disjunctive act of *S*'s informing that *P* or informing that *~P*. The precondition and effect of the *informif* action are determined by the disjunctions of the preconditions and effects of the constituent *inform* actions [30, 39]. Thus, the precondition can be shown to be *knowif(S,P)*. *in_body* shows that an *inform* action is part of the body of the *informif* action.

We define one more predicate to provide the unbound or unknown variables in the *Action*. In order to execute any *Action*, the system needs to know what that *Action* is [73]. Therefore, we define

*unk_oblig_arg(Action, Role:Var#Type)*

to say that *Agt* does not know the value for one of the obligatory variables (namely, *Var*, which is the value, of type *Type*, of the given role *Role*) for an action she intends to execute. That is,

*\+knowref (Agt,Var^intend(Agt, Action, Q))*

with *Var* being a free variable in *Action*. If *Agt* does not know the value of the variable, but a value for that variable is required for the successful execution of *Action*, the system will eventually create a *pgoal* to *knowref* that value which may then lead to planning a question (this process would be repeated for all other unknown obligatory arguments of *Action*).

# 4 Reasoning about Mental States and their Combinations

We now discuss how Eva reasons with the above formulas. There are two meta-interpreters used to reason about modal formulas, one for proving (*istrue*) and one for asserting (→). Modal formulas are proven with *istrue* invoked from a standard Prolog rule ':-'. Non-modal formulas are put into negation normal form (negations only take literals as their scope), and are proven using a standard Prolog interpreter. The assertional meta-interpreter → handles assertions of modal formulas, whereby instead of asserting the left-hand side (LHS), the right-hand side (RHS) is asserted. With it, we ensure that the least embedded formula possible is entered into the database, subject to the logical semantics. This also means that the LHS clause would not be found in the database because the → meta-interpreter is rewriting the LHS into the RHS. Finally, we have a rule operator '=>' for planning and plan recognition rules. The LHS of => is proven using *istrue*, and the right-hand side is asserted with '→'.

The system is driven by its maintaining the rational balance among mental states. Thus, it is able to reason about one agent's goals to get an agent (itself or another agent) to believe or come to know whether a formula is true, to intend to perform an action, or come to know the referent of a description. The semantics of the individual modalities is expressed in the possible worlds framework, which describes the meanings of the combinations of these formulas and the implications of their becoming true. Moreover, the theory of intention in Cohen and Levesque [1] is able to provide a semantics for the internal commitments that the intender takes on, and the conditions under which those commitments can be given up. The system is built to provide an operational semantics that accords with the logical ones through its inference rules and BDI architecture.

In addition to the axioms for belief and goal found in Section 3.1, we give below examples of inference and rewriting rules that the system employs with respect to cross-modal inference. These rules together are an attempt to maintain the least embedded formulae possible, subject to the semantics of *bel*, *goal/pgoal*, *intend*, *knowif*, and *knowref*. Sometimes the same formula will be in both forward and backward reasoning rules below because there may be cases where a condition needs to be proved, even though the system has not yet tried to assert it and thus create the more compact version.

## 4.1 Meta-logical Interpreter: Proving

Proving via ***istrue***: If trying to prove the left side, prove the right side (as in Prolog)[28]

- *istrue(bel(X, bel(X,P))) :- istrue(bel(X, P))*
- *istrue (bel(X, P ∧ Q)) :- istrue(bel(X,P)), istrue(bel(X,Q))*
- *istrue(bel(X, pgoal(X,P))) :- istrue(pgoal(X,P))*
- *istrue(bel(X, knowref(X, Var^Pred))) :- istrue( knowref(X, Var^Pred)))*
- *istrue(bel(X, knowref(X, Var^(Pred, Cond)))) :- istrue(knowref(X, Var^Pred))), istrue(Cond)*
- *istrue(bel(X, exists(Var^Pred))) :- istrue(knowref(X, Var^Pred))*
- *istrue(knowif(X, P)) :- istrue(bel(X, P))* ∨ *istrue(bel(X, ~P))*
- *istrue(bel(X, P)) :- istrue(bel(X, (P :- Q))), istrue(bel(X,Q))* This models the system's being able to reason about another agent's belief reasoning. Here, the agent (*X*) has a belief about a Horn clause rule '*P :- Q'*. There would be assertions in the database about the agent's domain specific Horn clause rules.[29] It approximates[30] with Horn clauses the material implication in axiom K:

  $\models bel(X,P \supset Q) \supset (bel(X,P) \supset bel(X,Q))$

There are also a number of other rules of no particular interest (e.g., meta-interpreting conjunctions and disjunctions).

### 4.1.1 Proving *done* as applied to complex action expressions

There are several predications that take action expressions as an argument, namely *do*, *done* and *doing*. These predicates take a list of arguments, the first of which specifies the agent, then the action, and finally the location and time. We take *do* and *done* to be satisfied at a future/prior time, respectively (as they take a specific time as an argument). In addition to taking primitive actions as arguments, these predicates are defined over complex action expressions, such as conditional, sequential, and disjunctive action expressions. For example, a disjunctive action expression has been done at some location and time if either of its disjuncts has been done at that location and time.

*istrue(done(action(Agt, disj(Act1, Act2), Constr), Loc, Time)) :-*

*istrue(done(Agt, Act, Constr), Loc, Time)* ∨ *istrue(done(Agt, Act2, Constr), Loc, Time))* **.**

Currently, we have found it sufficient to say that a conditional action *condit(P,Act)* has been done if *Act* has been done and the predicate *P* is true.[31]

*istrue(done(action(Agt, condit(Pred, Act), Constr), Loc, Time)) :-*

*istrue(Pred)* ∧ *istrue(done(action(Agt, Act, Constr), Loc, Time)).*

---

[28] Comma ',' is used both to separate arguments of terms, and also as a meta-language conjunction. So, if P → Q, R, if the forward meta-interpreter is asked to assert P, it will assert Q and will assert R, but is not asserting the object language conjunction of *and(Q, R)*.

[29] This rule also applies to the system's beliefs, which eventually cashes out *istrue* into the system's Prolog belief database. However, for efficiency, we would insert a predication that *X* is not equal to *system*.

[30] For example, material implication allows contrapositive reasoning, but Horn clause reasoning does not.

[31] To be more precise with respect to *done*, Eva could check if the condition *P* held at the time the action was performed (*Time*) rather than at the present time. To do the former, we would need a time argument for every predicate (fluent), and to know when the interval of its being true terminates (or has not yet terminated). The classic way to do this is with a *holds(Pred,Timeinterval)* predicate. This would be easy to adopt.

We will say a sequential action of *seq(Act1,Act2)* has been done if *Act1* has been done and *Act2* has been done afterwards in a circumstance in which *Act1* has been done.

*istrue(done(action(Agt, seq(Act1, Act2), Constr), Loc, Time)) :-*

*istrue(done(action(Agt, Act1, Constr), Loc1, Time1)) ,*

*istrue(done(action(Agt, condit(done(action(Agt, Act1, Constr), Loc1, Time1), Act2), Constr), Loc, Time)).*

The *doing* predicate applies to actions that have a hierarchical decomposition, such that once the higher-level action has been decomposed into its body, and the system is executing one element of the body, then the system asserts that it is *doing* the higher-level action. If the system has *done* the last element of the body, then, for the higher-level action *doing* is retracted and *done* is asserted.

We currently do not reason with the full linear temporal logic of Cohen and Levesque [1]; see Gutierrez et al. [87] for an example of how to do that in a multi-agent context.

## 4.2 Meta-logical Interpreter: Asserting as Rewriting

The meta-logical interpreter uses rules of the form (*LHS* → *RHS*) to infer that, whenever it is supposed to assert the left-hand side (*LHS*), instead it should assert the right side (*RHS*). Importantly, these are rewriting rules, not inference rules, in that *LHS* is not asserted to be true. In the rules below we will use the expression 'and' to mean that what follows it is a constraint. That is, when the agent is trying to assert the first literal, if the formula following 'and' is true, then the the literal is rewritten with the right-hand side.

- *bel(X, bel(X,P))* → *bel(X, P)*
- *bel(X, P ∧ Q)* → *bel(X,P), bel(X,Q)*
- *bel(X, knowref(X, Var^Pred))* → *knowref(X, Var^Pred))*
- *knowref(X, Var^bel(X, Pred))* → *knowref(X, Var^Pred)*
- *knowref(X, Var^Pred))* and *Var* is a constant → *bel(X, Pred)*
- *pgoal(X, P ∧ Q, R)* → *pgoal(X, P, R), pgoal(X, Q, R)*[32]
- *pgoal(X, pgoal(X, P, Q), Q)* → *pgoal(X, P, Q)*[33]
- *pgoal(X, intend(X, A, Q), Q)* → *intend(X, A, Q),* where *A* is a (potentially complex) action. If agent X wants to intend to do action A, then the agent in fact intends to do A.
- *pgoal(X, knowref(X, Var^Pred), Q)* and *Var* is a constant → *pgoal(X, bel(X, Pred), Q)*

  During reasoning, *Var* can become bound to a constant. Because *knowref* is defined to existentially quantify *Var* into the agent's beliefs, the possible worlds semantics shows that the agent believes *Pred* is true of that constant.
- *bel(X, pgoal(X, P, Q))* → *pgoal(X,P,Q)* — If the system tries to assert that the agent believes it has a goal, then assert that it does have the goal.

Finally, a word about soundness and completeness. We make no claims that the reasoning system is complete, however relative to the approximations we have made, we do believe it is sound.[34] Interestingly, there is a propositional variant of the logic in Cohen and Levesque [1], from which we draw, that has been shown to be sound and complete [89]. Rao and Georgeff [90] also proposed sound and complete branching-time multi-modal propositional logics for belief, desire, and intention. Their intention operator corresponds to Cohen and Levesque's goal operator, and they have no analogue to persistent goal or intend as we use them here. In comparison to both approaches [89, 90], however, we cannot use a propositional logic for our system, since it would make it impossible to reason about *knowref*, which is so essential to task-oriented dialogue. Likewise, we need to have arguments to actions that may be existentially quantified ("I intend that *someone* tow the car to the repair shop"). Another major difference with the aforementioned logics [89, 90] is that they do not consider intentions to perform actions, but only to make formulas true. For Eva, its intentions to perform (possibly complex) actions is key to its functioning.[35]

[32] CL 90 require that the pgoals P and Q be true at the same time, but we ignore that here.

[33] Here is an example in which the left-hand side (LHS) and the right-hand side (RHS) cannot co-exist in the database because an agent's having a pgoal means that the agent believes the content is false, which contradicts the RHS. This rewriting rule essentially keeps the database consistent if some reasoning step attempted to create the LHS.

[34] Researchers have fretted about the complexity of reasoning with such formulas and have suggested using programmatic BDI architectures to avoid the problems [88]. However, the restrictions inherent in such systems would not support the kinds of dialogues we target. On the other hand, many advocate end-to-end trained neural network-based dialogue systems (e.g., [46]). Apart from simple slot-filling, it is arguable whether dialogue systems based on neural networks or large-scale language models can engage in reasoning, much less modal Horn-clause reasoning. Furthermore, large-scale language models are known to not tell the truth, so cannot be said to engage in sound reasoning.

### 4.3 Equality and Reference Resolution

In the course of planning, the system generates equalities between variables appearing in different goals. Unlike graph representations that can accommodate multiple pointers to the same node, sentential reasoning requires that variables appearing in different formulas be explicitly stated as being equal[36]. For example, we record that the covid vaccination center that the user (*u1*) wants to go to is the same as the covid vaccination center at which the user wants to be vaccinated. Likewise, the time that the user intends to be vaccinated is the same as the time at which the user wants to have an appointment.

In the sample dialogue, at some point the system reasons that 'CVS' is a covid vaccination center that satisfies the user's goals, which then enables all of equalities to be resolved if needed.

In general, entities that are equal should be intersubstitutable in formulas. However, equality reasoning is prevented from crossing modal operators. For example, Frege's [91] famous examples "I believe that the morning star = evening star = Venus", but "John does not know that the morning star = the evening star" need to be jointly satisfiable. I cannot use my beliefs of equality to reason about what John believes. However, the system can reason with "the X that John believes to be the morning star = the Y that Mary believes is the evening star" (quantifying X and Y into their respective beliefs). Thus, because the variables are not in the scope of the agents' beliefs (or in other cases, pgoals), the system should be able to reason about X and Y, but not attribute those beliefs to John or Mary. Eva reasons about equality among variables by maintaining equivalence classes of equalities. **Figure 4** shows an example of such equivalence classes created after the first sentence of the example shown in Section 1.1. Notice that these are all quantified-in variables, but the system does not attribute these equalities to the user's beliefs because the semantics of quantified-in goals is that the value is the same in all the agent's goal worlds, but not necessarily in all the agent's belief worlds.

This same equality reasoning mechanism enables the system to represent and resolve co-referential and anaphoric references. Because the equality relationship is declarative, if an ambiguous reference is detected, the system can generate a request to the user to disambiguate the intended referent (See Section 5 for the definitions of speech acts):

*intend(sys, request(sys, usr, disj(inform(usr, sys, (<referring expression> = 'a'),*
*inform(usr, sys, (<referring expression> = 'b'))), Q)*

Of course, how the system finds candidates for coreference or resolves them is a topic of much research (e.g., [92–94]), but not one we have concentrated on to date.

## 5 Speech Acts

In the study of semantics, philosophers and linguists once only analyzed utterances as being true or false. But in his seminal book "How to do things with words," the philosopher Austin [27] upended this approach by arguing that utterances are actions that change the world. Dubbed illocutionary acts or "speech acts", this radical shift in thinking began the study of linguistic pragmatics. John Searle ([76], and in subsequent books) provides a detailed analysis of

[35] One could perhaps use the encoding in Cohen and Levesque [1] of the agent X's intention to perform an action as a persistent goal to make it the case that eventually the action has been done by X (believing that is what it was about to do), but reasoning with the defined concept (intend) is easier for building the system.

[36] Prolog forces this representation as it generates new variables for any formula that is asserted into the database.

**equivalence class 1**

[the X#covid_vaccination_center such that U wants to go from some place to a covid vaccination center]

[the X#covid_vaccination_center such that U intends to be vaccinated for covid at a covid vaccination center]

[the X#covid_vaccination_center such that U may want someone to vaccinate U for covid at a covid vaccination center]

**equivalence class 2**

[the X#time such that U wants someone to vaccinate U for covid]

[the Y#time such that U wants to go from some place to a covid vaccination center]

[the X#time such that U intends to be vaccinated for covid at a covid vaccination center]

[the Y#time such that U wants u1 to have an appointment at a covid vaccination center]

**equivalence class 3**

[the X#date such that U wants U to have an appointment at a covid vaccination center]

[the X#date such that U intends to be vaccinated for covid at a covid vaccination center]

**Figure 4. Equivalence classes of typed variables (of the form X#Type). U is the user.**

many different types of speech acts, at the level of philosophical argumentation. Based in part on some initial analyses of Bruce [10], the plan-based theory of speech acts [8, 14, 52] argued that people plan their speech acts to affect their listener's mental and social states, and showed how speech acts could be modeled as operators in a planning system. Thus, a system can likewise plan speech acts to affect its listener's mental states, and reason about the effects that its listener's speech acts were intended to have on its own mental states. Planning and plan recognition became essential to such pragmatic analyses of language in context because speech acts could then be incorporated into an agent's task-related plans when the agent determines that it needs to know/believe something, or needs to get the user to intend to perform an action.

Given the logic we have provided, especially the tools for describing actions, below are some of the speech acts implemented to date. Note that speech act definitions are domain independent.

## 5.1 *inform*

The first action expression that we will consider is that of an inform by a Speaker to a Listener that formula Pred is true. The precondition for this action is that the Speaker believes what she is saying. The intended effect is stated as the Listener's believing that Pred holds. Recall that we said the listed effect of this action description does not become true as a result of performing the action. Rather, the Listener comes to believe that the Speaker had a persistent goal that the effect holds. In the descriptions of the speech acts we will ignore the constraint parameter and trivial applicability conditions.[37]

***inform****(Speaker, Listener, Pred)*

*precondition: bel(Speaker,Pred),*

*effect: bel(Listener,Pred)*

[37] Cohen and Levesque [7] and Perrault [22] gave differing accounts of the conditions under which even these effects do not hold. Cohen and Levesque analyzed a notion of sincerity, whereas Perrault described the effects in terms of default logic. For purposes of this paper, we only need to state that the Listener's believes the speaker wanted to achieve the stated effect.

## 5.2 *assert*

Assertions are different from informs in that the intended effect is to get the listener to believe that the speaker believes the propositional content, whereas the intended effect of an inform is that the speaker comes to believe the propositional content. Thus, we have:

***assert****(Speaker, Listener, Pred)*

*precondition: bel(Speaker,Pred),*

*effect: bel(Listener, bel(Speaker, Pred))*

## 5.3 *informref*

An *informref* is a speech act whose intended effect is that the listener know what the value of the variable *Var* is such that *Pred* is true of it (*Var* must be free in *Pred*). For example, when the user says "Monday" in response the wh-question "*when do you want to eat?*", the intended effect is that the listener come to know that the referent of "*the date the user wants to eat*" is Monday. The precondition is that the speaker knows what the value of *Var* is such that *Pred*.

***informref****(Speaker, Listener, Var^Pred),*

*precondition: knowref(Speaker, Var^Pred),*

*effect: knowref(Listener, Var^Pred)*

## 5.4 Informing whether

The *informif(S, L, P)* speech action can be defined as a disjunctive action [30]:

*informif(S, L, P) ≡ disj(inform(S, L, P) , inform(S, L, ~P))*

## 5.5 *assertref*

*assertref* is similar to *informref* in that the intended effect of the speech act is that the listener come to know the referent of the variable such that the *speaker believes Pred* is true of it. *assertref* can be used to find out what the speaker believes, even if the speaker is not trying to convince the listener. For example, teacher-student questions or verification questions employ *assertref.*

***assertref****(Speaker, Listener, Var^Pred),*

*precondition: knowref(Speaker, Var^Pred),*

*effect: knowref(Listener, Var^bel(Speaker,Pred))*

Note that *assertref* can be defined in terms of *informref* as:

*assertref(S, L,V^P) ≡ informref(S,L,V^bel(S,P))*

## 5.6 Wh-Questions

A *speaker* asks a wh-question to *Listener* about the referent of *Var* such that *Pred* is true (as usual, *Var* must be free in *Pred*): [38]

---

[38] The *wh-question* can be decomposed into a *Request* to do an *Informref* speech act followed by an *Informref***.** The freedom we have to define such complex speech acts is discussed in Cohen and Levesque [7]. At times, Eva needs to be able reason with these speech acts individually, e.g., when it is is planning a third-party speech act such as "*Ask Bob to tell me the time he wants the appointment*" [8]. The *wh-question* action definition assumes the listener is the same as the agent who will tell the speaker the answer, but having the individual speech acts means the system can plan them directed toward different agents.

***wh-q****(Speaker, Listener, Var^Pred)*

*precondition: knowref(Listener, Var^Pred)*

*effect: knowref(Speaker, Var^Pred)*

Recall that we do *not* claim the *Effect* of a speech act becomes true in virtue of the act's being performed. Because these are planning operators, the *Effect* becomes a *pgoal* of the planning agent. Conversely, on observing an agent performing an action, including a speech act, the observer comes to believe the planning agent had the *Effect* as a *pgoal*.[39] So, on hearing a wh-question, the listener comes to believe that the speaker has a *pgoal* that the speaker come to know the referent of the description.

During backward-chaining, the planner may unify the content of a goal formula with the *Effect* of an action and choose to consider that action as a means to achieve the goal. However, the matching of the *Effect* may not provide a binding for the *Listener*. If the *Listener* is not specified, evaluating the *Precondition* may enable it to determine who knows the answer and ask that agent the wh-question.

### 5.7 Yes-No Questions

A yes-no question is described as:

***ynq****(Speaker, Listener, Pred)*

*precondition: knowif(Listener, Pred),*

*effect: knowif(Speaker, Pred)*

Thus, if the system has the *pgoal* to achieve a *knowif* formula, it can adopt the *pgoal* to perform a *ynq* directed at someone whom it believes knows whether *Pred* is true. A yes-no question from speaker *S* to listener *L* whether predicate *P* is true can be decomposed as a sequence of the speaker's requesting that *L* do an *informif* action, followed by the *informif* action. That is:

*ynq(S, L, P) ≡ seq(request(S, L, informif(L, S, P)), informif(L, S, P))*

### 5.8 Requests

Requests are a paradigmatic example of a larger class of speech actions, the directives [76] that also includes commands, recommendations, suggestions, etc. The intended effect of the request is that the listener form the intention to do the requested action. From this class, Eva currently uses requests and recommendations, which differ in terms of whether the action being requested/recommended benefits the speaker or the listener, respectively. Notice that some of the parameters must be computed based on the embedded *Act.*

***request****(Speaker, Listener, Act)*

*constraint: bel(Speaker, Cond)*

*precondition: bel(Speaker, Pre),*

*effect: intend(Listener,do(action(Listener, Act, Cond),Loc, Time), Q)*

where *Pre* and *Cond* are, respectively, the precondition and the constraint of the requested *Act*, which benefits *Speaker*:

*precondition(Act, Pre),*

*constraint(Act, Cond),*

---

[39] The speech act literature, and works of ours [7] and colleagues [22] was concerned with precisely which formula became true after the speech action was executed. Notwithstanding our remarks above about the *Effect* not holding, we are having Eva assert that (1) *bel(Listener, pgoal(Speaker, Effect*)) become true. However, this might not be true if the speaker was found to be insincere or deceptive. Apart from reasoning about trust directly [7], we can also have Eva assert that *default(bel(Listener, pgoal(Speaker, Effect))).* That means that if Eva comes to believe that formula (1) is false, the default stipulation will block the conclusion that *Listener* in fact believes *Speaker* wanted the *Effect*. However, further detail is beyond the scope of this paper.

*benefits(Act,Speaker).*

### 5.9 Verification Questions

A number of application scenarios require that a user be verified by answering questions for which the system already has an answer in its database. The system's goal here is *not* for the system to come to know the answer, but for the system *to come to know what the user thinks is the answer.* This can be accomplished via the *assertref* action. Thus, in planning the verification question, the system *requests* this *assertref* action.[40] Notice that the effect of the *assertref* involves an existential quantifier whose scope is the *Listener's* belief of the *Speaker's* belief. We leave as an exercise for the reader to derive the preconditions and effects of the verification question from the preconditions and effects of the constituent actions.

***verifyref**(Speaker, Listener, Var^Pred) ≡*

*seq(request(Speaker, Listener, assertref(Listener, Speaker, Var^Pred)),*

*assertref(Listener, Speaker, Var^Pred) )*

## 6 Collaborative Planning

As discussed in Section 1, collaboration is so essential to society that we teach our children to be collaborative at a very early age [2]. However, present day conversational systems generally do not know how to be helpful or collaborate, stemming from their inability to infer and respond to the intention that motivated the conversant's utterance. To overcome this failing, we have built a collaborative planning-based system designed to assist its conversant(s) in achieving his/her goals. As previously mentioned, the approach dates back to work done at Bolt Beranek and Newman [10–12], at the University of Toronto [9, 13, 14], and at the University of Rochester (e.g., [15–19]). Such systems attempt to infer their conversants' plan that resulted in the communication, and then to ensure that the plans succeed. Indeed, as noted in Section 1, a central feature of our approach is that the system will attempt to infer as much of the user's plan as it can, will try to identify obstacles to its success, and plan to overcome those obstacles in order to help the user achieve his/her higher-level goals. Thus, plan recognition and planning are essential to Eva's architecture and processing. Below we provide a description of Eva's collaborative planning and plan recognition.

### 6.1 Planning Rules

Rather than representing a plan as a pure graph structure of actions, Eva's plans consist of a web of interdependent logical forms describing mental states, notably beliefs, persistent goals and intended actions (cf. [37]). Based on the epistemic planning approach first described in [8, 9, 13, 31, 52] and recast in the logic and speech act theory provided in Cohen & Levesque [1, 7], Eva's planning and plan recognition make extensive use of reasoning, as it derives and attributes new mental states of the user. The system generalizes the original plan-based dialogue research program by planning with multiple declaratively-specified mental states, including *persistent goal, intention, belief, knowif, knowref.*

The system has a hybrid planning algorithm [95] as it both engages in backward chaining from desired effect to one or more chosen actions that could achieve those effects, and decomposes hierarchically defined actions into more primitive ones as a hierarchical task network planner does [80, 96–98]. Others have described plan recognition in terms of "inverse planning" [99, 100]. Our planning algorithm is similar in some respects to that used by BDI systems in that the purpose is to determine the next action to perform, rather than to plan sequences of actions, though it can do that. One would not expect a dialogue planning system to engage in advance planning of back-and-forth interactions as in a game, unless the world were very constrained. We also are not interested in finding a machine-learned "optimal" response, given the rather arbitrary numerical weights/probabilities and the vast space of potential logical forms (not to mention natural language utterances) that a learning system driven by a user simulator [101] might generate. Rather, we want the system to do something appropriate and reasonable, as long as it has the means to recover from errors, respond to users' clarification questions, etc. Because the system interleaves planning, execution and so-called "execution-monitoring" that involves observing the user's actions in response to the system's (i.e., to have a dialogue), there are many opportunities to revector a dialogue towards success.

[40] A similar ***assertif*** action is used when the system wants to verify that the user knows whether a predicate is true.

Eva's planning involves the following rules, for which we use '=>' to separate antecedent and consequent. The formulas in the antecedent would be proven to be true via the meta-interpreter's *istrue* rules). The result of applying a planning rule is to assert the consequent. These assertions can be rewritten by the → rules.

***Effect-Action****:* If an agent *Agt* has *pgoal* to achieve a proposition *P*, and *Agt* can find an action *Act* that achieves *P* as an effect, then the planner creates a *pgoal* to do the action relative to the *pgoal* to bring about *P*.

*(P1) pgoal(Agt, P, Q)* and *effect(Agt, Act, P) => pgoal(Agt, done(Agt, Act), pgoal(Agt, P, Q))*

Given the definitions of intention provided earlier, the formula on the right side is the expansion of the intention to do *Act*:

*intend(Agt, Act, pgoal(Agt, P, Q))*

We will use intend formulas wherever possible.

If more than one action can be found, the planner creates a disjunctive action (see also Sadek et al., 1997).

***Act-Applicability Condition****:* If the planning agent believes an *Act's* applicability condition, *AC*, is false, the action is impossible, the intention is marked as *impossible.* During the main loop of the system, intentions that are impossible are retracted, as are any persistent goal or intentions that are relativized to it.

We adopt the following rule for applicability conditions: given action *A*, applicability condition *AC*, agent *Agt*, and a relativizing condition, *Q*:

*(P2) applicability_condition(A, AC)* and *bel(Agt, ~AC) =>*

*bel(Agt, impossible(done(Agt, A))), ~intend(Agt, A, Q)*

Recall that for a given action, the applicability conditions cannot be made true. The above rule means that, if *AC* is an applicability condition to do action *A*, and the agent believes it is false, the agent itself cannot possibly do anything to make *AC* true, so then the agent would drop (or not adopt) an intention to do *A*. I

If the planning agent *Agt* does not know whether *AC* holds, the following rule is used to create a *pgoal* to *knowif* that *AC* is true, relative to the intention to do the *Act.*

*(P3) intend(Agt, Act, Q)* and *applicability_condition(Act, AC)* and *\+knowif(Agt, AC) =>*

*pgoal(Agt, knowif (Agt, AC), intend(Agt, Act, Q))* and *blocked(intend(Agt, Act, Q))*

The created *pgoal* to *knowif* potentially leads the agent to ask a question. In addition, the persistent goal/intention to perform the *Act* is *blocked*, such that no more expansion of any plans passing through that action can be accomplished until the system knows whether the *AC* holds. Considering the system as agent, if the system comes to believe *AC* holds, the relevant blocked persistent goal/intention becomes unblocked and planning to achieve that pgoal/intention continues. Hereafter, we suppress the condition on all rules that the intention and/or persistent goal to do an action is not blocked.

***Act-Precondition****:* In backward-chaining, if the planner determines that a precondition (*PC*) to an intended *Act* is believed to be false, the planner creates a persistent goal to achieve PC, relative to the intention to perform Act.

*(P4) intend(Agt, Act, Q)* and *precondition(Act, PC), bel(Agt, ~PC) => pgoal(Agt, PC, intend(Agt, Act, Q))*

If the agent does not know whether or not *PC* holds, then it adopts the *pgoal* to *knowif PC* is true.

*(P5) intend(Agt, Act, Q)* and *precondition(Act, PC) and ~bel(Agt, PC) =>*

*pgoal(Agt, knowif(Agt, PC), intend(Agt, Act, Q))*

***Act-Body****:* This rule enables the planner to engage in hierarchical planning. When the planner creates an intention to perform an *Act* that has a decomposition (*Body*), it creates an intention to perform *Body* relative to the higher-level intention to perform *Act*. The intention to perform *Body* could then lead to planning with conditionals, disjunctions, and sequences.

*(P6)* *intend(Agt, Act, Q)* and *body(Act, Body) => intend(Agt, Body, intend(Agt, Act, Q))*

As discussed in Section 3.5.2, various expansions and relativizations are created between the *Body* action and the higher-level action. Note that the preconditions and effects of the higher-level action are derived from the structure of the *Body* action. In particular, the precondition of the higher-level act is the precondition of the first act in the decomposition. The effect of the higher-level act depends on the decomposition. For instance, the effect of a sequence is the effect of the last act in the sequence. The effects of the intermediate acts may or may not persist until the end, so we do not attempt to derive their status. Other rules are provided by the forward meta-interpreter which handles the assertion of intending complex actions in terms of intending its decomposed constituents.

***Act-Knowref:*** If an agent *Agt* has an intention to do an action *Act* (relative to *Q*), the agent has a *pgoal* to *knowref* the value for each of the obligatory arguments of that action relative to the intention that it does now know. If it does not know what the values are, it creates for each *Var*, a persistent goal to know what the *Var* is such that the agent intends to do the *Act* for which that *Var* is a parameter.

*(P7)* *intend(Agt, Act, Q)* and *unk_oblig_arg(Act, Role:Var#Type) =>*

*pgoal(Agt, knowref(Agt, Var^intend(Agt, Act, Q)), intend(Agt, Act, Q))*

The creation of such goals may lead to planning and executing wh-questions (so-called "slot-filling" questions) by the agent. Conversely, because of helpful goal adoption (Section 6.3.1) after the user asks a question indicating that the user has a *pgoal* to know the value of the variable in a predicate, the system may come to have a *pgoal* that the user know what that value is and may infer an action that the user wants to perform. Such a goal could lead the system to tell the user what s/he needs to know in order to do an inferred action, even if not explicitly asked. Also, if the user changes his/her mind about intending *Act*, the system can drop the *pgoal* to find out the values of *Act's* parameters.

### *Intended complex actions:*

If an agent intends to do a conditional action and does not know whether the condition holds, the agent forms a persistent goal to come to know whether the condition holds, relative to the intention to do the conditional action.

*(P8)* *intend(X, condit(P,A)) and \+knowif(X,P) => pgoal(X, knowif(X,P), intend(X, condit(P,A)))*

If an agent intends to do a conditional action, and believes the condition is true, then the agent intends to do the action relative to the intention to do the conditional action.

*(P9)* *intend(X, condit(P,A), Q)* and *bel(X,P)) => intend(X, A, intend(X, condit(P,A), Q))*[41]

Intending a mutually exclusive disjunctive action *disj(A,B)* results in two intentions: an intention to do action *A* provided action *B* has not been done, and similarly for *B*. So, whichever gets done first, will cause the other intention to be removed because the relativized intention for the disjunctive act has been achieved.

*(P10)* *intend(X, disj(A,B), Q) => intend(X, condit(~done(X,B), A), intend(X, disj(A,B), Q))* and

*intend(X, condit(~done(X,A), B), intend(X, disj(A,B), Q))*

An agent *X*'s intending to do the sequential action *seq(A,B)* results in two intentions: first in the agent *X's* intending to do the first action *A*, relative to the intention to do the sequence, and in *X*'s intending to do the second action when *done(X,A)* is true, again relative to the intention to do the sequence.[42]

---

[41] Note relativization of the consequent intention.

[42] In Cohen and Levesque [1], *done* meant the action has just finished. Because the Eva system maintains the time any action is done, it can predicate whether *done* is satisfied to whatever temporal degree it is needed. Indeed, it is generally unclear under what time constraint the subsequent action in any arbitrary sequence must be performed relative to the prior one. For example, if the first

*(P11)* *intend(X, seq(A,B), Q) => intend(X, A, intend(X, seq(A,B), Q))* and

*intend(X, condit(done(X,A),B), intend(X,seq(A,B), Q))*

## 6.2 Plan Recognition Rules

The system engages in plan recognition by expanding the user's plan, expressed in terms of appropriately relativized persistent goals and intentions, according to various rules similar to those of Allen and Perrault [9].

***Act-Effect****:* If the system has attributed to the user *Agt* a persistent goal/intention to perform an action *Act*, then assert that *Agt* has a persistent goal to achieve the effect *E* of *Act relative to the intention to do Act*.

*(P12)* *intend(Agt, Act, Q)* and *effect(Agt, Act, E) => pgoal(Agt, E, intend(Agt, Act, Q))*

***Precondition-Act:*** If the system has attributed to *Agt* a *pgoal(Agt, P, Q)* and *P* is the precondition to an act *Act*, then attribute to *Agt* the intention to do *Act* relative to that *pgoal*.

*(P13)* *pgoal(Agt, P, Q)* and *precondition(Agt, Act, P) => intend(Agt, Act, pgoal(Agt, P, Q))*

Note that *P* could enable multiple acts, e.g., *A1* and *A2*. The system would then attribute to *Agt* the intention:

*intend(Agt, disj(A1, A2), pgoal(Agt, P, Q)).*

***Know-if-exists****:* If *S* has attributed to *Agt* the *pgoal* to know whether or not there is an *X* such that *Pred*, where *Pred* is a schematic predicate variable that has a free variable *Var*, then attribute to *Agt* the *pgoal* to know what "the *P*" is. Formally,

*(P14)* *pgoal(Agt, knowif(Agt, ∃X^Pred ) Q) => pgoal(Agt, knowref(Agt, X^Pred) , R)* and

*Q = pgoal(U,knowref(Agt, X^Pred) , R)*

For example, if *Agt* wants to know whether there is a nearby vaccine center, then attribute to *Agt* the *pgoal* to know the referent of "nearby vaccine center". This would enable the ***Val-Action*** inference below.

***Val-Action****:* If *Agt* has a *pgoal(Agt, knowref(Agt, X#Type^Pred)),* and *X#Type* is a required argument in some action *Act* and *Act* has a constraint predicate *C*, then create a persistent goal to have done *Act* additionally constrained by *Pred*.

*(P15)* *pgoal(Agt, knowref(Agt, X#Type^Pred))* and *unk_oblig_arg(Agt, Act, Role:X#Type) =>*

*pgoal(Agt, knowref(Agt, X#Type^pgoal(Agt, and(done(Agt, Act), constraint(Agt, Act, Pred)), Q))*

For example, if *Agt* wants to know the location of the nearest vaccination center, then *Agt* may want to go to that location.[43]

***Knowif-Action:*** If *pgoal(Agt, knowif(Agt,P), Q)*, and *P* is an applicability condition for an *Act*, then attribute to the *Agt* the pgoal to have *done* that *Act* (i.e., the intention to do *Act).* Notice that because this is a plan recognition rule, the relativization argument of the *pgoal* to *knowif* is the intention to perform the *Act.* Formally:

*(P16)* *pgoal(Agt, knowif(Agt,AC), Q)* and *applicability_condition(Act, AC) => intend(Agt, Act, R)* and

*Q = intend(Agt, Act, R)*

---

action was a year ago, is doing the second action now satisfactory? Of course, that all depends on the actions and the agent's intentions.

[43] Of course, there may be many actions for which the variable *X#Type* is an argument. Rather than create a disjunction of all of them, we may assert that *Agt* wants to do ***some*** action for which the typed variable is an argument. Subsequent reasoning might then infer what is the best action to choose.

***Normal-Activity:*** If *Agt* has a *pgoal* to be located at a place *P*, then ascribe to *Agt* the *pgoal* to do the normal activity one does at location *P*. For example, if *Agt* has a *pgoal* to be located at a movie theater, then attribute to *Agt* the *pgoal* to watch a movie.[44]

*(P17)* *pgoal(Agt, location(Agt, Place), Q)* and *normal_activity(Place, Act) =>*

*intend(Agt, Act, pgoal(Agt, location(Agt, Place), Q))*

***Negative State*:** If *Agt* is in a negative state (of which there are a list of domain dependent types), infer that the agent wants to be in the corresponding positive state. For example, if the agent has lost her phone, infer that the agent wants to have found her phone. If the agent's phone has been damaged, infer that *Agt* wants her phone to be repaired.

*(P18)* *bel(Agt, state_of(Agt, NegState))* and *bel(Agt, positive_state(NegState, PosState)) =>*

*pgoal(Agt, state_of(Agt, PosState), Q)*

Finally, if the probability of an inferred intention of the user is below a modifiable threshold (which could be user-dependent), the system generates a goal to know whether the user is wanting/intending that intention. We see this in utterance 3 of the Sample Dialogue in Section 1.1.

## 6.3 Other Ways that Goals Arise

Eva is driven by its persistent goals, which results in its planning to achieve them, and/or helping its user to achieve his/her goals. We described above how many persistent goals are generated. Below we elaborate on other ways that *pgoals* arise.

### 6.3.1 Collaborative goal adoption

Because the system is collaborative, it will adopt as its own those goals that it attributes to the user. For example, if it believes the user wants to be vaccinated, it will adopt the goal that the user be vaccinated. However, such goal adoption is not absolute. Thus, if it believes the user or the system is not allowed to adopt the goal, it will not. More formally, if the system believes the user has *P* as a *pgoal*, and the system does not have a pgoal that *~P*, then the system adopts the *pgoal* that *P* relative to the user's *pgoal*:[45]

*(P19)* *pgoal(usr, P, Q)* and *\+pgoal(sys, ~P, R) => pgoal(sys, P, pgoal(usr, P, Q)) ).*

For example, if the system believes the user wants to *knowref P* (e.g., *P* = the secret formula for Coca Cola), and the system does not want the user not to know it, the system adopts the goal that the user *knowref P*. However, should the system not want the user to *knowref P*, then the system does not have to adopt the user's goal and plan to satisfy it. Notice also that if the system comes to believe that the user no longer wants *P*, then the system can abandon its *pgoal* that *P***,** which would then lead to its dropping any intentions it had created to achieving *P*.

Among the consequences of the theory of joint intentions [1, 4, 7] are communications that must occur when joint intentions are achieved or are impossible. Rather than implement the entire theory of joint intentions declaratively, we choose to incorporate some of its consequences as rules within Eva. Thus, we have:

*(P20)* *done(Agt, condit(bel(sys, bel(usr, pgoal(sys, P, pgoal(usr, P, Q))))),Act)* and *bel(sys,P))) =>*

*pgoal(sys, bel(usr,P), bel(sys,P))*

That is, if some action *Act* has been done prior to which the system believed that the user believed the system had a *pgoal* to achieve *P* relative to the user's wanting the system to do so, and after the act, the system comes to believe *P*, then the system then has a *pgoal* to get the user to believe *P* (relative to the system's believing *P*). Notice that in virtue of the definition of *pgoal*, the system no longer has the *pgoal* to achieve *P* because it believes *P* is true. For example, if the system offers to make an appointment, and the offer is accepted, then once the appointment is made, the system will inform the user of that fact. Likewise, if the user directly requests the system to make an appointment, and the system agrees, then the system will inform the user once the appointment has been made. In general, these should be *mutual* beliefs that the system has a pgoal dependent on the user, but we are refraining from incorporating that theoretical construct for the time being.

[44] There are, of course, many possible actions one might do at a place, with different prior probabilities [102]. Each of the pgoals has a probability argument, which would be affected by the prior. Again, one might consider inferring that U wants to do *something* at that location, rather than expand to every possible atomic action.

[45] One could be more specific here to require the system not to have a *pgoal* that *P* never be true. For now, this detail is irrelevant to our applications, but may well become so. Likewise, we could have a deontic operator for *S*'s not being permitted to achieve *P*.

Notice also that similar reasoning would apply if the system came to believe that *P* is impossible. That is,

*(P21)* *done(Agt, condit(bel(sys, bel(user, pgoal(sys, P, pgoal(usr, P, Q))))), Act)* and *bel(sys, impossible(P))))*

*=> pgoal(sys, bel(usr, impossible (P)), bel(sys, impossible (P)))*

In this case, the system will inform the user that *P* is impossible.

### 6.3.2 Generating goals to *knowif* by rule decomposition

Eva generates goals to know whether or not a proposition *P* is true in numerous ways. First, if *P* is a precondition to an intended action *A*, Eva will generate the goal *pgoal(sys, knowif(sys, P), intend(sys, A, Q))* (see Section 6.1**).** If Eva later learns that *P* is false, it may then attempt to make it true. If the intention to do *A* is dropped, the pgoal to *knowif(P)*, and anything depending on it, such as the likely intended Yes/No question, can be dropped as well. Second, if *P* is an applicability condition to an intended action *A*, Eva will attempt to prove that *knowif(sys, P)*. If it cannot prove it, Eva also generates the goal *pgoal(sys, knowif(sys, P), intend(sys, A, Q)).* In both cases, it *blocks* the intention to do *A* such that no further planning is done with that intention until it comes to *knowif(sys, P)***.** If it comes to believe the applicability condition is false, then *P* is impossible for it to achieve, so it retracts the intention and unwinds the plan subtree that depends on it. If it comes to believe a precondition is false, it will attempt to create a plan to achieve it.

Given a *pgoal(sys, knowifP)),* the system can plan a yes-no question (YNQ that *P*), provided it can find a listener *L* whom it can ask and of whom it believes *knowif(L, P).* This may involve asking someone other than the user (in the example, it is the pharmacy CVS).

Another special case of generating goals to *knowif(sys, P)* arises when *P* is defined in terms of a disjunction of Prolog clauses. For example, one might state a rule that a person is eligible for the Covid vaccine if:

*Clause1* ~ The person's age is greater than 65, or

*Clause2* ~ The person's age is between 50 and 64, and the person is caring for someone who is disabled, or

*Clause3* ~ The person's age is less than 50 and the person is an essential worker

If the system has a pgoal to know whether the user is eligible for a covid vaccine, i.e.,

*pgoal(sys, knowif(sys, eligible(User, covid_vaccine))),*

Eva also generates:

*pgoal(sys, knowif(sys, Clause1), pgoal(sys, knowif(sys, eligible(User, covid_vaccine)))) ,*

as well as pgoals to *knowif Clause2* and *Clause3*. Notice that these three pgoals are made relative to the eligibility pgoal; if any of the pgoals is achieved, eligibility becomes true, and the other pgoals are dropped.

## 7 Semantics of Slots

In the present plan-based dialogue system, slots are quantified-in goals, such as *the time you want to the system to make an appointment for a vaccination* (see also [13, 52]). Assuming the user wants the system to make an appointment, the system's goal adoption and planning would generate pgoals of knowing the date and the time for which the user wants to make an appointment.

Using the definition of *knowref* and the semantics in Appendix A, we can now provide a meaning to statements like:

*The system wants to know the date for which the user wants it to make an appointment for the user at some business b.*

This is represented as (showing an existential binding for other variables that we have so far been suppressing):

*pgoal(sys, knowref(sys, Day^pgoal(usr, ∃Time^done(sys, make_appointment(usr, b, Day, Time))), Q)*

Expanding *knowref* into its definition, we see that this formula essentially quantifies into two levels of modal operators – *bel* and *pgoal*, namely:

*pgoal(sys, ∃Day^bel(sys, pgoal(usr, ∃Time^done(sys, make_appointment(usr, b, Day, Time)), Q))*

or, in words:

*The system wants there to be a Day of which the system thinks the user wants there to be a time such that system makes an appointment for the user at business b on that day and time.*[46]

To make sense of such statements, consider that *bel(A, P)* means that *P* is true in all of agent *A*'s *B*-related possible worlds (see Appendix A). The meaning of *∃X^bel(A, p(X))* is that there is some value of *X* (call it *d*) assigned to it by the semantics of the "valuation" function *v* in the world in which the formula is evaluated, such that the same value assigned to X in *all* of *A*'s *B*-related worlds satisfies *p(X)*. Because the existential quantifier out-scopes the universal quantifier, the chosen value *d* is the same choice in every related world such that *p(d)*. As modal operators are embedded, the corresponding chains of *B* and *G*-relatedness continue, with *the same d* being chosen in all of them (see **Figure 7**).

Assume the system has a *pgoal* that there be a day on which the system thinks the user wants the system to make an appointment at a vaccine center. This would likely result in a question like *"When would you like me to make the appointment?"*. The user model contains a default assertion that the user knows what s/he wants, so by default the user knows when s/he wants to have appointments. However, the user might say "*I don't know*," or might say "*Mary knows the day*," or say, "*Ask Mary*". We adopt the Gricean heuristic that the user would have said the day if s/he knew it, and s/he didn't, so s/he doesn't know the day. The general default still holds, but a specific *neg(knowref(usr, Time^pgoal(…)))* would then be asserted, which would cause that default to be inapplicable. This would prevent the system from asking the same question again, as the precondition would no longer hold.

The system plans the wh-question when the effect of the speech act (**Figure 5**) matches the content of the system's *pgoal***,** provided that the system believes the precondition holds, i.e., the system believes that the user knows the referent of "*the time User wants to make an appointment*". If the system does not believe the user knows the referent, but knows of someone else who does, it could then plan a question to that other agent.

[46] There would likely be a similar *knowref* for choosing the time. Note also that Time2 is the date/time for the *making* of the appointment, not for the date/time of the appointment. Finally, the other missing arguments to the action are bound by existential quantifiers within the scope of the pgoal.

***System asks a Whq to User about the time User wants to make an appointment***

***"Slot"***

***Precondition: User knows the referent of the time User wants to make an appointment***

***Effect: System knows the referent of the time User wants to make an appointmen***

**Figure 5. Slot-filling question**

### 7.1 Handling Constraints on Slots

Every wh-question has a Variable and a Predicate which constrains its value. When the user conjoins another predicate, it further constrains that value. So, if the system wants to know the time that the user wants an appointment, the system has the following *pgoal*:

*pgoal(sys, knowref(sys, Time^pgoal(usr, done(sys, make_appointment(usr, b, Day, Time),* ***Cond****), R)), Q)*

When the user says: *"after 10 am",* the system then has (assuming the variable *Day* has already been given a value):

*pgoal(sys, knowref(sys, Time^pgoal(user, done(sys, make_appointment(usr, b, Day, Time),* ***and(Cond, after(Time, 10am)))****), R)), Q)*

Critically, as shown here, the *after* constraint needs to be *in the scop*e of the *user's pgoal***,** because it is not simply that the system believes the time is after 10am, but that the user *wants* it to be after 10am.[47] Another example of a constraint ("the earliest time available") can be found in the sample transcript.

In the course of this processing, Eva asserts that:

*neg(knowref(user, Time^pgoal(user, done([sys, make_appointment(usr, b, Day, Time), Cond), R)))*

That is, the user does not know what time s/he wants the system to make the appointment.

## 8 Operational Semantics – BDI architecture

Belief-desire-intention architectures have been researched for many years, beginning with Bratman et al. [75]. Inspired by philosophical and logical theories, reactive BDI architectures such as PRS [103] essentially sensed changes in the world that updated the system's state (its "beliefs"). The architecture determined which of its pre-packaged "plans" could be used to achieve its "goals" in that state. These architectures expand the plans hierarchically to decide what to do. Sardiña et al. [104] show that the inner loop of such BDI architectures is isomorphic to HTN (Hierarchical Task Network) planning. However, neither BDI architectures nor HTN-based systems engage in plan formation (e.g., [81], nor do they reason about other agents. Sardiña et al. [104] and de Silva et al. [105] present formal theories of how to incorporate plan formation in such architectures, in part using declarative statements of the system's goals. In our case, the Eva system needs to be more declarative still in order to reason about the *user's* beliefs, goals, and intentions. There is also some confusion in the BDI architecture literature in the use of the terms 'intention' and 'plan'. For other BDI architectures, intentions consist of plans, which are contained in a fixed plan library, corresponding to our actions and their hierarchical definitions. In contrast, Eva's plans consist of its intentions and goals, whose contents are actions (that are contained in a library) and propositions to be made true. Eva's plan is created at run-time by the hybrid planning algorithm, whereas other BDI architectures' plans are fixed and reactively executed. In virtue of the

---

[47] In addition to adding constraints, Eva can also revise the *pgoal*, including embedded values or constraints, such as would occur with "*how about 9 am?*"

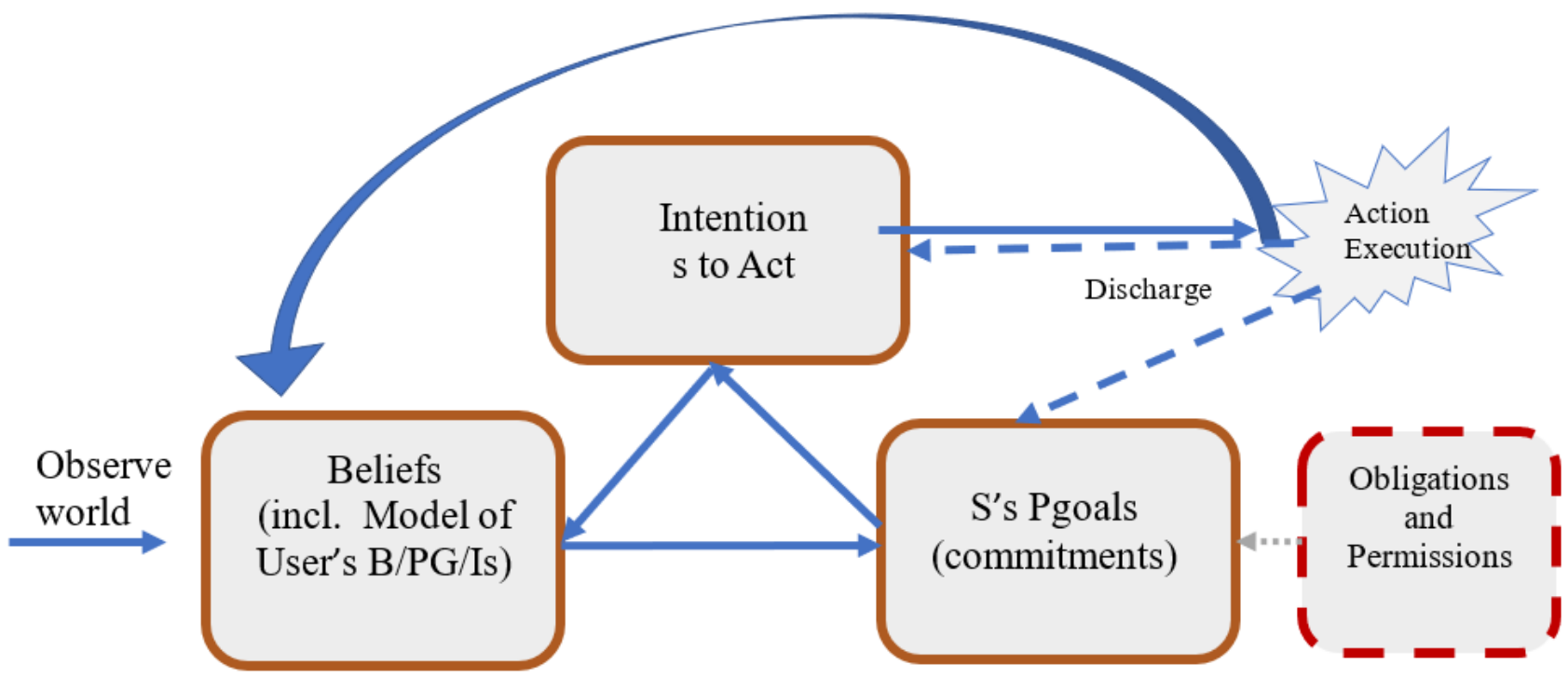

**Figure 6. The Basic BDI Architecture**

Cohen and Levesque [1] semantics, Eva's intentions also contain a relativization parameter, enabling it to unwind its plans appropriately.

The Eva system performs the basic loop described below, and depicted in **Figure 6**. It relies on the declarative representation and reasoning processes described above, as well as the intention formation and commitment processes in ([1]) that state the relationships among intentions, persistent goals, and beliefs.

## 8.1 Eva's Main Loop

Eva's operation could be described as looping through the following steps (where S identifies the system, and U stands for the user):

1) Observe the world, including U's action(s) $A_U$, including speech acts.
2) Assert that U wants (i.e., has *pgoal* for) the effect of $A_U$; If $A_U$ is a speech act, assert that U wants S to believe U wants the effect. If S trusts U, this will collapse "U wants S to believe U wants" into "U wants" the effect.[48]
3) Assert that U believes that the precondition of $A_U$ holds, and if S believes the action $A_U$ was successful, assert that S believes the precondition holds. [49]
4) Apply plan recognition rules until no new beliefs, persistent goals, or intentions are created.
5) Debug U's plan, i.e., check the applicability conditions for U's intention(s) in the plan.[50]
    i) If the applicability condition to an act is false, plan alternative ways to achieve the higher-level effect of the act.
    ii) Retract any intentions or pgoals that are marked as *not_applicable*, and remove their subtree of dependent pgoals/intentions. If there are no possible actions that can be performed to achieve the applicability condition, inform the user of the failure of that condition (e.g., no vaccination center has vaccine available).[51]
6) Adopt U's *pgoal*s to achieve *P* as the system's own *pgoals* -- i.e., *pgoal(U, P, Q)* → *pgoal(S, P, pgoal(U, P, Q))* if *P* does not conflict with system's existing pgoals.[52]

---

[48] However, if S has reason not to trust U, then S need not have the lesser embedded *pgoal*, This enables S to protect its innermost mental states from untrustworthy agents, which will alter its goal adoption and what it then plans to do.

[49] We are ignoring the temporal consideration that U believed the precondition held before the act.

[50] Shvo et al. [53] discuss how to find discrepancies in the inferred user plan based on what the system believes to be the user's incorrect beliefs (see also [37]).

[51] Eva does not retract the *not_applicable* predication on the act, thereby preventing the act from being planned again.

[52] Notice the relativization to the user's *pgoal*. If the system learns the *user* no longer has the *pgoal*, the system can drop its adopted *pgoal*.

7) For S's *pgoal* to achieve proposition *E*, S plans to achieve *E* by finding a (possibly complex) action $A_S$ that achieves it, resulting in an intention to perform action $A_S$. If $A_S$ benefits the user, S also creates a persistent goal to know whether the user would want S to perform $A_S$.[53]

8) If S does not know whether the applicability condition *AC* for $A_S$ is true, formulate a question to find out whether it is. Also, block the intention to perform $A_S$ until the truth or falsity of *AC* is believed.

9) Execute (some of the) intended act(s) $A_S$. We provide details regarding the choice of what intentions to execute below.

10) Remove the intention to perform $A_S$ if $A_S$ was *done.*

11) If $A_S$ is deemed impossible (e.g., the applicability condition for $A_S$ is believed to be false), unwind the intention to perform $A_S$ via the relativization conditions, continuing to unwind through the chain of pgoals and intentions that depend on $A_S$.

12) If $A_S$ terminates a higher-level act that S was *doing*, retract the *doing* predicate and assert that the higher-level act was *done*.

13) If the execution of $A_S$ fails (e.g., failure has been propagated to Eva from some backend system), remove the intention to achieve *As* and plan again to achieve the effect for which $A_S$ was planned. Inform the user of the failure of the action (e.g., the credit card could not be charged), and if the backend provides that information, the reason for the failure (e.g., card number is invalid).

These steps are repeated until no more formulas have been asserted.

## 8.2 Intention Satisfaction, Abandonment, and Revision

The logic of intention prescribes when intentions must be abandoned. For example, in Step 10, if an intended action has been performed, the intention is removed because it is satisfied. Likewise in Step 11, if the system discovers that an applicability condition for an intended action is false, then (because it can do nothing to achieve false applicability conditions), it concludes the intended action is impossible, and so abandons it and unwinds the subtree of pgoals/intentions that depends on the intention to perform the impossible action. For example, if the system forms the intention to achieve the conditional act, *condit(P, Act),* and it does not know whether *P* holds, it forms the *pgoal* to *knowif(S, P),* relative to the intention to do *condit(P, Act).* That *pgoal* to *knowif(S, P),* may well lead to an intention to ask a yes-no question as to whether *P* holds. Because of the chain of relativizations, if the intention to do the conditional action is abandoned, say because S no longer wants the effect of *Act* to which the conditional *Act* was relativized, the *pgoal* to achieve the *knowif(S, P)* will eventually also be dropped.

Intentions may also be abandoned by the system when users changes their mind about their goals. For instance, if the system has a persistent goal to achieve *P*, relative to the user's persistent goal that *P*, and the system comes to believe that the user no longer wants to achieve *P*, the system can drop its pgoal to achieve *P*, and unwind the plan subtree that depends on *P*. The plan subtree that depends on *P* possibly contains additional intentions and goals that have not yet been acted on, but were placed on the agenda in order to achieve *P*.

We note that in the scenario described in our example (Section 1.1), U has changed her mind *after* the system has already affected the environment to achieve U's previously specified goal (i.e., an appointment for Monday at 9 has been made). In our example, in order to achieve U's goal of not having an appointment on Monday at 9, the system offers to reschedule U's appointment, since the effect of the rescheduling action is both that U no longer has an appointment at the old date and time, and also that U has an appointment at a new date and time. Thus, the rescheduling action is an example of an action to 'undo' the consequences of S's having achieved U's old goal.

Regarding Step 13, we assume for now that the system can in fact execute any of its intentions for which the applicability condition and precondition are true at the time of execution. However, ability to execute an action does not necessarily translate into successful execution when the execution is delegated to some external agent. For example, assume Eva has gathered credit card information and sends the appropriate request to the credit card company to charge the card, only to receive a failure with a response that the card is over the credit limit. In this case the system's act *failed*. Eva would inform the user of that failure as an impossibility because it behaves according to the principles of joint intention theory [1], which entails informing the user when a joint intention is impossible to achieve. However, it then would replan to achieve the higher-goal of finishing the transaction by asking the user for another credit card (or some other plan). It would not retry that specific card because the intention to charge that card is impossible to achieve.

[53] We initialize the system with *benefits(Act, Agt)* predications for all actions it knows about. Actions can benefit the system, user, or both. For example, the *make_appointment* act benefits the user, not the system. So, if the system plans to do it, it needs to find out whether the user wants it to do so.

Icard et al. [106], Shoham [107], and Van der Hoek et al. [108], all consider the problem of intention revision with logics similar to or inspired by the one discussed here. The approach we are taking distinguishes itself in several ways. First, Eva relativizes intentions and goals to one another so that if a *pgoal* to achieve *P* is undermined because the intention/pgoal to which it is relativized has been given up, then that *pgoal* to achieve *P* may be given up as well. Second, if an intention is adopted relative to a belief, and for some reason that belief is retracted, so may the intention (e.g., "*I intend to take my umbrella because I believe it may rain*"). We cannot adopt Shoham's [107] use of a simple database of intentions, because we need to quantify into multiple levels of modal operators (see Appendix A.2). In both van der Hoek et al. [108] and Icard et al. [106], a distinction is made between beliefs that the agent adopts because it believes its intentions will be successful, and beliefs that occur from observing the world. Unlike the cited authors, Eva does *not* infer beliefs because of the effects of actions stated in the action description. Those descriptions are *not* dynamic logic expressions *per se*. The agent has a *pgoal* for the effects, but until the agent does the action, and determines that it was successful[54], it does not come to believe the effect. This simplifies joint intention and belief revision, as well as dialogue. Finally, as emphasized in van der Hoek et al. [108], there may be multiple reasons for forming an intention. They argue that the intention should only be retracted if all those reasons are themselves retracted. In Eva's case, the system might have two intentions to perform the same action *A*, each relativized to a different proposition. If one relativization is given up causing its relativized intention to be given up, the other intention to perform *A* that was relativized differently would still remain.

# 9 What Intentions to Execute

At any time, the system may have multiple intentions, including communicative intentions, that could be executed. For example, at the same time it could have a rapport-building action ("*that's terrible…*", "*I'm sorry that…*"), a confirmation, one or more informatives, a question to ask, and a request to pose. Some of these actions may change the "topic" of discussion, or begin work on a new goal, which can be introduced by discourse markers such as "*OK, now*", "*so*", etc. There may be mixed initiative digressions created by the user, such as seen in the example above when the user answers the question "*how old are you?*" with another question "*why do you ask?*", and with the conversation eventually being returned back to the original topic with a repetition of the prior speech act. Thus, there may be many possible choices for what action(s) to do next. Although one could imagine learning a policy for deciding among the actions to execute, the needed data for a new domain would be difficult to obtain, even with user simulators [46, 101]. As a result, we have implemented an initial policy in which Eva executes speech actions from its set of enabled intentions in the following order (with examples drawn from to the above dialogue):

1. Rapport-building (*"Sorry to have to ask again"*)
2. Confirmations (*"OK, you are eligible for the Covid vaccine"*)
3. Informatives on the same topic (*"The reason why …."*)
4. Repetition of a prior speech act in order to return to a prior goal/topic (*"but how old are you?"*)
5. Informatives on a new topic
6. A single directive or interrogative action, requiring a response from the user (*"Are you caring for someone who is disabled?", "What date would you like the appointment?"*). Again, continuing on the same topic is preferred to switching to a new one.

In a single turn Eva may execute multiple speech acts, but no more than one directive/interrogative.

Of course, how one defines "topic" for various speech acts and goals is difficult (cf. [19, 33]). We have provided Eva with an initial topic analysis based on the structure of the logical form at issue, but it is a matter of future research.

---

[54] How it does that is a different, but very interesting matter, especially for the intended effects of speech acts. See [7].

## 10 Use of Context

Eva maintains numerous types of context. First, at the logical form and linguistic levels, it maintains a chronologically-ordered *done* representation of all utterances and their speech acts, as well as domain actions, that have been performed by the dialogue participants. In addition, it keeps track of *pgoals* and *intentions* that it currently has, and also previously had but have been retracted. It also maintains the instantiated intended effects of both parties' speech acts. For example, the system's wh-question to the user "*how old are you?*" is represented as the act:

*whq(sys, usr, Age#years^age_of(usr, Age#years)),*

whose intended effect the system would maintain in the context database as:

*knowref(sys, Age#years^age_of(usr, Age#years)).*

This contextual information is used to derive a full logical form when the user answers a wh-question with a fragment, say "*45 years old*". In this case the parser generates the expression *45#years*, which is unified with the typed variable *Var#Term* of the contextual *knowref* formula above. This determines the predicate in question, enabling the Eva to identify the user's speech act as an *informref(user, sys, 45#years^age_of(usr, 45#years)).* Likewise, in processing answers to yes-no questions, Eva searches its context database for a *knowif(sys, Pred),* which, if successful, results in the user's speech act being identified as an *inform(user, sys, Pred)* or *inform(usr, sys, ~Pred).* Context also informs numerous other aspects of the system's processing (e.g., to generate anaphoric expressions[55]).

## 11 Explanation

A major advantage of planning-based systems is that they provide an immediate mechanism to support the generation of explanations [109]. Unlike black-box machine-learned systems (e.g., [46]), the present system has a plan behind everything that it says or does, such that it can answer questions like "*why did you say that?*". The explanation finds the path in the plan starting from the action being referred to, and follows the chain of achievements, enablements, and relativizations backwards to the intentions and persistent goals that led to the action to be explained. For example,

*S: "how old are you?"*

*U: "why do you ask?"*

*S: "The reason is that I need to determine whether you are eligible for the Covid vaccine"*

The *pgoals* needed to answer this request for explanation are described in Section 6.3.2. What would constitute a "good" explanation is a subject of considerable research [67, 110–112]. For example, in the dialog above, a good explanation would *not* be that the system asked for the user's age because it wanted to know the answer! Although that is the closest *pgoal* in the plan, that answer is likely something that the user already knows (S and U both are taken to know what the desired effect of a wh-question speech act is).[56] On the other hand, a good explanation is also not one where Eva asks the question because the user wants to be vaccinated, which is at the top of the plan tree. For the answer to the explanation request to be reasonable, Eva finds the lowest *pgoal* in the plan whose content the user does not already believe. In doing so, Eva needs to provide the explanation that the precondition for being vaccinated is eligibility.

The system also engages in proactive explanation by informing the user of preconditions of actions as goals the system needs to achieve. For example, in an insurance domain, it says: "*In order to help you file a claim for this incident, I need to have identified your phone*". Importantly, as with any action, the system checks whether the effect is already true, i.e., for an inform, whether the user already believes the propositional content. In particular, some of the preconditions are stated as being general common knowledge. For example, to have someone receive a vaccination at a vaccination center, the person needs to be located at the vaccination center. Because that proposition's being a precondition (as opposed to its truth) is common knowledge, the system does not plan an inform speech act of the need for this proposition to be achieved. But notice that it does recognize the user's plan to achieve her being at the vaccination center, and collaborates by telling the user what she would need to know, namely how to drive there. We leave more work on explanation to further research. However, we note that the overall planning-based framework inherently supports explanation since elements of the plan are causally linked to one another; this feature stands in stark contrast with much of the current AI research (conversational or otherwise) that relies on practically inscrutable models for generating answers or predictions.

---

[55] Natural language generation in Eva is beyond the scope of this paper.

[56] Kaptein et al. [113] show that adults prefer goal-based explanations to belief-based ones. However, their algorithm employs a fixed goal-tree for their domain of study. Their algorithm will find the goal that immediately dominates the action to be explained. In our planning system, this would generate obvious explanations, so we provide a higher-level goal that is not immediately obvious to the listener. Of course, one would need a theory of obviousness, but that must await future research.

# 12 Other Related Work

In addition to the references mentioned above, this section discusses three threads of research that strongly relate to the topics discussed here: epistemic planning, slot-filling dialogue systems, and plan-based dialogue systems.

## 12.1 Epistemic Planning

The planning community has considered planning coupled with sensing actions in order to overcome incomplete knowledge (e.g., [51, 66, 114, 115] ). Epistemic planning systems generate plans to influence agents' beliefs, both those of the planner and of other agents. We argue that an essential feature for any epistemic planner to support a dialogue system is that it be able to handle the incomplete belief/knowledge described by *knowref*, and be generalized to modal operators other than belief. However, most epistemic planning systems rely on the content of a belief (or multiply-embedded belief operator) being propositional (i.e., not first-order) and therefore do not support such reasoning. There are a few notable exceptions, including the work of Liberman et al. [116] who studied a first-order extension of dynamic epistemic logic (DEL) and appealed to term-modal logics to define the semantics for their language and corresponding epistemic planning system. Liberman et al.'s framework was applied to epistemic social network dynamics, which concerns the flow of information and knowledge through social networks, and how individuals' and groups' beliefs and behaviors are impacted (e.g., one might wish to model the spread of misinformation) [117]. Eva goes beyond Liberman et al.'s work by offering a complete implementation of a collaborative dialogue system. Importantly, in addition to belief or knowledge, Eva can reason with a number of modal operators essential to collaborative dialogue (e.g., persistent goal, intention) and can quantify into all of these modalities. Finally, Liberman et al. [116] offer a compact characterization of action schemas, inspired by PDDL, which bridges between research on planning formalisms and DEL. Future work could explore whether Liberman et al.'s action and domain representations could be used for our collaborative dialogue purposes.

The closest epistemic planning work to ours is the PKES planner of [66, 115, 118]. They developed a modal logic-based dialogue planning framework that uses one database to keep track of an agent's knowing whether a proposition is true, along with a different database for an agent's knowing the values of terms (cf. [52]). Driven by PKES, Petrick and Foster [119] developed an impressive human-robot social interaction system in a bartending domain. While the problems are similar to ones we tackle, the approach itself is limited by the use of a single level of databases, and its use only for the system's belief modality. In order to plan speech acts in dialogues, Eva's planning system needs to create plans to influence another agent's beliefs, (persistent) goals and intentions. Specifically, we showed here and in Cohen [49] the desirability of having quantifiers whose scope includes multiple modal operators. For example, the representation of *"John wants to know whether Mary knows the date that Sue wants to eat"*, which could lead to the question to Mary *"when does Sue want to eat?"*. To represent these would require a goal database within which is a know-if database, containing a know-value database within which is a goal database. Also, one could in fact quantify over agents to have: "*John wants to know <u>who</u> knows the secret*." Then it is not clear which database should be used. Eva currently reasons and creates plans with quantified-in formulas directly and without such databases.

## 12.2 Slot-Filling Dialogue Systems

The dialogue research community has concentrated in recent years on a form of task-oriented dialogue that emphasizes slot-filling, which dates back to the Gus frame-based dialogue system [120]. Wen et al. [46] say this about task/goal-oriented dialogue (emphasis ours):

> Given a user input utterance, $u_t$ at turn $t$ and a knowledge base (KB), the model needs to parse the input into actionable commands Q and access the KB to search for useful information in order to answer the query. Based on the search result, the model needs to summarise its findings and reply with an appropriate response $m_t$ in natural language.[57]

Building a system solely to execute actionable commands is a very limited conception of goal-oriented dialogue. For example, the original study of task-oriented dialogue from Grosz [121], had the *system* giving *the user* instructions on performing a task (assembling an air-compressor). Numerous researchers (e.g., [14, 19, 122]) have emphasized the need for *collaborative* tasks, in which both parties can get the other to perform actions, though typically they have not emphasized slot-filling per se. Eva is squarely in the camp of collaborative dialogue, for which slot-filling is a necessary component.

Although slot-filling is an important step, intent classification and slot-filling are only part of what it takes to engage in a task-oriented dialogue. The Dialogue State Tracking Challenge [56, 123] attempts to standardize a corpus-based test for systems that acquire values for slots. The most explicit definition of "slot" we can find is from Henderson [56] in describing the Dialog State Tracking Challenge (DSTC2/3):

> The slots and possible slot values of a slot-based dialog system specify its domain, i.e. the scope of what it can talk about and the tasks that it can help the user complete. The slots inform the set of possible actions the system can take, the possible semantics of the user utterances, and the possible dialog states… For each slot $s \in S$, the set of possible values for the slot is denoted *Vs*. ….
>
> The term *dialog state* loosely denotes *a full representation of what the user wants* at any point from the dialog system. The dialog state comprises all that is used when the system makes its decision about what to say next. … the dialog state at a given turn consists of:
>
> - The goal constraint for every informable slot $s \in S_{inf}$. This is an assignment of a value $v \in Vs$ that the user is specifying as a constraint, or a special value *Dontcare*, which means the user has no preference, or *None*, which means the user is yet to specify a valid goal for the slot.
> - A set of requested slots, the current list of slots that the user has asked the system to inform. This is a subset of $S_{req}$. …

Thus, slots are considered to be parameters of an action that either are filled by an atomic symbol, are unfilled, or are filled by the atoms *dontcare*, *dontknow*, or *none*. As we discussed in Cohen [49], the meaning of these values, or lack of values are unclear — are they quantified variables? Is there a negative somehow involved in *dontcare* and *dontknow*? If so, how are those embedded negatives used in reasoning? The notion of "informable" and "requestable" slots are intended to be approximations of modal operators and incomplete knowledge about the user's beliefs and desires, though without any accompanying semantics.

Overall, this intent+slot meaning representation is far too limiting, e.g., it does not handle true logical forms including Booleans, conditionals, superlatives, comparatives, temporal qualifications, etc. More difficult still, dialogues often provide constraints on the values for slots, rather than providing an atomic value, such as '*7 pm*'. Indeed, we showed in Cohen [49], examples in which the conversants *collaboratively* fill a slot, rather than just one party's doing so. A more general approach is needed (see Cohen [49] and the next section for further discussion).

## 12.3 Planning-Based Dialogue Systems

Although we have provided extensive references to planning-based dialogue systems throughout the paper, there are a number of important works with which to compare. The closest implementations have been the ARTIMIS system [39] and systems from Allen's group at the University of Rochester, (Trips, Trains, Cogent), and a recent plan-based dialogue system from IBM [124].

[57] In a footnote, the authors qualify their goal as: "Like most of the goal-oriented dialogue research, we focus on information seek type dialogues." Unfortunately, this strand of research is overly restrictive, and the limitations are deeply embedded in the mathematics of the approach.

First, there have been research systems that partially attain some plan-based capabilities. RavenClaw [125] employed a fixed hierarchical descriptions of dialogue moves, but did not engage in on-the-fly planning and reasoning. Trindikit [126] provides the dialogue system developer with generic tools to build rule-based dialogue systems using a simplified dialogue state and set of communicative actions. Although these systems take us part of the way to our desired type of plan-based system, they eschew representations of mental states of the participants, which Eva adopts and reasons with directly.

Beyond the early plan-based dialogue work at the University of Toronto, the first system to incorporate a variant of the Cohen and Levesque [1] logic was ARTIMIS [30, 39], which reasoned about beliefs and intentions for a deployed system that engaged users in spoken dialogue about finding services in the French Audiotel telephone network. The system engaged in mixed-initiative question-answering, and had to deal with substantial numbers of speech recognition errors. While very similar in spirit, the Eva system develops and makes more extensive use of the plan structure, collaboration (plan recognition, plan debugging, planning), goal/subgoal relativization, and explanation. The use of quantified modal operators in Eva is also more extensive, especially as applied to quantification through multiple belief/pgoal/intend operators.

Lemon et al. [127] developed the WITAS multimodal conversational system that drives an autonomous helicopter. WITAS bears some similarities to Eva in terms of its ability to reason about domain actions and maintaining multiple threads of conversation. The system is a derivative of the Information State approach to dialogue [126]. The Information State implementation that WITAS maintains include a/an: Activity Tree, Dialogue Move Tree (DMT), System Agenda, Pending List, Salience List, and Modality Buffer. The Activity Tree corresponds to Eva's hierarchical action expressions, but only seems to allow decompositions into sequential actions vs. Eva's use of sequential, conditional, and disjunctive actions. The actions are described as having preconditions and effects, but plans are not composed on-the-fly into future-directed plans. The actions can be obviated if the effects are found to be true in the system's database, but there is no mechanism to create goals to find out whether the preconditions are true if the system cannot prove those preconditions. Thus, unlike Eva, the robot system is essentially operating in a closed world. The Dialogue Move tree precomputes how user speech acts can "attach" to nodes in the DMT, though which nodes receives the attachment seems to depend solely on the dialogue act types rather than the effects of those acts (which are not specified). The generation of output employs the Gemini unification grammar [128] that renders logical forms as utterances. This is more general than Eva's natural language generator, which uses a structural decomposition of logical forms to generate simple utterances. WITAS' LFs are either on the system agenda or on the "pending list", which stores questions that the system has asked (but not those that the user has asked). Eva's pending list keeps track of questions from both parties, and also incorporates requestive types of speech acts, enabling it to more generally engage in multi-threaded interactions. WITAS' generation process may "aggregate" multiple clauses, rendering "*I will fly to the tower and I will land at the parking lot*" as "*I will fly to the tower and land at the parking lot*" (see also Appelt (1985) for similar plan optimizations). Finally, anaphoric expressions are generated if an object in a logical form is at the top of a "salience list". Eva generates an anaphoric expression for a logical form element if that element is present in the LF of the last dialogue turn (including both party's contributions). Apart from these differences, WITAS appears to do no plan recognition, nor obstacle detection or helpful behavior. Finally, its state representation is only of facts that the system believes to be true and goals that the system is pursuing, not a general-purpose representation of both system and user's mental states.

An interesting and useful point of comparison is the COGENT framework [19], the latest evolution of the TRIPS system developed by James Allen's team at University of Rochester, and, more recently, the Institute of Human and Machine Cognition (IHMC). Like Eva, COGENT is specifically addressing the need for dialogue systems to tackle much more complex tasks than what the current generation of conversational assistants are capable of [129]. Interestingly, COGENT evolved from the same ideas of using planning and plan recognition to drive dialogue [9], with speech acts serving as essential operators in such a plan [8, 9, 13]. Unlike Eva, however, the TRAINS/TRIPS/COGENT series of systems placed emphasis on natural language understanding, built around the TRIPS parser [129, 130], to create a very rich, domain-independent utterance meaning. The pursuit of domain independence led the group towards a theory of dialogue based on Collaborative Problem Solving (CPS), whereby the agent's competence in carrying out collaborative dialogue is completely separated from the competence to carry out domain actions. Thus, in COGENT the dialogue model proper is incorporated in the CPS Agent (CPSA), whereas domain-specific planning and execution are incorporated in a separate Behavioral Agent (BA). Dialogue in COGENT is reflected in updates to the CPS state, which keeps track of the status of joint intentions, and those updates are only happening after fairly extensive communication between the CPSA and the BA. In this model, a user's communicative act leads to a CPS act that change the CPS state, which in turn leads to a change in the problem solving state of the BA; generation of communicative acts by the system follows this path backwards.[58] This division of labor between managing the dialogue itself and the generic linguistic aspects of the interaction on one hand, and managing the individual problem-solving aspects of the system on the other hand is meant to make the COGENT framework attractive to potential developers of collaborative dialogue systems without requiring them to have sophisticated linguistic expertise. However, they would still need to master the complex logical form language embedded in the CPS acts to create mappings to and from their internal representations. And, when utterances in their chosen domain fail to parse, they are out of luck.[59]

By contrast, Eva uses a ML-based semantic parser, which may generate a less-detailed LF, but is much easier to train and improve. Although Eva currently does not have as explicit a model of joint intention as COGENT does, its dialogue model does, in fact, ensure that the system's and user's plans mesh appropriately in true collaborative fashion. Eva has a more explicit model of the participants' beliefs and intentions, which provides it with a solid basis for implementing additional reasoning about joint intentions and shared knowledge, if need be (cf. [131]). In fact, COGENT lacks a deep model of user's beliefs, relying on the BA to do so. CPSA's dialogue model does not depend on having a "theory of mind", which is a potentially severe limitation. For example, COGENT cannot model a multi-user conversation, in part because it cannot distinguish the different users' mental states. We think these mental states are crucial, so much so that they are at the core of Eva's planning-based dialogue model. Whereas Eva has the ability to explain its own behavior, COGENT's current model has the ability to justify answers, and proposed modifications or failures of the CPS acts, but in its current incarnation it cannot answer a question such as "*why do you ask?*" [60]

Finally, we consider the recent work of Muise et al. [124], which we will refer to as PGODS (based on the paper title). This work is very much on the same track as ours, but concentrates on the planning aspect rather than the dialogue per se. PGODS attempts to provide a high-level specification methodology for simple dialogue systems centered around a restricted form of planning. For PGODS, the system developer specifies dialogue actions and representations of back-end actions, as well as system responses for each, from which a planner essentially compiles a large tree of all possible actions/utterances the system and end user can take. Based on this, the authors nicely show how a planning-based approach is more compact than having to specify each step of a scripted dialogue.[61] Prominent among approaches to dialogue from which PGODS leverages is the classical "intent+slots" approach, which we have discussed in Section 12.2. Unlike those approaches, the PGODS approach allows one to specify more general types of dialogues, but still restricted in their complexity.

A few specific comments will serve to differentiate the PGODS approach from ours. First, the FOND (Fully Observable Non-Deterministic) planner employed [132] uses the PDDL language [83] for expressing preconditions and effects, which only allows atomic symbols to be expressed, rather than the modal logic expressions that Eva uses. This very much restricts the logical forms that the system can handle and the kind of reasoning it can support. PGODS (and PDDL) does not distinguish between preconditions and applicability conditions, such that the former can be made true, but the latter cannot. Likewise, Eva's use of hierarchical action descriptions is more expressive than PDDL (and the hierarchical variant, HDDL). The planning formalism used by PGODS encodes all of the possible outcomes of an action in that action's definitions, which would need to be similarly encoded for every action. For example, all the possibilities that might arise when an action is deemed to be impossible are folded into that action's PDDL representation. For all such conditions, PGODS assumes that the developer is specifying the system's response utterances by hand. In Eva, these are technically not part of the plan but result from its execution and are handled by the BDI architecture, which itself depends on the semantics of intention (specifically, the conditions for giving up an intention).

A very substantial difference between the PGODS system and Eva is the representation of incomplete knowledge about the user's mental states. In particular, the PGODS system appeals to the well-known 0-approximation [133] and represents knowledge and uncertainty via a simple 3-valued logic that supports the use of efficient planning tools.

---

[58] This is an over-simplification, but for the purpose of this necessarily superficial description it will do.

[59] For example, we tested the initial sentence in our initial dialogue, "*Are there any Covid vaccination centers nearby?*" on a version of the TRIPS parser (http://trips.ihmc.us/parser/cgi/drum-dev) and the resulting parse has an incorrect interpretation for "*nearby*", which can be problematic for generating the correct response from a suitable BA.

[60] However, we do believe COGENT's model can be extended to handle such questions.

[61] The plan-based dialogue community knew this from the start, but since scripted dialogues were not considered to be a major research advance, the argument was never made explicitly.

In contrast, Eva's modal logic framework offers a much richer way of encoding (multiple) agents' knowledge as well as a myriad of other mental states such as beliefs, goals, and intentions.

Finally, while not its focus, the PGODS system does not handle multi-agent settings, plan recognition, obstacle detection, goal adoption, and collaboration. There are many more differences, but suffice it to say that Eva operates at a more theoretically-grounded level.

Recent systems [134, 135] have parsed utterances into "data flow" graphs that provide a graph query or retrieval "plan" of execution or operations on that graph. Data flows from one query action to another is exactly analogous to shared logical variables in a unification framework. Indeed, the parsing of utterances into logical forms, dating back at least to the Chat system of Warren and Pereira [136] as executed by a Prolog-based interpreter essentially provides "data flow" during execution. The representation of meta-operations on the dataflow graph, such as referring to an entity, or replacing one entity or subset of logical form elements with another, can be handled by conjoining additional constraints, and by replacing predicates in logical forms with others. More generally, the above-cited works do not discuss domain or speech action planning, plan recognition, nor collaboration and thus do not provide a framework for collaborative planning-based task-oriented dialogue.

# 13 Limitations

This initial system has many limitations. First, it incorporates algorithms for which some researchers have surely built superior renditions for isolated situations. However, they have not typically been adapted for dialogue. For example, there are better probabilistic planners and plan recognizers that have been developed for academically interesting problems. Yet their limitations may preclude their use here if they have not been adapted for multi-agent interaction, or for reasoning with adequate representations of mental states. Of course, it goes without saying that such a system will eventually need to deal with uncertainty. We have left room for probabilistic reasoning as well as utilities with regard to planning and plan recognition. Likewise, the belief operator takes a probability argument, and reasoning could in principle take advantage of it. However, Eva often has to deal with embedded beliefs and goals. It is still an open question how probabilistic reasoning would incorporate those embedded operators. Still, we believe the basic structure of this system can function quite well until researchers have further developed probabilistic multi-agent reasoning for dialogue, which would then require a massive data collection effort in order to incorporate reasonable probabilities. Our contribution here is to situate the problems and initial approaches in the context of a useful cooperative dialogue system. Other examples of algorithms incorporated here that could use improvement are the modal logic reasoner (e.g., to deal better with equality, negation, defaults, uncertainty, and causality), semantic parsing of natural language that incorporates anaphoric expressions, failed presuppositions, multi-utterance speech acts, multi-act utterances, time and tense, etc. Likewise, the language generation component can be greatly improved to deal with a variety of syntactic constructions: grounding (cf. [54, 137]), and confirmations [4, 138], generation of anaphoric expressions [139], and of discourse markers. Most importantly, a tool needs to be created that enables developers to provide domain actions and business processes that will drive the planning and reasoning. Overall, our purpose here is to put all the pieces together in a fashion that enables researchers to see where their specialty fits in the larger scheme.

# 14 Concluding Remarks

The Eva dialogue system engages its users in cooperative planning-based task-oriented dialogues by inferring the user's plans, and planning to facilitate them. It formalizes the planning-based approach to speech acts [8–10, 13, 52] with the analysis of intention and speech acts in [1, 7]. It adapts that approach with principles derived from existing theories of collaboration [4, 6] in order to provide a principled analysis and reasoning process underlying conversational interaction. Whereas there have been many research works that have investigated aspects of this general approach to dialogue, none recently have been done within a *declarative* BDI reasoning and planning framework. It has been supposed by many that this approach is too computationally intensive to function as a real-time conversational agent. Eva is a counterexample to that supposition, as it engages in domain-dependent real-time mixed-initiative collaborative dialogues, using multimodal inputs and outputs. The conversational capabilities include its planning of a variety of speech acts, context-dependence, constraint processing in response to slot-filling questions, mixed initiative, over-answering by system and user, multi-agent dialogue based on models of the mental states of more than one agent. The system's conversational capabilities are domain independent as applied to domain dependent actions and knowledge. But, how should we or a system gather the knowledge needed to incorporate a new domain (e.g., banking, automotive service)? We are so far agnostic on that issue – perhaps systems can read domain-specific documents providing the domain constraints (if such documents exist). We are also investigating easy-to-use knowledge building tools to impart specific domain knowledge. Rules coupled with domain knowledge can be written at high levels of abstraction (such as what our BDI planner uses), covering many cases, which can overcome challenges faced by the dialogue community in learning all dialogue behavior from data.

Unlike many current dialogue systems, we do not adopt an end-to-end trained approach to task-oriented dialogue because we believe it misses significant generalities and requires systems to relearn how to engage in dialogue for each domain. Moreover, they typically will use pre-trained language models that have difficulty telling the truth, and that are so far incapable of explaining their responses in a way that is causally-connected to what it said. We advocate a domain-independent model of dialogue processing that is applied to domain-dependent knowledge. One reason one might advocate for the end-to-end approach is to argue that it copes with the variability so often found in dialogue. However, we believe there is much more variability in the natural language processing *per se* than there is in the dialogue processing, so that training of the natural language engines (as we do) can proceed independently of building the dialogue processing engine.

Moreover, on the topic of language models, numerous assertions in the literature tout the emergence of reasoning capabilities in pretrained Large Language Models (LLMs), which are large neural network models built upon the Transformer architecture that leverage the power of extensive training data [140, 141]. However, recent work has cast doubt on the capacity of LLMs to effectively engage in planning (e.g., [142]), even in classical planning settings with deterministic actions and full observability. In contrast, our system operates in much richer settings involving multiple agents and partial observability where planning must take into consideration agents' beliefs, goals, and intentions. In addition to the seeming failures of LLMs in relatively simple planning settings, and despite claims that Theory of Mind capabilities have emerged in LLMs [143, 144], recent work has demonstrated how these reasoning skills falter in the face of even minor modifications to Theory of Mind tasks (e.g., [145, 146]). Thus, as stated earlier, these models, in their current form, cannot be entrusted with carrying out the kind of collaborative dialogue envisioned in this paper in an end-to-end fashion.

However, in addition to documenting the failures of LLMs, recent research has also endeavoured to create neuro-symbolic systems that synthesize the powerful pattern matching capabilities of LLMs with the provably correct reasoning capabilities and inherent interpretability and explainability of symbolic AI techniques (e.g., [147−150]). Our neuro-symbolic system combines a BDI-based collaborative dialogue system with neural network-based natural language understanding and (in ongoing work) generation techniques. The aforementioned enhancement of planning via LLMs offers a potential path forward for further neuro-symbolic integration within our system. To this end, we are currently investigating various instantiations of such integration with LLMs within the planning and dialogue components of our system.

Lastly, there is always a request among academics for competitive evaluation. We believe such evaluation is premature here as the purpose of this paper is to show that a more *declarative* specification of a collaborative plan-based BDI dialogue management engine can drive a complete task-oriented dialogue system in real-time. On the other hand, there have been challenges from the symbolic AI community to the effect that 1) the logic of intention with which we defined speech acts [1, 7] was too complex to be functional and useful, and 2) that anything "logical" is fruitless (except for writing academic papers). We propose this paper and system as a kind of existence proof in having built a dialogue system based on a well-understood analysis of intention that: knows what it is doing, can explain its behavior, and plans to communicate its "mental" state (along with performing domain tasks, of course) in order to influence those of its users.

### 14.1 Ethical Considerations

Finally, in the not-so-distant future, with better analyses of the system's and users' mental states, such dialogue systems will need to be handled with care. Unscrupulous system developers could build competent but nefarious dialogue agents that could plan to take advantage of unsuspecting people, maximizing their developers' utility. With deep fake technology and detailed user models, they could be instruments of crime and social unrest. The first steps in dealing with this issue is for a system to be able to decide whether it trusts a given user, and if not, to refrain from concluding that the user wants the typical utterance effect to hold (cf. [29]). Rather it could infer something weaker, e.g., that the system believes the user *wants* the system to *think* the effect holds [7, 22]. From there, the system could continue the dialogue, but not perform requested actions, could send out appropriate warnings, terminate the conversation, etc. [151, 152]. In general, the AI community will need to incorporate technologies into *defensive dialogue systems* whose purpose is to shield users and corporations from those who would do them harm.

## 15 Acknowledgements

Many thanks to James Allen, David McGee, Reza Haffari, Zhuang Li, Eli Pincus, Sebastian Sardiña, Candy Sidner, and Raj Tumuluri for their valuable comments and suggestions.

# 16 Bios

*Dr. Phil Cohen,* Chief Scientist Emeritus at Openstream.ai®, has long engaged in research in the fields of human-computer dialogue, multimodal interaction, and multiagent systems. He is a Fellow of the Association for Computing Machinery, Fellow of the Association for the Advancement of Artificial Intelligence, Fellow and Past President of the Association for Computational Linguistics, and the recipient of the 2017 Sustained Accomplishment Award from the International Conference on Multimodal Interaction. He is a co-inventor of plan-based dialogue systems, and of the Open Agent Architecture at SRI International, which led to Apple's Siri digital assistant. Cohen is one of the two winners of the Inaugural Influential Paper Award from the International Foundation for Autonomous Agents and Multi-Agent Systems for his paper with Hector Levesque "Intention is Choice with Commitment", *Artificial Intelligence 42(2-3),* 1990. He is also an Adjunct Professor of Data Science and Artificial Intelligence, in the Faculty of Information Technology, Monash University, Melbourne, Australia.

*Dr. Lucian Galescu* is the Director of Dialogue Research at Openstream.ai®, where he focuses on planning-based conversational AI. Earlier, he was a research scientist with the Florida Institute for Human and Machine Cognition. Dr. Galescu's research has spanned a broad spectrum of topics in dialogue systems, natural language understanding, knowledge extraction, intention recognition, and statistical language modeling. He has a number of widely cited publications in conversational AI and holds two patents in related technologies.

*Dr. Maayan Shvo* is a Research Scientist at Openstream.ai®, where he focuses on planning-based conversational AI. Earlier, as part of his doctoral research at the University of Toronto, he focused on epistemic planning and Theory of Mind reasoning and their role in a variety of important social reasoning tasks such as explanation, plan recognition, and assistance.

## 17 Bibliography

1. Cohen, P. R., & Levesque, H. J. (1990). Intention is choice with commitment. *Artificial Intelligence*, *42*(2–3), 213–261.
2. Warneken, F., & Tomasello, M. (2006). Altruistic helping in human infants and young chimpanzees. *Science*, *311*(5765), 1301–130. https://doi.org/10.1126/science.1121448.
3. Clark, H. H., & Wilkes-Gibbs, D. (1986). Referring as a collaborative process. *Cognition*, *22*(1), 1–39.
4. Cohen, P. R., & Levesque, H. J. (1991). Teamwork. Nous.
5. Grosz, B. J., & Kraus, S. (1996). Collaborative plans for complex group action. *Artificial Intelligence*, *86*(2), 269–357.
6. Grosz, B. J., & Sidner, C. (1990). Plans for discourse. In P. R., J. Cohen, Morgan, M. E., & Pollack (Eds.), *Intentions in Communication*. Cambridge, MA: MIT Press.
7. Cohen, P. R., & Levesque, H. J. (1990). *Rational Interaction as the Basis for Communication*. Intentions in Communication.
8. Cohen, P. R., & Perrault, C. R. (1979). Elements of a plan-based theory of speech acts. *Cognitive Science*, *3*(3).
9. Perrault, C. R., & Allen, J. F. (1980). A plan-based analysis of indirect speech acts. *Computational Linguistics*, *6*, 3-4 ,.
10. Bruce, B. (1975). Case systems for natural language. *Artificial Intelligence*, *6*(4), 327–360.
11. Sidner, C., Bates, M., Bobrow, R., Brachman, R., Cohen, P., Israel, D., … Woods, W. (1981). *Research in knowledge representation for natural language understanding*. Bolt, Beranek and Newman Inc.

12. Brachman, R. J., Bobrow, R., Cohen, P., Klovstad, J., Webber, B., & Woods, W. (1979). *Research in natural language understanding*. National Technical Information Service.
13. Perrault, C. R., Allen, J. F., & Cohen, P. R. (1978). Speech acts as a basis for dialogue coherence. In D. Waltz (Ed.), *Proc. of Theoretical Issues in Natural Language Processing (TINLAP-2)* (pp. 125–132).
14. Allen, J. F. (1979). *A plan-based approach to speech act recognition* (PhD Thesis). Dept. of Computer Science, University of Toronto.
15. Allen, J. F., Schubert, L. K., Ferguson, G. H., Hwang, P., H., C., Kato, T., … R, D. (1995). The TRAINS project: A case study in building a conversational planning agent. *Journal of Experimental and Theoretical Artificial Intelligence*.
16. Allen, J., & Ferguson, G. (2002). Human-Machine Collaborative Planning. In *Proceedings of the Third International NASA Workshop on Planning and Scheduling for Space*. Houston, TX.
17. Geib, C., Weerasinghe, J., Matskevich, S. K., P., C., B., & Petrick, P. A. (2016). Building helpful virtual agents using plan recognition and planning. In *Proc. of the 12th AAAI Conference on Artificial Intelligence and Interactive Digital Entertainment (AIIDE-16)* (pp. 162–168).
18. Ferguson, G., & Allen, J. (2011). A Cognitive Model for Collaborative Agents. In *Proceedings of the AAAI 2011 Fall Symposium on Advances in Cognitive Systems*. Washington, DC.
19. Galescu, L., Teng, C. M., F., A. J., & Pereira, I. (2018). Cogent: A Generic Dialogue System Shell Based on a Collaborative Problem Solving Model. *Proceedings of SigDial*, 400–409.
20. Freedman, R., Levine, S., Williams, B., & Zilberstein, S. (2017). Helpfulness as a Key Metric of Human-Robot Collaboration. In *Artificial Intelligence for Human-Robot Interaction: Trust & Explainability in Artificial Intelligence for Human-Robot Interaction, AAAI Fall Symposium Series*.
21. Shvo, M., & McIlraith, S. A. (2020). Active goal recognition. *Proceedings of the AAAI Conference on Artificial Intelligence*, *34*(06), 9957–9966.
22. Perrault, C. R. (1990). *An application of default logic to speech act theory, Intentions in Communication*. (P. R. Cohen, J. Morgan, & M. E. Pollack, Eds.). MIT Press.
23. Kumar, V., Maheshwari, B., & Kumar, U. (2002). Enterprise resource planning systems adoption process: a survey of Canadian organizations. *International Journal of Production Research*, *40*(3), 509–523.
24. Jennings, N. R., & Wooldridge, M. (1995). Applying agent technology. *Applied Artificial Intelligence an International Journal*, *9*(4), 357–369.
25. Tambe, M. (1997). Towards flexible teamwork. *Journal of artificial intelligence research*, *7*, 83–124.
26. Grice, H. P. (1957). Meaning. *The Philosophical Review*, *66*, 377–88.
27. Austin, J. (1962). Speech acts. Oxford.
28. Cohen, P. R., Levesque, H. J., & Persistent. (1987). Intention, and Commitment. In M. P. Georgeff & A. L. Lansky (Eds.), *Proceedings of the 1986 Workshop on Reasoning about Actions and Plans* (pp. 297–340). Los Altos, CA: Morgan Kaufmann.
29. Traum, D. R., Swartout, W., Gratch, J., & Marsella, S. (2008). A virtual human dialogue model for non-team interaction. In *Recent Trends in Discourse and Dialogue* (pp. 45–67). Springer.
30. Bretier, P., & Sadek, D. (1996). A rational agent as the kernel of a cooperative spoken dialogue system: Implementing a logical theory of interaction. In J. P. Müller, M. J. Wooldridge, & N. R. Jennings (Eds.), *Intelligent Agents III Agent Theories, Architectures, and Languages. ATAL 1996. Lecture Notes in Computer Science* (Vol. 1193). Berlin, Heidelberg: Springer. https://doi.org/10.1007/BFb0013586
31. Cohen, P. R., Perrault, C. R., & Allen, J. F. (1982). Beyond question-answering. In W. Lehnert, M. Ringle, & L. E. Associates (Eds.), *Strategies for Natural Language Processing*.
32. Ferguson, G., & Allen, J. (2007). Mixed-Initiative Dialogue Systems for Collaborative Problem-Solving. *AI Magazine*.
33. Grosz, B. J., & Sidner, C. (1986). Attention, intentions, and the structure of discourse. *Computational Linguistics*, *12*(3), 175–204.
34. Litman, D. J., & Allen, J. F. (1987). A Plan Recognition Model for Subdialogues in Conversations. *Cognitive Science*, *11*, 163–200.
35. McRoy, S., & Hirst, G. (1995). The Repair of Speech Act Misunderstandings by Abductive Inference. *Computational Linguistics*, *21*(4), 435–478.
36. McShane, M., English, J., & Nierenburg, S. (2021). Knowledge engineering in the long game of Artificial Intelligence: The case of speech acts. *Proc. of Advances in Cognitive Systems*, *9*.
37. Pollack, M. E. (1990). Plans as complex mental attitudes. In P. R. Cohen, M. E. Pollack, & J. Morgan (Eds.), *Intentions in Communication* (pp. 77–104). Cambridge: MIT Press.
38. Power, R. (1979). The organisation of purposeful dialogues. *Linguistics*, *17*, 107–152.

39. Sadek, D., Bretier, P., & Panaget, F. (1997). ARTIMIS: Natural dialogue meets rational agency. In *Proc. IJCAI-15* (pp. 1030–1035).
40. Sidner, C. L. (1985). Plan parsing for intended response recognition in discourse. *Computational Intelligence, 1*(1). https://doi.org/10.1111/j.1467-8640.1985.tb00054.x
41. Traum, D. R., & Allen, J. F. (1994). *Discourse Obligations in Dialogue Processing*. Proc. Association for Computational Linguistics.
42. Carberry, M. S. (1985). *Pragmatic modeling in information system interfaces (goals, dialogue, plans, ill-formedness)*. University of Delaware.
43. Koller, A., & Stone, M. (2007). Sentence generation as a planning problem. In J. Carroll, A. van den Bosch, & A. Zaenen (Eds.), *ACL 2007, Proceedings of the 45th Annual Meeting of the Association for Computational Linguistics, June 23-30, 2007, Prague, Czech Republic*. The Association for Computational Linguistics.
44. Lesh, N., Rich, C., & Sidner, C. L. (1999). Using plan recognition in human-computer collaboration. In *UM99 User Modeling: Proceedings of the Seventh International Conference* (pp. 23–32). Springer.
45. Poesio, M., & Traum, D. R. (1997). Conversational actions and discourse situations. *Computational intelligence*, *13*(3), 309–347.
46. Wen, T. H., Y., M., Blunsom, P., & Young, S. (2017). Latent Intention Dialogue Models. In *Proceedings of the 34th International Conference on Machine Learning*. Sydney, Australia, PMLR 70.
47. Mrkšić, N., Séaghdha Ó., D., Tsung-Hsien, W., Blaise, T., & Steve J., Y. (2017). Neural Belief Tracker: Data-Driven Dialogue State Tracking. In *Proceedings of the 55th Annual Meeting of the Association for Computational Linguistics* (pp. 1777–1788).
48. Chu-Carroll, J., & Carberry, S. (1994). A plan-based model for response generation in collaborative task-oriented dialogues. In *Proceedings of the Twelfth AAAI National Conference on Artificial Intelligence* (pp. 799–805).
49. Cohen, P. R. (2019). Foundations of task-oriented dialogue: What's in a slot? In *Proceedings of the 20th SigDial meeting*. Stroudsburg: PA:Association. for Computational Linguistics.
50. Teixeira, M. S., & Dragoni, M. (2022). A review of plan-based approaches to dialogue management. *Cognitive Computation*, *14*, 1019–1038.
51. van Der Hoek, W., & Wooldridge, M. (2003). Towards a logic of rational agency. *Logic Journal of the IGPL, 11(2),* 135–160.
52. Cohen, P. R. (1978). *On knowing what to say: Planning speech acts* (PhD Thesis). Dept. of Computer Science, University of Toronto.
53. Shvo, M., Hari, R., O'Reilly, Z., S., A., Wang, S. Y. N., McIlraith, S. A., & A, S. (2022). Proactive Robotic Assistance via Theory of Mind. In *International Conference on Intelligent Robots and Systems (IROS 2022)*. Kyoto, Japan.
54. Traum, D., & Allen, J. F. (1992). A "speech acts" approach to grounding in conversation. In *2nd International Conference on Spoken Language Processing (ICSLP 1992)*.
55. Gašić, M., Mrkšić, N., Rojas-Barahona, L. M., Su, P.-H., Ultes, S., Vandyke, D., … Young, S. (2016). Dialogue manager domain adaptation using Gaussian process reinforcement learning. *Computer Speech and Language*, *45,* 552–569.
56. Henderson, M. (2015). Machine learning for dialog state tracking: A review. In *Proceedings of The First International Workshop on Machine Learning in Spoken Language Processing*.
57. Shah, P., Hakkani-Tür, D., Tür, G., Rastogi, A., Bapna, A., Nayak, N., & Heck, L. (2018). Building a conversational agent overnight with dialogue self-play. *arXiv preprint arXiv:1801.04871*.
58. Wang, Y., Berant, J., & Liang, P. (2015). Building a semantic parser overnight. In *Proc. of Assoc. for Comp* (pp. 1332–1342). Ling.
59. Li, Z. (2022). *Machine translation and active learning for multilingual semantic parsing* (Chapter 6, Semantic Parsing in Limited Resource Conditions, PhD Thesis). Faculty of Information Technology, Monash University.
60. Gratch, J., & Marsella, S. (2004). A domain-independent framework for modeling emotion. *Cognitive Systems Research*, *5*, 269–306.
61. Kripke, S. A. (1963). Semantical Considerations on Modal Logic. *Acta Philosophica Fennica*, *16,* 83–94.
62. Hintikka, J. (1962). *Knowledge and Belief: An Introduction to the Logic of the Two Notions*. Ithaca, NY: Cornell University Press.
63. Rao, A. S., & Georgeff, M. P. (1991). Modelling rational agents within a BDI- architecture. In *Proc. 2nd International Conference on Principles of Knowledge Representation*. AAAI Press.

64. Fagin, R., Halpern, J. Y., Moses, Y., & Vardi, M. Y. (2003). *Reasoning About Knowledge*. Cambridge, MA, USA: MIT Press.
65. Rendsvig, R., & Symons, J. (2019). Epistemic logic.
66. Steedman, M., & Petrick, R. (2007). Planning dialogue actions. In *Proc. of SigDial*.
67. Miller, T., Felli, P., Muise, C., Pearce, A. R., & Sonenberg, L. (2017). Knowing whether. In *Proper Epistemic Knowledge Bases, Proc. of AAAI*.
68. Barcan, R. C. (1946). A Functional Calculus of First Order Based on Strict Implication. *Journal of Symbolic Logic*, *11*.
69. Quine, W. V. O. (1956). Quantifiers and propositional attitudes. *Journal of Philosophy*, *53*(5), 177–187.
70. Kaplan, D. (1968). Quantifying in. *Synthese*, *19*(1/2), 178–214.
71. Appelt, D. (1985). *Planning English Sentences*. Cambridge, UK: Cambridge University Press.
72. Appelt, D. K. & A. (1987). A Computational Model of Referring. In *Proc. of the International Joint Conference on Artificial Intelligence*.
73. Moore, R. C. (1977). Reasoning about knowledge and action. In *Proc. of the International Joint Conference on Artificial Intelligence*.
74. Anscombe, G. E. M. (1957). *Intention*. Oxford: Basil Blackwell.
75. Bratman, M. I. (1987). *Plans, and Practical Reason*. Cambridge, MA: Harvard University Press.
76. Searle, J. R. (1969). *Speech acts: An essay in the philosophy of language* (Vol. 626). Cambridge university press.
77. Dennett, D. C. (1971). Intentional Systems. *The Journal of Philosophy*, *68*(4).
78. Dennett, D. C. (1987). *The Intentional Stance*. Cambridge: MIT Press.
79. Poole, D. (1988). A logical framework for default reasoning. *Artificial intelligence*, *36*(1), 198827–47.
80. Sacerdoti, E. D. (1977). *A Structure for Plan and Behavior*. Elsevier-North Holland.
81. Bordini, R., & Hübner, J. F. (2005). BDI Agent Programming in AgentSpeak Using Jason. *Computational Logic in Multi-Agent Systems: 6th International Workshop, CLIMA VI*, *6*, 143–164.
82. Cohen, P. R., & Levesque, H. J. (1997). *Communicative Actions for Artificial Agents, Software Agents*. (J. Bradshaw, Ed.). AAAI Press.
83. McDermott, D., Ghallab, M., Howe, A., Knoblock, C., Ram, A., Veloso, M., … Wilkins, D. (1998). *PDDL—The Planning Domain Definition Language* (Technical Report CVC TR98003/DCS TR1165.). New Haven, CT: Yale Center for Computational Vision and Control.
84. Höller, D., Behnke, G., Bercher, P., Biundo, S., H., F., Pellier, D., & R, A. (2019). HDDL—A Language to Describe Hierarchical Planning Problems. In *Proc. of International Workshop on HTN Planning (ICAPS*.
85. Bercher, P., Keen, S., & Biundo, S. (2014). Hybrid planning heuristics based on task decomposition graphs. In *Proceedings of the International Symposium on Combinatorial Search* (Vol. 5, pp. 35–43).
86. Harel, D., Kozen, D., & Tiuryn, J. (2000). *Dynamic logic*. Cambridge: MIT.
87. Gutierrez, J., Kraus, S., Perelli, G., & Wooldridge, M. (2022). Giving Instructions in Linear Temporal Logic. In R. Posenato & S. Tonetta (Eds.), *Proc. of 29th International Symposium on Temporal Representation and Reasoning (Time 2022), Alexander Artikis* (Vol. 15, pp. 1–15). Germany: Dagstuhl Publishing.
88. Rao, A., & Georgeff, M. (1995). BDI-agents: From theory to practice. In *Proceedings of the First International Conference on Multiagent Systems*.
89. Herzig, A., & Longin, D. (2004). C&L intention revisited. In *Proceedings of the 9th International Conference on Knowledge Representation and Reasoning* (pp. 527–535). AAAI Press.
90. Rao, A., & Georgeff, M. (1998). Decision procedures for BDI logics. *Journal of Logic and Computation*, *8*(3).
91. Frege, G. (1952). On sense and reference (in the Translations from the Philosophical Writings of Gottlob Frege).
92. Mitkov, R. (1999). *Anaphora resolution: the state of the art*. Wolverhampton, UK: School of Languages and European Studies, University of Wolverhampton.
93. Mitkov, R. (2014). *Anaphora resolution*. Routledge.
94. Sukthanker, R. (2020). Anaphora and coreference resolution: A review. *Information Fusion*, *59*, 139–162.
95. Sardiña, S., & Padgham, L. (2011). A BDI agent programming language with failure handling, declarative goals, and planning. *Autonomous Agents and Multi-Agent Systems*, *23*, 18–70. https://doi.org/10.1007/s10458-010-9130-9
96. Geib, C., & Goldman, R. (2011). Recognizing plans with loops represented in a lexicalized grammar. In *Proc. of the 25th Conference on Artificial Intelligence (AAAI)* (pp. 958–963). AAAI Press.
97. Nau, D. S., Au, T. C., Ilghami, O., Kuter, U., Murdock, J. W., Wu, D., & Yaman, F. (2003). SHOP2: An HTN planning system. *Journal of Artificial Intelligence Research*, *20*, 379–404.

98. Wilkins, D. E. (2014). *Practical planning: extending the classical AI planning paradigm*. Elsevier.
99. Ramírez, M., & H, G. (2009). Plan Recognition as Planning. In *Proceedings of the International Joint Conference on Artificial Intelligence*.
100. Baker, C. L., Saxe, R., & Tenenbaum, J. B. (2009). Action understanding as inverse planning. *Cognition*, *13*(3), 329–349.
101. Schatzmann, J., Thomson, B., Weilhammer, K., Ye, H., & Young, S. (2007). Agenda-based user simulation for bootstrapping a POMDP dialogue system. In *Proc. of NAACL-HLT*.
102. Jiang, T., & Riloff, E. (2018). Learning prototypical goal activities for locations. In *Proc. of Assoc. for Comp* (pp. 1297–1307). Ling.
103. Georgeff, M. P., & Lansky, A. L. (1987). Reactive reasoning and planning. In *Proc. of AAAI-87* (pp. 677–682).
104. Sardiña, S., Silva, L., & Padgham, L. (2006). Hierarchical planning in BDI agent programming languages: A formal approach. In *Proceedings of the Fifth International Joint Conference on Autonomous Agents and Multiagent Systems* (pp. 1001–1008).
105. De Silva, L., Sardina, S., & Padgham, L. (2009). First principles planning in BDI systems. In *Proceedings of The 8th International Conference on Autonomous Agents and Multiagent Systems* (pp. 1105–1112). International Foundation for Autonomous Agents and Multiagent Systems (IFAAMAS).
106. Icard, T., Pacuit, E., & Shoham, Y. (2010). Joint revision of belief and intention. In *Proceedings of the Twelfth International Conference on Principles of Knowledge Representation and Reasoning (KR 2010* (pp. 163–200). Palo Alto: AAAI Press.
107. Shoham, Y. (2009). Logical Theories of Intention and the Database Perspective. *Journal of Philsophical Logic DOI*. https://doi.org/10.1007/s10992-009-9116-8
108. van Der Hoek, W., Jamroga, W., & Wooldridge, M. (2007). Towards a theory of intention revision. *Synthese*, *155*, 265–290.
109. Winograd, T. (1972). Understanding natural language. *Cognitive Psychology*, *3*(1), 191,. https://doi.org/10.1016/0010-0285(72)90002-3.
110. Shvo, M., Klassen Q., T., & McIlraith, S. A. (2020). Towards the role of theory of mind in explanation. In D. Calvaresi, A. Najjar, M. Winikoff, & K. Framling (Eds.), *Explainable, Transparent Autonomous Agents and Multi-Agent Systems* (pp. 75–93). Springer International Publishing.
111. Mueller, S. T., Hoffman, R. R., Clancey, W., Emrey, A., & Klein, G. (2019, February). Explanation in Human-AI Systems: A Literature Meta-Review Synopsis of Key Ideas and Publications and Bibliography for Explainable AI, DARPA XAI Program.
112. Swartout, W. R., & Moore, J. D. (1993). *Explanation in second generation expert systems, Second Generation Expert Systems*. Berlin: Springer.
113. Kaptein, F., Broekens, J., Hindriks, K., & Neerincx, M. (2017). Personalised Self-Explanation by Robots: The Role of Goals versus Beliefs in Robot-Action Explanation for Children and Adults. In *Proc. of ROMAN 2017*. https://doi.org/10.1109/ROMAN.2017.8172376
114. Bolander, T., & Andersen, M. B. (2011). Epistemic planning for single- and multi-agent systems. *Journal of Applied Non-Classical Logics*, *21*(1), 9–34.
115. Petrick, R. P., & Bacchus, F. (2002). A Knowledge-Based Approach to Planning with Incomplete Information and Sensing. In *AIPS* (Vol. 2, pp. 212–222).
116. Liberman, A. O., Achen, A., & Rendsvig, R. K. (2020). Dynamic term-modal logics for first-order epistemic planning. *Artificial Intelligence*, *286*, 103305.
117. Liberman, A. O., & Rendsvig, R. K. (2019). Dynamic Term-Modal Logic for Epistemic Social Network Dynamics. In *International Workshop on Logic, Rationality and Interaction* (pp. 168--182).
118. Bacchus, F., & Petrick, R. (1998). Modeling an agent's incomplete knowledge during planning and execution. In *PRINCIPLES OF KNOWLEDGE REPRESENTATION AND REASONING-INTERNATIONAL CONFERENCE-* (pp. 432–443). Morgan Kaufmann Publishers.
119. Petrick, R. P. A., & Foster, M. E. (2013). Planning for social interaction in a robot bartender domain. In *International Conference on Automated Planning and Scheduling (ICAPS)* (pp. 389–397).
120. Bobrow, D. G., Kaplan, R. M., Kay, M., Norman, D. A., Thompson, H., & Winograd, T. (1977). GUS, a frame-driven dialog system. *Artificial Intelligence*, *8*(2), 155–173.
121. Grosz, B. J. (1977). *The representation and use of focus in dialogue understanding.* University of California, Berkeley.
122. Rich, C., & Sidner, S. L. (1997). COLLAGEN: When agents collaborate with people. In M. Huhns & M. Singh (Eds.), *Readings in Agents*. San Francisco: Morgan Kaufmann Publishers.

123. Williams, J. D., Henderson, M., Raux, A., Thomson, B., Black, A., & Ramachandran, D. (2014). The dialog state tracking challenge series. *AI Magazine*, *35*(4), 121–124.
124. Muise, C. M., Chakraborti, T., Agarwal, S., Bajgar, O., Chaudhary, A., Lastras-Montaño, L. A., … Wiecha, C. (2019). Planning for goal-oriented dialogue systems. https://doi.org/10.48550/arXiv.1910.08137.
125. Bohus, D., & Rudnicky, A. I. (2009). The RavenClaw dialogue management framework. *Computer Speech and Language*, *23*, 332–361.
126. Larsson, S., & Traum, D. R. (2000). Information state and dialogue management in the TRINDI Dialogue Move Engine Toolkit. *Natural Language Engineering*, *6*(3–4), 323–340.
127. Lemon, O., Bracy, A., Gruenstein, A., & Peters, S. (2001). The WITAS multi-modal dialogue system I. In *Seventh European Conference on Speech Communication and Technology*.
128. Shieber, S., Noord, G., Pereira, F. C. N., & Moore, R. C. (1990). Semantic head-driven generation. *Computational Linguistics*, *16*(1), 30–42.
129. Allen, J., Galescu, L., Teng, C. M., & Perera, I. (2020). Conversational Agents for Complex Collaborative Tasks. *AI Magazine*, *41*(4), 54–78. https://doi.org/10.1609/aimag.v41i4.7384.
130. Allen, J., Dzikovska, M. O., Manshadi, M., & Swift, M. (2007). Deep linguistic processing for spoken dialogue systems. In */ACL 2007 Workshop on Deep Linguistic Processing* (pp. 49–56).
131. Subramanian, R. A., Kumar, S., & Cohen, P. R. (2006). Integrating joint intention theory, belief reasoning, and communicative action for generating team-oriented dialogue. In *Proceedings of the National Conference on Artificial Intelligence* (Vol. 21, p. 1501). Menlo Park, CA; Cambridge, MA; London; AAAI Press; MIT Press; 1999.
132. Muise, C., McIlraith, S., & Beck, C. (2012). Improved non-deterministic planning by exploiting state relevance. In *Proceedings of the International Conference on Automated Planning and Scheduling* (Vol. 22, pp. 172–180).
133. Baral, C., & T.C, S. (1997). Approximate Reasoning about Actions in Presence of Sensing and Incomplete Information. In *ILPS* (Vol. 97, pp. 387–401).
134. Andreas, J., Bufe, J., Burkett, D., Chen, C., Clausman, J., Crawford, J., … Zotov, A. (2020). Task-oriented dialogue as dataflow synthesis. In *Transactions of the Association for Computational Linguistics* (Vol. 8, pp. 556–571).
135. Tellman, Z. (2023). Designing a framework for conversational interfaces. *Communications of the ACM, 66*(10), 44–49.
136. Warren, D. H. D., & Pereira, F. C. N. (1982). An Efficient Easily Adaptable System for Interpreting Natural Language Queries. *American Journal of Computational Linguistics*, *8*(3–4), 110–122.
137. Geib, C., George, D., Khalid, B., Magnotti, R., & Stone, M. (2022). An integrated architecture for common ground in collaboration. *Advances in Cognitive Systems*, *10*, 1–20.
138. McGee, D. R., & Cohen, P. R. (2001). Creating tangible interfaces by augmenting physical objects with multimodal language. In *Proceedings of the 6th international conference on Intelligent user interfaces* (pp. 113–119).
139. McCoy, K. F., & Strube, M. (1999). Generating anaphoric expressions: pronoun or definite description? In *Proc. Workshop on the Relation of Discourse Structure and Reference*. Association for Computational Linguistics.
140. Vaswani, A. (2017). Attention is all you need. *Advances in neural information processing systems*, *30*.
141. Brown, T., Mann, B., Ryder, N., Subbiah, M., Kaplan, J. D., Dhariwal, P., … Agarwal, S. (2020). Language models are few-shot learners. *Advances in neural information processing systems*, *33*, 1877–1901.
142. Valmeekam, K., Sreedharan, S., Marquez, M., Olmo, A., & Kambhampati, S. (2023). On the planning abilities of large language models (a critical investigation with a proposed benchmark. *arXiv preprint arXiv:2302.06706*.
143. Kosinski, M. (2023). Theory of mind may have spontaneously emerged in large language models. *arXiv preprint arXiv:2302.02083*.
144. Bubeck, S., Chandrasekaran, V., Eldan, R., Gehrke, J., Horvitz, E., Kamar, E., & Zhang, Y. (2023). Sparks of artificial general intelligence: Early experiments with GPT-4. *arXiv preprint arXiv:2303.12712*.
145. Sap, M., LeBras, R., Fried, D., & Choi, Y. (2022). Neural theory-of-mind? on the limits of social intelligence in large LMs. *arXiv preprint arXiv:2210.13312*.
146. Ullman, T. (2023). Large language models fail on trivial alterations to theory-of-mind tasks. *arXiv preprint arXiv:2302.08399*.

147. Silver, T., Hariprasad, V., Shuttleworth, R. S., Kumar, N., Lozano-Pérez, T., & Kaelbling, L. P. (2022). PDDL Planning with Pretrained Large Language Models. *NeurIPS 2022 Foundation Models for Decision Making Workshop*.
148. Guan, L., Valmeekam, K., Sreedharan, S., & Kambhampati, S. (2023). Leveraging Pre-trained Large Language Models to Construct and Utilize World Models for Model-based Task Planning. *arXiv preprint arXiv:2305.14909*.
149. Silver, T., Dan, S., Srinivas, K., Tenenbaum, J. B., Kaelbling, L. P., & Katz, M. (2023). Generalized Planning in PDDL Domains with Pretrained Large Language Models. *arXiv preprint arXiv:2305.11014*.
150. Chu-Carroll, J., Beck, A., Burnham, G., Melville, D. O., Nachman, D., Özcan, A. E., & Ferrucci, D. (2024). Beyond LLMs: Advancing the Landscape of Complex Reasoning. *arXiv preprint arXiv:2402.08064*.
151. Dalton, A., Aghaei, E., Al-Shaer, E., Bhatia, A., Castillo, E., Cheng, Z., … Dorr, B. (2020). Active defense against social engineering: The case for human language technology. In *Proceedings for the First International Workshop on Social Threats in Online Conversations: Understanding and Management* (pp. 1–8).
152. Sarkadi, S., Panisson, A. R., Bordini, R. H., McBurney, P., Parsons, S., & M, C. (2019). Modelling deception using theory of mind in multi-agent systems. *AI Communications Journal*.

## Appendix A:  Semantics for the Logic

Taken from Cohen and Levesque [1].

### A.1 Syntax

The syntax of the logical language follows.  Semantics for the logical language can be found in Section A.2.

<Action-var> ::=A, B, $a_1$, $a_2$. . . . . $b_1$, $b_2$. . . . . e, $e_1$, $e_2$. . . . .
<Agent-var> ::= x, y, $x_1$, $x_2$. . . . . $y_1$, $y_2$. . . . .
<Regular-var> ::= i, j, $i_1$, $i_2$ . . . . . $j_1$,$j_2$. . . . .
<Variable> :: = <Agent-var> | <Action-var> | < Regular-var>

<functor> : an atomic function symbol

<Pred-symbol>: an atomic predicate symbol

<Time-term> ::= <date(Mo, Day,Year)>

<Term> ::= <functor>($Arg_1$, $Arg_2$, … $Arg_n$)> | <Time-term> | <Variable>

<Pred >:: = <Pred-symbol> (<Variable>1. . . . . <Variable>n), a Boolean predicate.

<Wff> ::= < Pred>  | ~<Wff > | <Wff> ∨ <Wff >| ∃ <variable > <Wff>  |  impossible(<Wff>),  <Term> = <Term>, a well-formed (Boolean) formula.

<Act_expr> ::=  <Action-var>

| <action_name>([agent:Agent#agent, … Role:Var#Type])

| seq(<Act_expr>1 ; <Act_expr>2): sequential action[62]

| disj(<Act_expr>$_1$ | <Act_expr>2):  nondeterministic mutually-exclusive disjunctive action

| condit(< Wff >, <Act_expr>): conditional action

<Action-expression> :: = action(agent:Agent#agent, act:<Act_expr>, constraint:<Wff>)

Linear time operators:

◊<Wff>:    <Wff>  "Eventually" —<Wff> is true at some time in the future.

□<Wff>  $=_{def}$ ~◊~<Wff>    "Always" — <Wff>  is true in all future states of the world.

before(<$Wff_1$>,  <$Wff_2$>) :  at any time,  if <$Wff_2$> is true, then <$Wff_1$> was true before then.

do([<Action-expression>, location:Loc#location, time:Time#time]):  <Action-expression> will be done at location Loc and time Time in the future.[63]

done([<Action-expression>, location:Loc#location, time:Time#time]): <Action-expression> has been done at location Loc and a past time Time.

---

[62] Extension to lists of sequential or disjunctive actions is straightforward.

[63] Note that **do** is actually redundant given ◊ and **done.**

doing([<Action-expression>, location:Loc#location, time:Time#time]): <Action-expression> is ongoing at location Loc and time Time.

## A.2 Model-Theoretic Semantics

A precise model theory and semantics for the logical language can be found in Cohen and Levesque [1], section 3.2. Informally, we quote from that paper (p. 224):

> We shall adapt the usual possible-worlds model for belief to deal with goals. Assume there is a set of possible worlds T, each one consisting of a sequence (or course) or events, temporally extended infinitely in past and future. Each possible world characterizes possible ways the world could have been, and could be. Thus, each world specifies what happens in the future. Agents usually do not know precisely which world they are in. Instead, some of the worlds in T are consistent with the agent's beliefs, and some with her goals, where the consistency is specified in the usual way, by means of an accessibility relation on tuples of worlds, agents, and an index, n, into the course of events defining the world (from which one can compute a time point, if need be). To consider what an agent believes (or has as a goal), one needs to supply a world and an index into the course of events defining that world. As the world evolves, agents' beliefs and goals change. When an agent does an action in some world, he does not bring about a new world, though he can alter the facts of that world at that time. Instead, after an event has happened, we shall say the world is in a new "state" in which new facts hold and the set of accessible worlds has been altered. That is, the agent changes the way he thinks the world could be and / or chooses the world to be.[64]

The point of having a formal semantics is that it gives meaning to and constrains the statements that the system manipulates. For example, as compared with so-called "slot-filling" systems, we now can say precisely what a slot is and how slot-filling questions and answers can be provided. As we have shown, the system plans questions when it is reasoning about what it needs to know. See Section 6.1 for more details.

### 3.2.1. *Model theory*

A model M is a structure $<\Theta, P, E, Agt, T, B, G, \Phi>$, where $\Theta$ is a set, $P$ is a set of people, $E$ is a set of primitive event types, $Agt \in [E \rightarrow P]$ specifies the agent of an event, $T \subseteq [\mathbb{Z} \rightarrow E]$ is a set of possible courses of events (or worlds) specified as a function from integers to elements of $E$, $B \subseteq T \times P \times \mathbb{Z} \times T$ is the belief accessibility relation, $G \subseteq T \times P \times \mathbb{Z} \times T$ is the goal accessibility relation, and $\Phi$ interprets predicates. Formulas will be evaluated with respect to some possible course of events, hereafter some *possible world,* and an "index" into that possible world, that is, at a particular point in the course of events.

### 3.2.2. *Definitions*

(1) $\Phi = \Theta \cup P \cup E^*$, specifying the domain of quantification. That is, one can quantify over things, people, and sequences of (types of) primitive events. Given this, $\Phi \subseteq [Pred^k \times T \times \mathbb{Z} \times D^k]$.

(2) $AGT \subseteq E^* \times P$, where $x \in AGT[e_1 \ldots\ldots e_n,]$ iff there is an i, such that $x = Agt(e_i)$. That is, AGT specifies the partial set of agents of a sequence of events.

### 3.2.3. *Satisfaction (Cohen and Levesque [1], p. 225)*

Assume $M$ is a model, $\sigma$ is a sequence of events, $n$, an integer, $v$, a set of bindings of variables to objects in the domain of quantification $D$, and if $v \in [Vars \rightarrow D]$, then $v(X,d)$ is that function which yields $d$ for and is the same as $v$ everywhere else. We now specify what it means for $M,\sigma,v,n$ to *satisfy* a wff $\alpha$, which we write as $M,\sigma,v,n \models \alpha$. Because of formulas involving actions, this definition depends on what it means for an expression a to occur between index points $n$ and $m$. This, we write as $M,\sigma,v,n[[a]]m$, and is itself defined in terms of satisfaction. The definitions are as follows:

---

[64] Many researchers (e.g., [63]) have suggested using a branching time model rather than the linear time one adopted here. For the purposes of building a dialogue system, these differences are of minor importance.

(1) $M,\sigma,v,n \models P(x_1. \dots . x_k)$ iff $<v(x_1) \dots v(x_k)> \in \phi[P, \sigma, n]$. Notice that the interpretation of predicates depends on the world $\sigma$ and the event index n.
(2) $M,\sigma,v,n \models \sim \alpha$ iff $M,\sigma,v,n$ does not satisfy $\alpha$.
(3) $M,\sigma,v,n \models \alpha \vee \beta$ iff $M,\sigma,v,n \models \alpha$ or $M,\sigma,v,n \models \beta$.
(4) $M,\sigma,v,n \models \exists x\, \alpha$ iff $M,\sigma,v(x,d),n \models \alpha$ for some $d$ in $D$.
(5) $M,\sigma,v,n \models (x_1 = x_2)$ iff $v(x_1) = v(x_2)$.
(6) $M,\sigma,v,n \models$ <Time-proposition> iff v (<Time-proposition>) = $n$.
(7) $M,\sigma,v,n \models$ impossible($\alpha$) iff $\Box\sim \boldsymbol{\alpha}$

The interested reader should refer to Cohen and Levesque [1], section 3.2.3].

Next, we provide a semantics for statements about beliefs and goals.

(1) $M,\sigma,v,n \models bel(x, \alpha)$ iff for all $\sigma^*$ such that $< \sigma,n> B[v(x)]\, \sigma^*$, $M,\sigma^*,v,n \models \alpha$. That is, $\alpha$ follows from the agent's beliefs iff $\alpha$ is true in all possible worlds accessible via $B$, at index $n$.
(2) $M,\sigma,v,n \models goal(X, \alpha)$ iff for all $\sigma^*$ such that $< \sigma,n> G\, [v(x)]\, \sigma^*$, $M,\sigma^*,v,n \models \alpha$. That is, $\alpha$ follows from the agent's goals iff $\alpha$ is true in all possible worlds accessible via $G$, at index $n$.

We can now see how formulas for quantifying into bel and goal are interpreted.

$M,\sigma,v,n \models \exists x\, bel(a, p(x))$ iff $M,\sigma,v(x,d),n \models \exists x\, bel(a, p(x))$ for some $d$ in $D$, such that for all $\sigma^*$ where $<\sigma,n> B[v(x)]\, \sigma^*$, $M,\sigma^*,v,n \models p(d)$.

In other words, the quantified belief formula is satisfied in some world at some time if there is a value *d* from the domain *D* picked out by the valuation function, such that for all *B*-related worlds, the formula p(*d*) is true of that value. The value *d* has to be the same in all of those worlds. Likewise, the same general satisfaction procedure applies to quantifying into goal formulas – for ∃x goal(a, p(x)) there is some value *d* for which in all *G*-related worlds, p(*d*) is satisfied. Notice that this can continue along chains of *B* and *G* related worlds. Thus, the semantics provides a meaning for ∃x *bel(*a, goal(a, p(x)) -- the valuation function picks out an element of the domain *D*, which is true in all *B*-related worlds, and then in all the *G*-related worlds to those such that p(*d*) is satisfied in those worlds. So, in our terminology, the agent knows what the value for *x* s/he wants p to be true of. Notice also that the agents could be different. See Figure 7.

## A.3 Constraints on the model

Formulas are evaluated in worlds and at times. Belief and goal worlds are related to one another as defined by two relations, B and G, respectively. Seriality means that the relations are (separately) consistent, with there always being a consistent world that is B-related or G-related, to a given world. The B relation is Euclidean, transitive, and serial, while the G relation is serial. Essentially, that means the B-related worlds that the agent thinks are possible form an equivalence relation, though those worlds may not include the real world.

In Cohen and Levesque [1], we adopted the "Realism" constraint such that the G worlds are a subset of the B worlds. In other worlds, the agent chooses among worlds that are among the agent's B-related worlds. Thus, the agent cannot choose what it believes to be impossible.

This possible-worlds diagram shows the current world $W_0$, which is belief-accessible for agent *A* to worlds $W_1$, $W_2$, and $W_3$. In each of these predicate *p* is true of element *d*. Worlds W4-W8 are G (goal)-related for agent *C*. In each of these, *p* is true of element *e*. Thus, agent *A* believes *p(d)*, agent *C* has a goal that *p(e)*, and *∃X bel(A, goal(C, p(X))* is satisfiable.

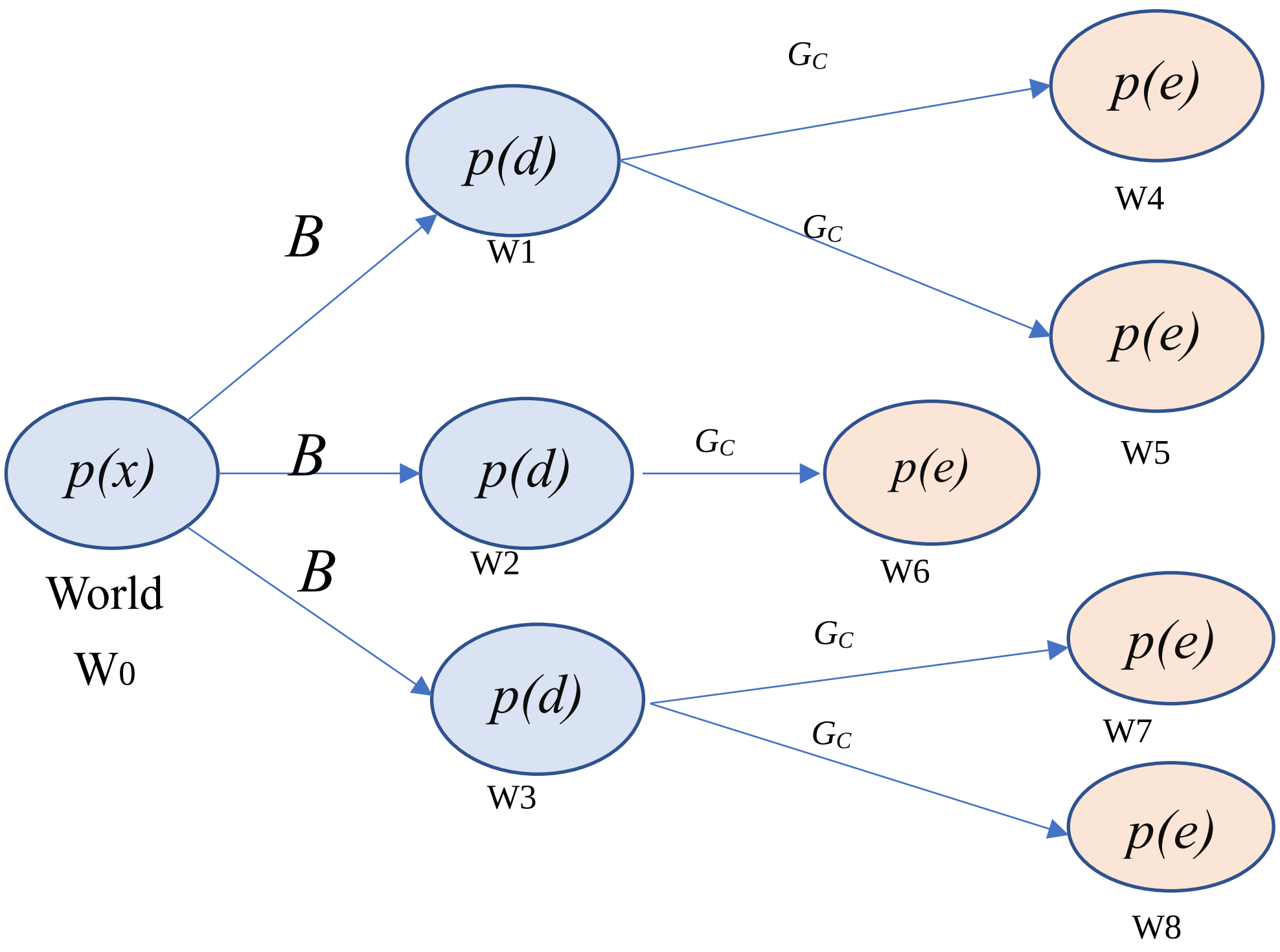

**Figure 7. Possible worlds analysis for ∃X *bel*(A, *goal*(C,p(X)))**

e.g., “A knows the date when A wants to eat”

## Appendix B: Detailed Dialogue Example

We now return to the example[65] given in the introduction, for which we need two additional action definitions to describe the making and rescheduling of an appointment at a business. This time, we include the applicability condition.

***make_appointment****(Agent, Patron, Business, Date, Time)*

*constraint: Cond,*

*precondition: true,*

*effect: have(Patron, appointment(Business, Date, Time))*

*applic_cond: available(appointment(Business, Date, Time))*

In other words, the intended *effect* of an *Agent's* making an appointment for a *Patron* at a *Business* at some *Date* and *Time*, is that the Patron has an appointment. The applicability condition for this action is that an appointment at that date and time must be available. For the moment, we will avoid constraints, and say the precondition is always true.

***reschedule_appointment****(Agent, Patron, Business, OldDate, OldTime, NewDate, NewTime)*

*constraint: Cond,*

*precondition: true,*

*effect: and(neg(have(Patron, appointment(Business, OldDate, OldTime))),*
*have(Patron, appointment(Business, NewDate, NewTime)))*

*applic_cond: and(available(appointment(Business, NewDate, NewTime)),*
*have(Patron, appointment(Business, OldDate, OldTime)))*

The intended *effect* of an *Agent's* rescheduling of an appointment for a *Patron* at a *Business* from some *OldDate* and *OldTime* to *NewDate* and *NewTime*, is that the patron has an appointment at the new date and time and no longer has an appointment at the old date and time. The applicability condition for this action is that the patron already has an appointment at the old date and time and that an appointment at the new date and time must be available.

For the example that follows, we provide the dialogue, an English gloss of the logical representations that the system is manipulating, and a graphical representation of the system's reasoning in the plan display (for clarity and economy of space, we only include the most relevant plan elements at each step). Whereas the reasoning employs the logical formulas shown above, the system provides a "clunky" paraphrase of them in the plan display, enabling us to know how it is reasoning. **Green** indicates acts that have been done, shades of **blue** signify pgoals, shades of **magenta** signify intentions, and **straw** indicates beliefs. The numbers before each plan element show the order in which they were created. The system operates primarily in a breadth-first fashion, but deepening via action decomposition. Black arrows indicate dependence of the node at the arrowhead on the node at the tail. Red dashed lines indicate a blockage and plum-colored edges indicate achievement relations.

[65] See also https://openstream.ai/hubfs/supp_mat_full_dialogue_graph.pdf for the complete plan graph representing Eva's reasoning throughout this example.

<table>
<tr><th colspan="2">Utterance</th></tr>
<tr><th>System reasoning</th><th>Plan Elements</th></tr>
<tr><td colspan="2">U: Are there any covid vaccination centers nearby</td></tr>
<tr><td>While this is interpreted as an existential Y/N question (pgoal #8), the system’s plan recognition component infers that the user asked this question possibly because they want to know the location of a nearby vaccination center. Furthermore, the reason why they might want to know this location is likely because they intend to go there.</td><td>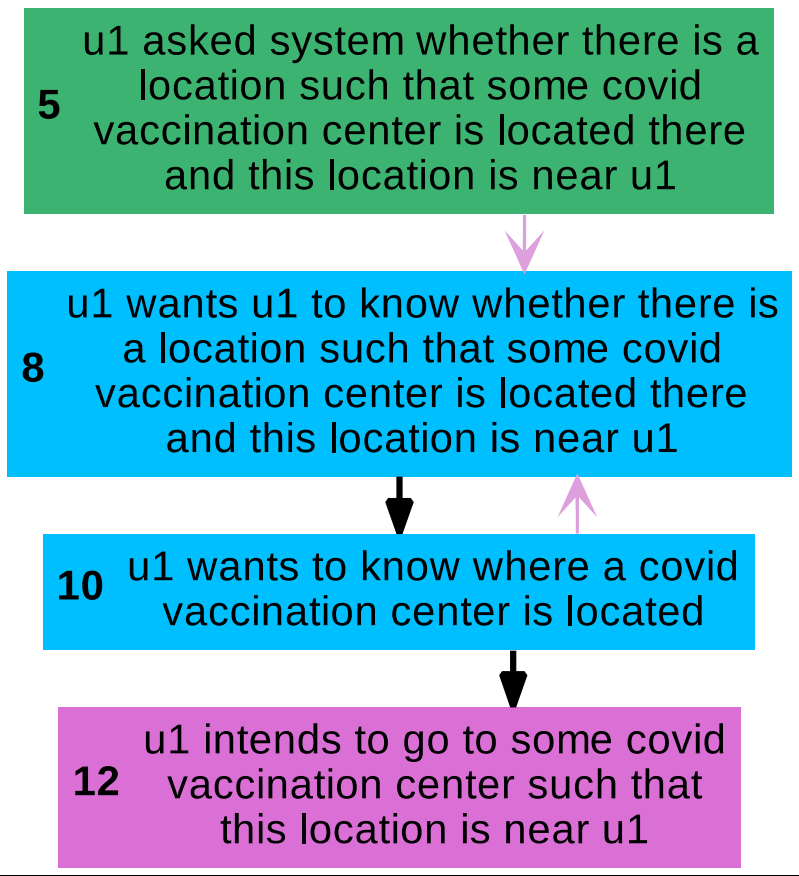
</td></tr>
<tr><td colspan="2">S: Yes, the CVS Covid vaccination center is located at 130 Main St and this location is 6 miles away from you via the following route: turn right, go straight, then turn left.</td></tr>
<tr><td>Given its inference about the user’s likely plan (specifically, pgoal #10), the system over-answers the Y/N question. It tells the user the route because it believes the user needs to know that. (In a multimodal implementation, the route might not need to be verbalized explicitly, but could simply be a hyperlink to a navigation app to display the route).</td><td>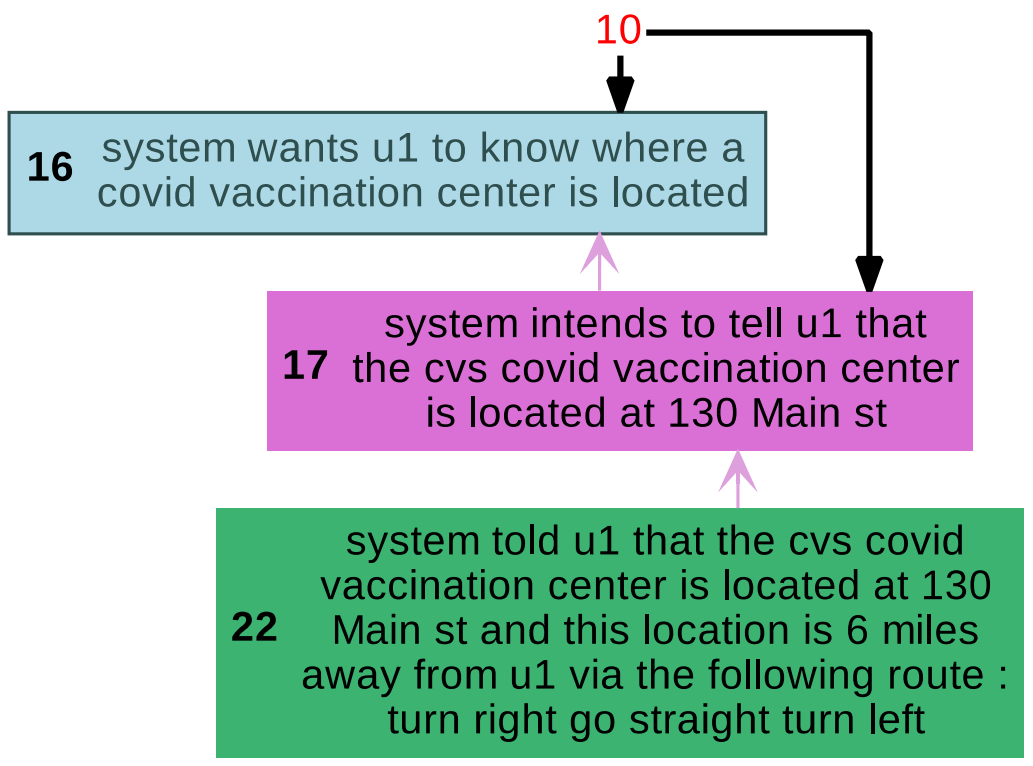
</td></tr>
<tr><td colspan="2">S: Would you like to be vaccinated at a Covid vaccination center?</td></tr>
</table>

<table>
<tr><td>The system's plan recognition component further infers that the reason one might want to go to a Covid vaccination center (intention #12) is to do what one typically does at such a place, namely to get vaccinated. However, the system doesn't have high confidence that its inference about the user's plan is accurate, so it plans to ask the user to confirm (or reject) its hypothesis. Until the system can confirm (or reject) its hypothesis, the user's intention to get vaccinated (represented in node 31) is blocked on the system's goal to know whether the user wants to get vaccinated (the blockage is represented by the red dashed line between nodes 31 and 28)<br><br>Note that the system knows that, in order to get vaccinated, one needs to set up an appointment (based on action preconditions).</td><td>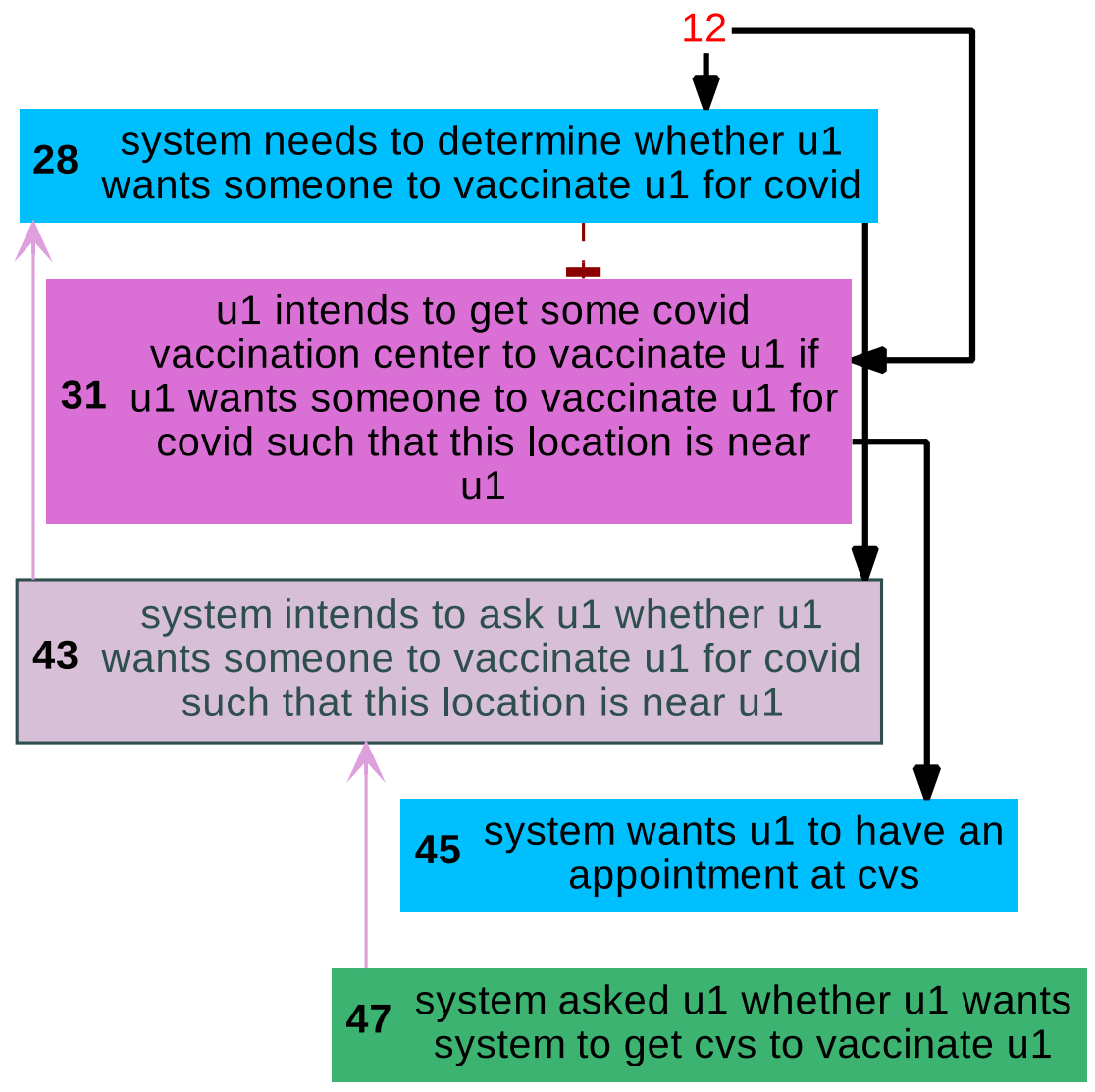
</td></tr>
<tr><td colspan="2">U: Yes</td></tr>
<tr><td></td><td>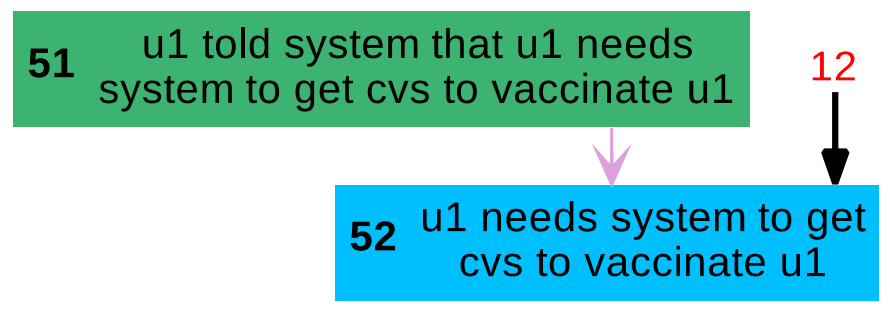
</td></tr>
<tr><td colspan="2">S to CVS: Do you have covid vaccine available?</td></tr>
<tr><td>S does not know if vaccine is available there, but believes that the vaccine center knows whether it is available. So, S plans and executes a Y/N question to CVS about availability (we're assuming here that this information is not obtained via an API, so the system uses a language channel available to any customer, such as calling the center's phone line, or interacting with a chatbot on the center's website).</td><td>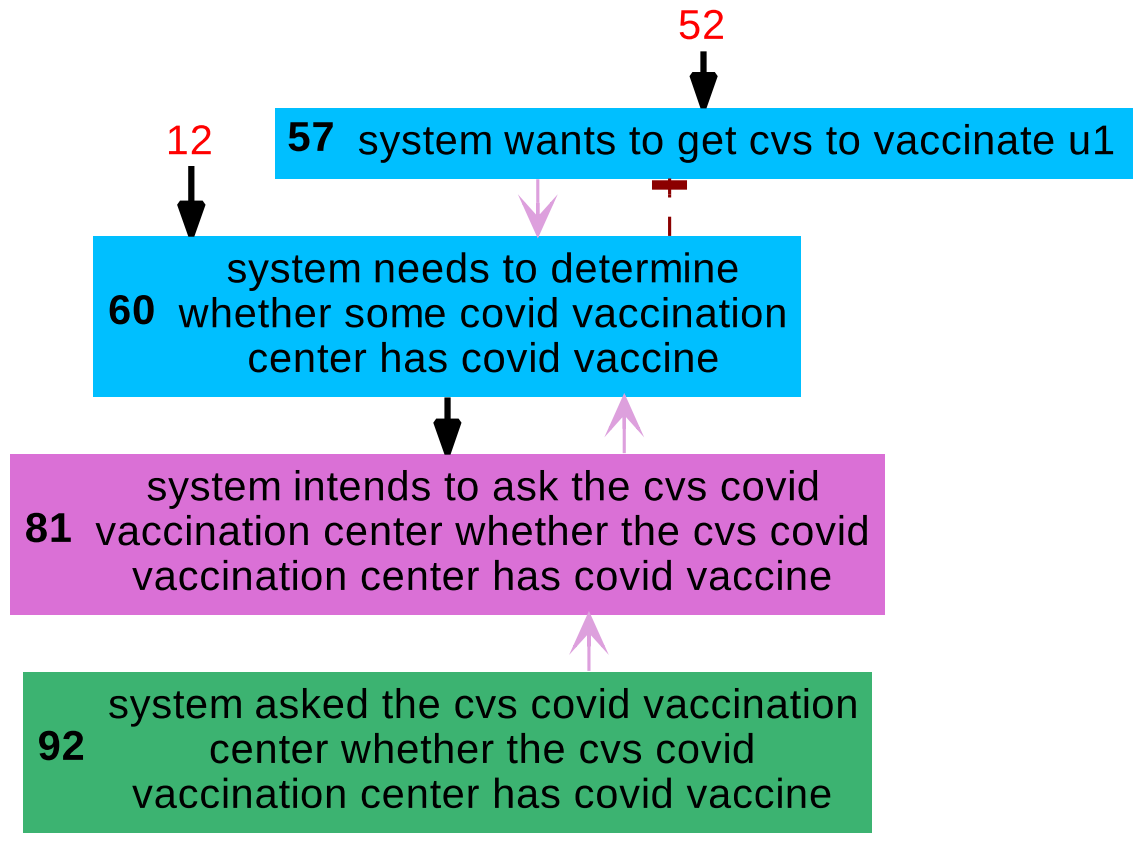
</td></tr>
<tr><td colspan="2">CVS to S: Yes<br>S to U: OK, I believe that the CVS covid vaccination center has Covid vaccine.</td></tr>
</table>

S comes to believe that the CVS vaccine center has vaccine and judges U ought to know this, as well, since this eliminates one potential hurdle in U's plan to get vaccinated.

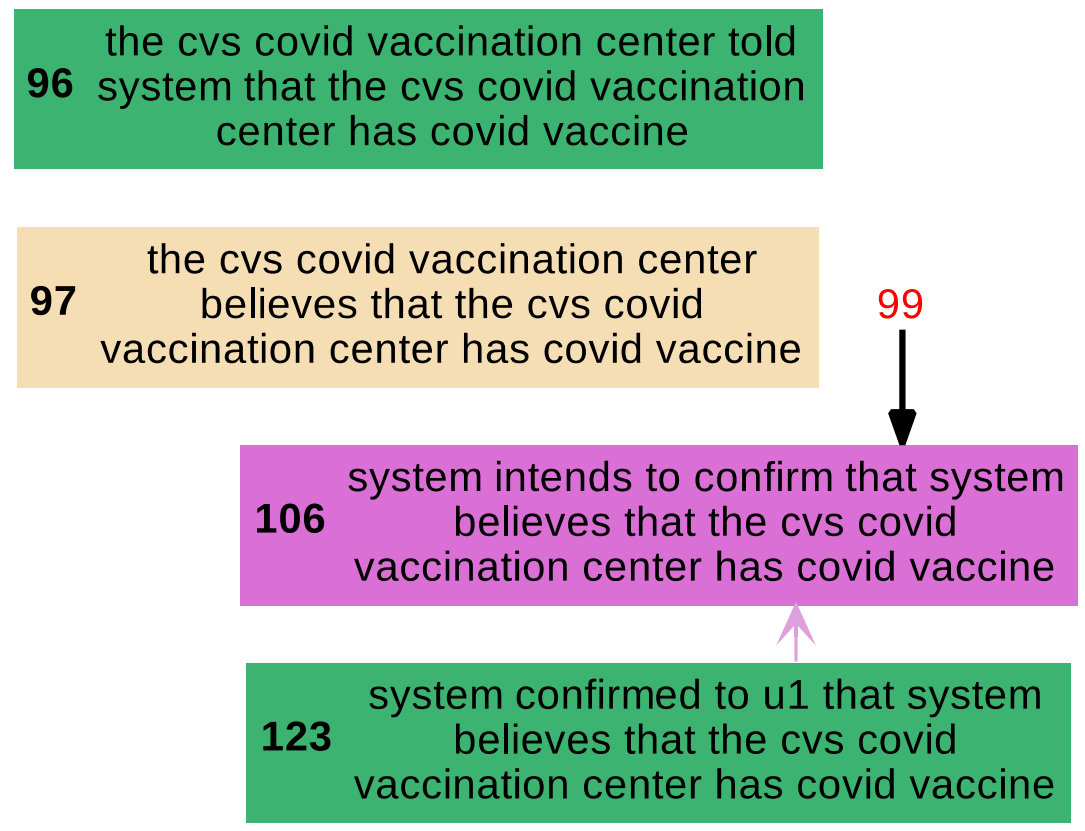

**S:** ***How old are you?***

Another requirement (precondition) for the user to be able to get vaccinated is that the user is eligible for the Covid vaccine. There are several conditions U may satisfy to be deemed eligible, and the system had already made a plan for checking U's eligibility. Some of the constraints require it to know U's age.

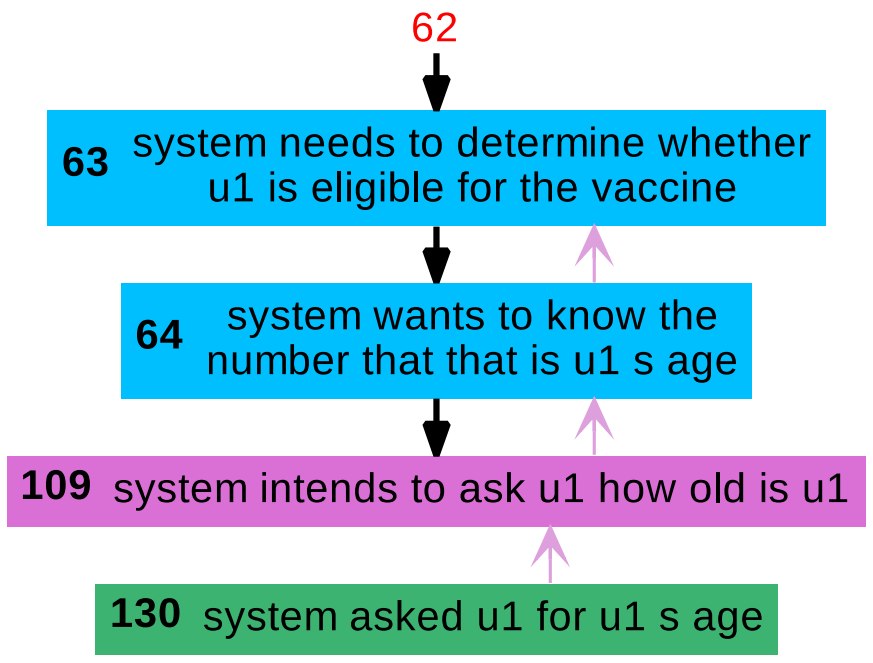

**U:** ***why do you ask?***

**S:** ***The reason is that I need to determine whether you are eligible for the vaccine.***

The previous system question may appear surprising to U, so they ask for an explanation. The system accesses the plan to determine a reasonable pgoal in the chain of reasoning leading to the intention (#109) to ask U about their age. Pgoal #63 is picked as the best explanation.

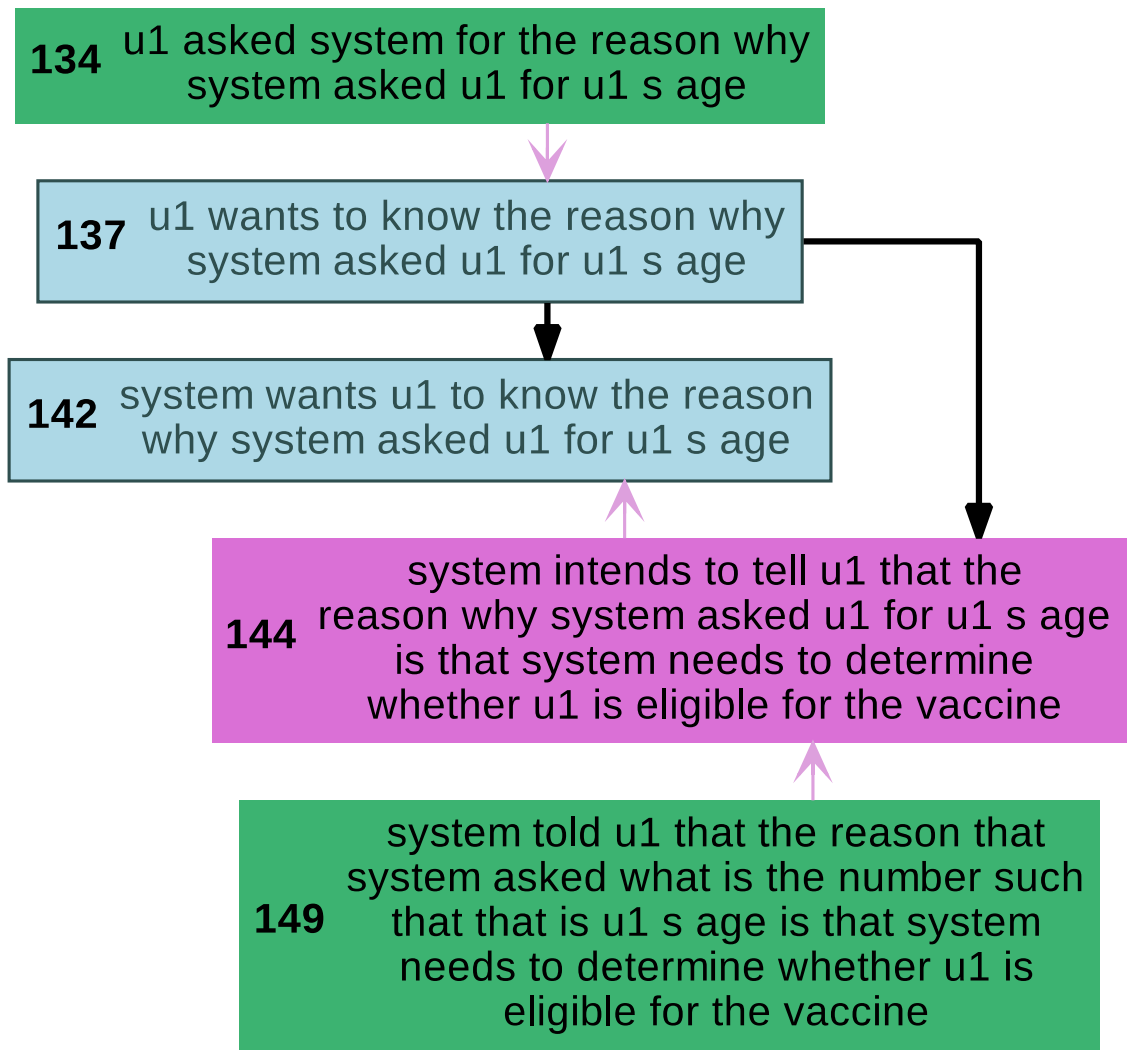

**S:** ***Sorry to repeat myself… How old are you?***

<table>
<tr><td>The system maintains its goal to achieve knowing how old the user is, causing it to repeat its question.</td><td>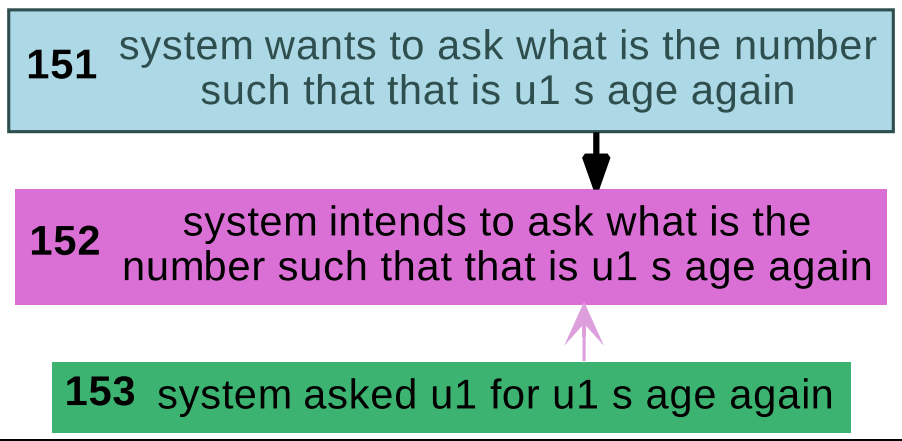
</td></tr>
<tr><td colspan="2">U: 45 years old</td></tr>
<tr><td></td><td>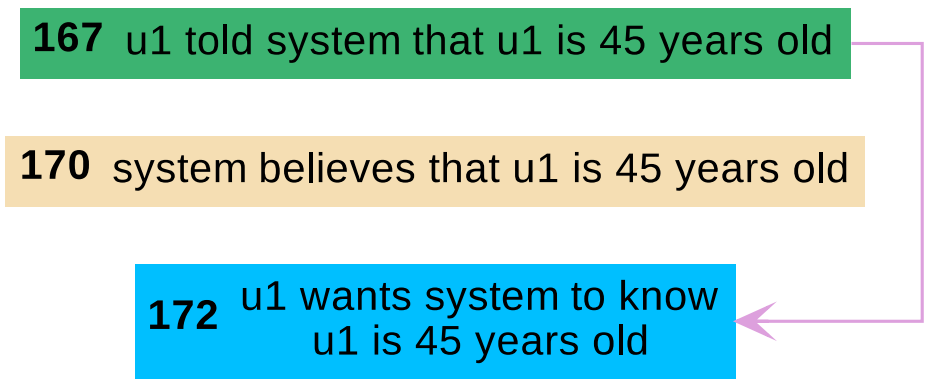
</td></tr>
<tr><td colspan="2">S: What is your occupation?</td></tr>
<tr><td>The specific eligibility rule says that a person is eligible for the Covid vaccine if they are under 50 years old and are an essential worker.</td><td>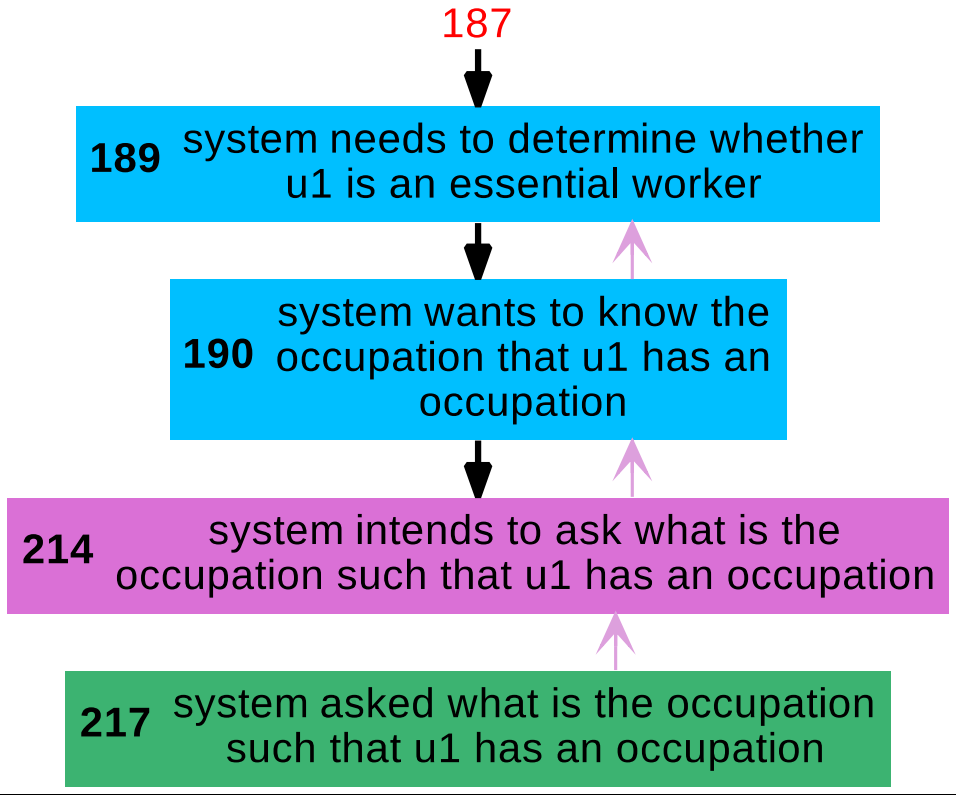
</td></tr>
<tr><td colspan="2">U: why?<br>S: The reason is that I need to determine if you are an essential worker.<br>S: Sorry, to, ask you again, but what is your occupation?</td></tr>
<tr><td>The rationale for each system action has to be sufficiently specific, but not trivial. In this case pgoal #189 is found to provide a good explanation. As above, system repeats the question, trying again to achieve the goal of knowing U's occupation.<br>The chain of pgoals and intentions is similar to the previous case. Here we highlight the beliefs the system forms about U's own beliefs as a result of the speech acts performed by U and the system itself.</td><td>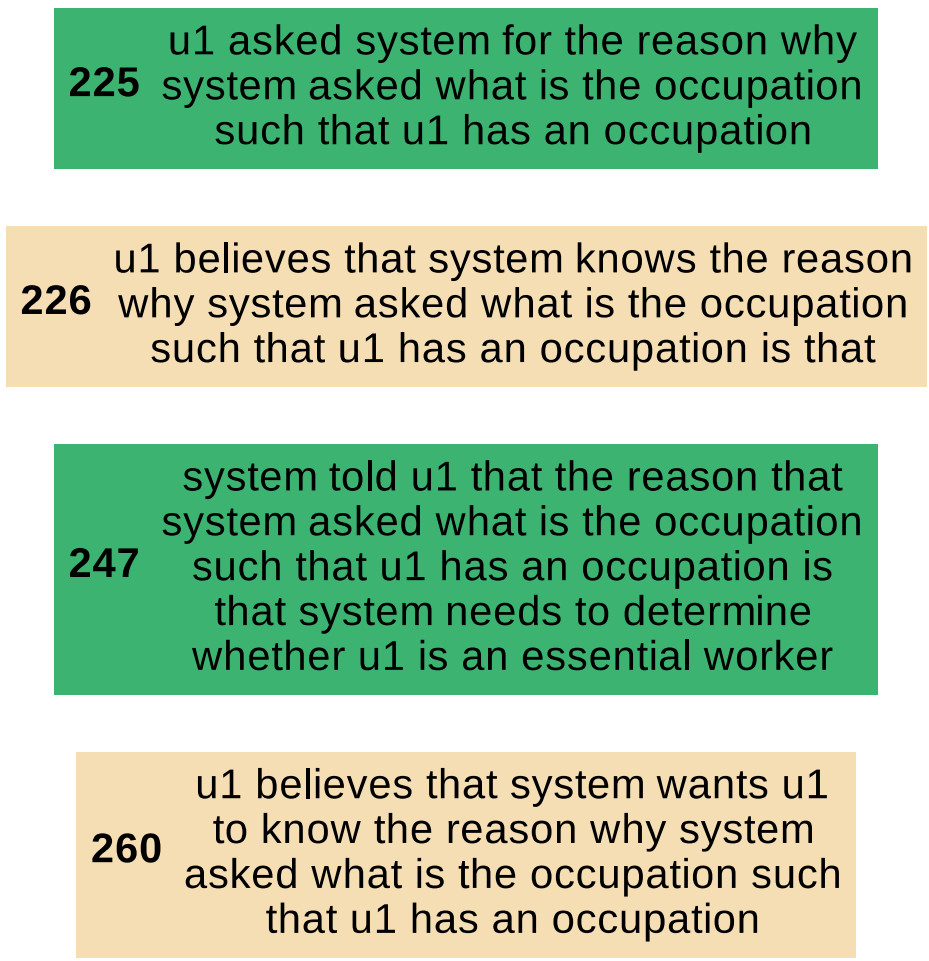
</td></tr>
<tr><td colspan="2">U: Teacher<br>S: You are eligible for the covid vaccine and appointments are available at CVS on June 26 at 9am.</td></tr>
</table>

<table>
<tr><td>The system's goal of knowing whether U is eligible and CVS having appointments is now satisfied. For this eventuality S had already planned to confirm this fact to U (similarly, it had planned to tell the user if this turned out not to be true); now it can perform this confirmation.</td><td>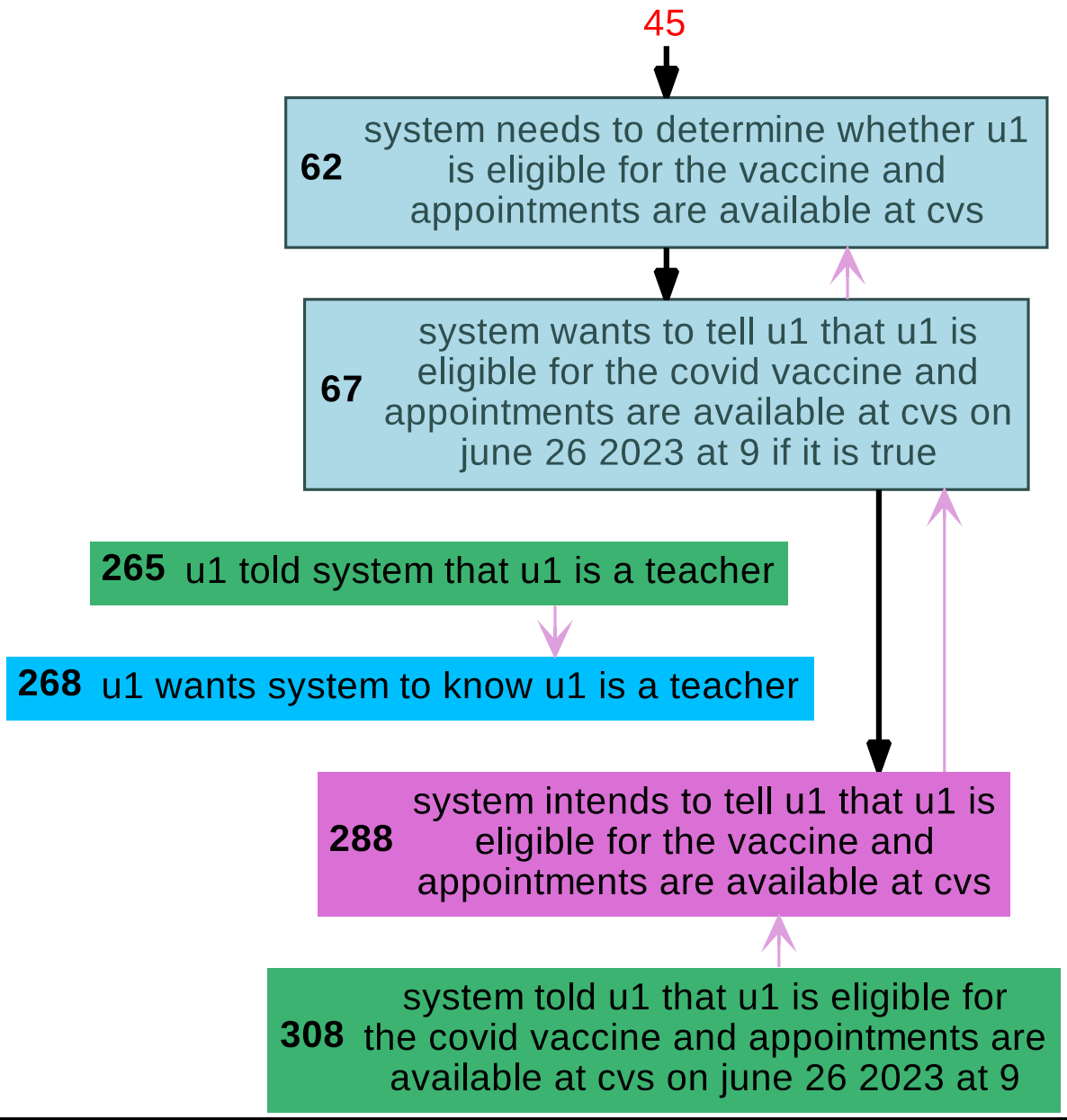
</td></tr>
<tr><td colspan="2">S: Would you like me to make an appointment for you at CVS?</td></tr>
<tr><td>In order to be vaccinated, the user needs an appointment, which is accomplished by making an appointment provided the user is eligible. Since the user is eligible, the conditional intention becomes unconditional and the system offers to make an appointment for the user.</td><td>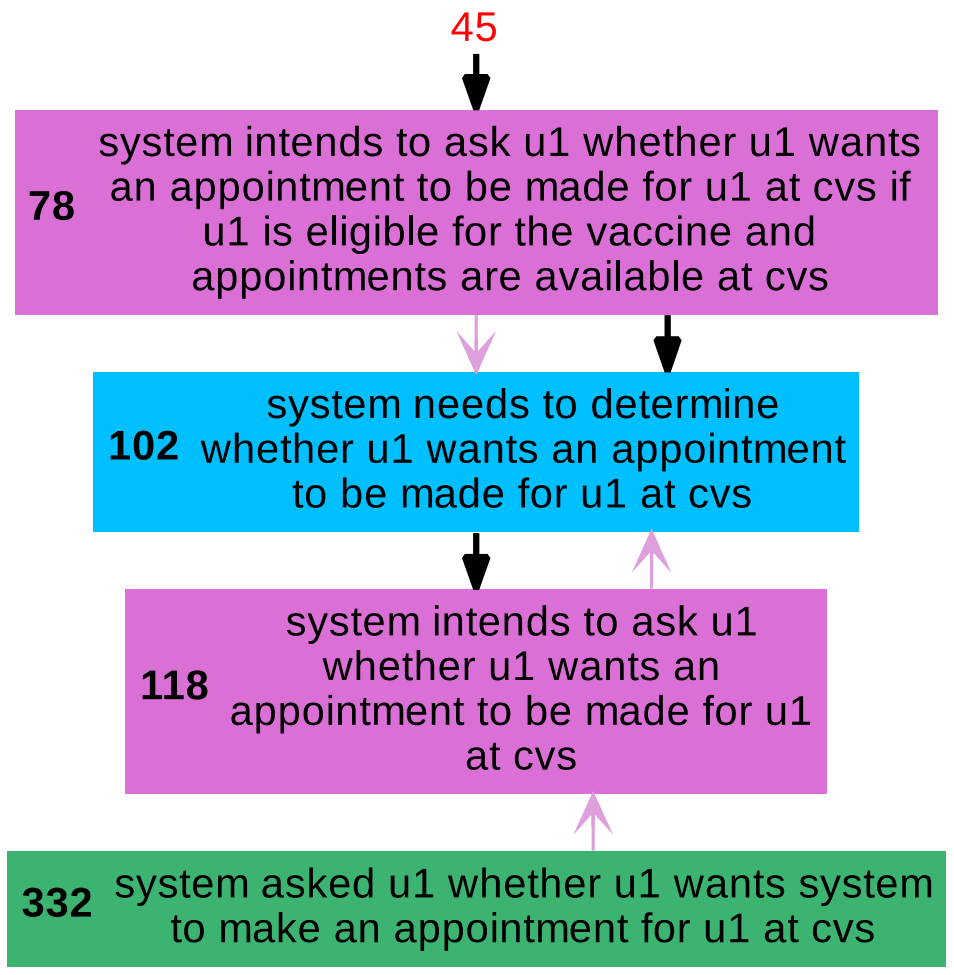
</td></tr>
<tr><td colspan="2">U: Yes, please<br>S: What time would you like me to make an appointment at a covid vaccination center for you?</td></tr>
</table>

Now that U confirmed that they want the system to make an appointment for them at CVS (relevant portions of the plan graph omitted to save space), the system is starting to pursue this goal. The (sub-)plan for setting up an appointment requires that it knows the date and the time of the appointment. This "slot-filling" is performed by reasoning about knowledge. Note that the system has generated multiple goals at the same time for the various values that it needs. This is useful for the ability to handle the situation when the user over-answers a question.

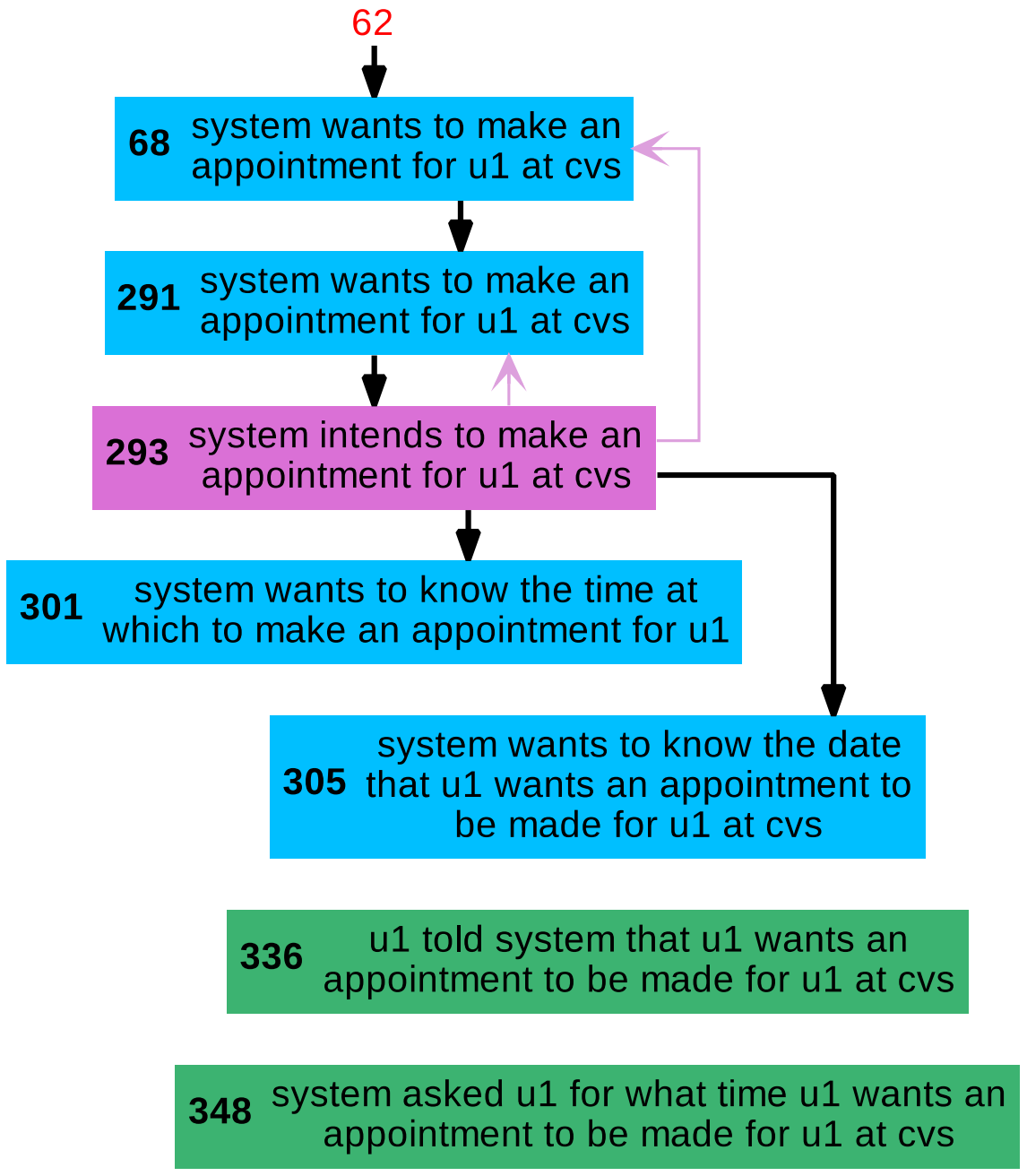

**U:** ***Monday the earliest time available***

The user over-answers, providing a date. However, for the time of the appointment, the user provides a constraint, which does NOT fill the slot. Hence the system keeps its *pgoal* to know the time, but updated with the new information about the date (#369).

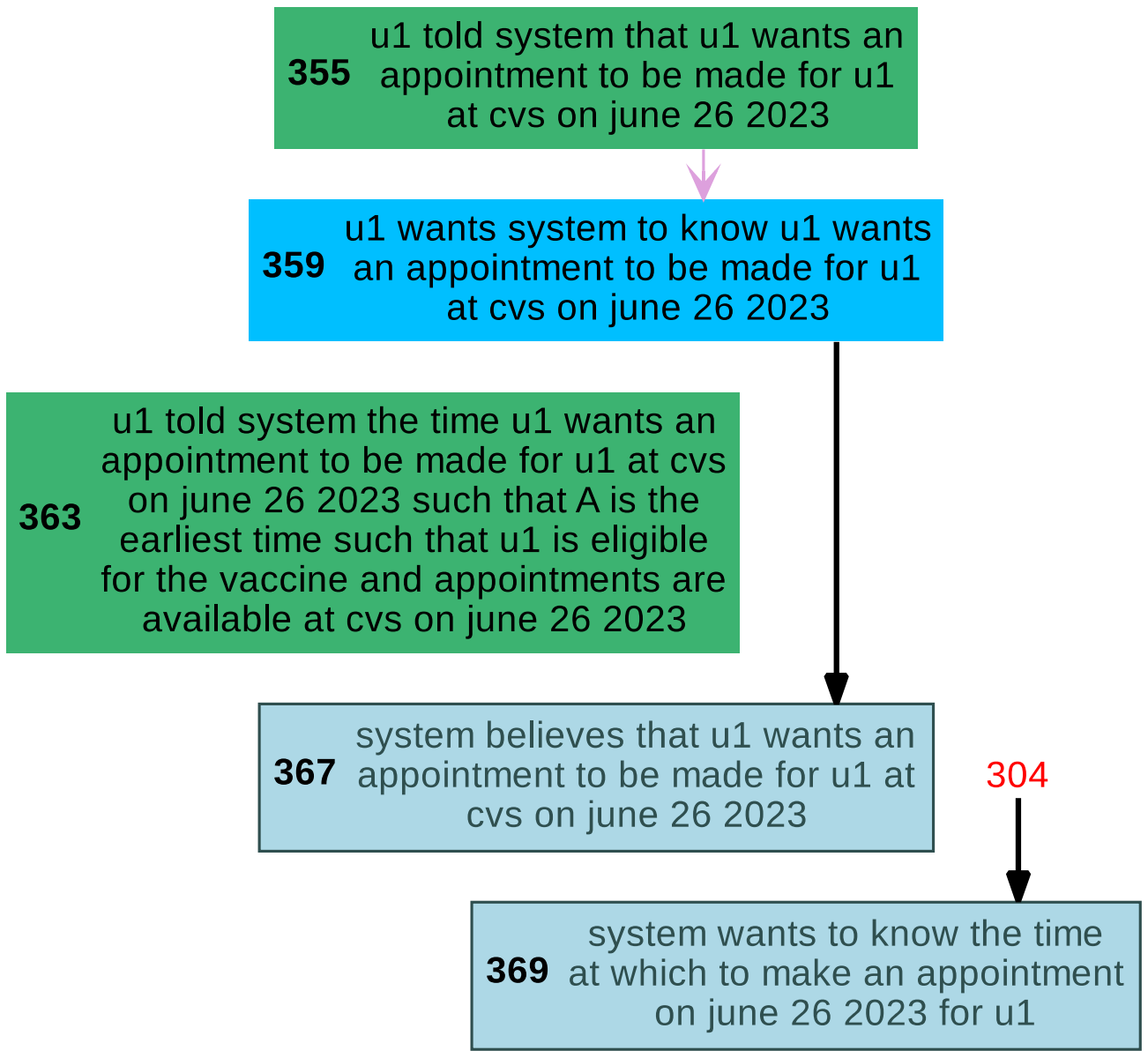

**S:** ***The earliest time available is 8 am.***

**S:** ***Would you like me to make an appointment for you at CVS on June 26 2023 at 8am?***

Because the user didn't provide a specific time, the system plans to propose one that matches the user's constraints. It finds the earliest appointment available on the date requested (in this case 8am), and plans a sequence of two speech acts: (i) to tell U that that satisfies the given constraint, and then (ii) to request that U confirm that that is, indeed, the time they want the appointment to be made for.

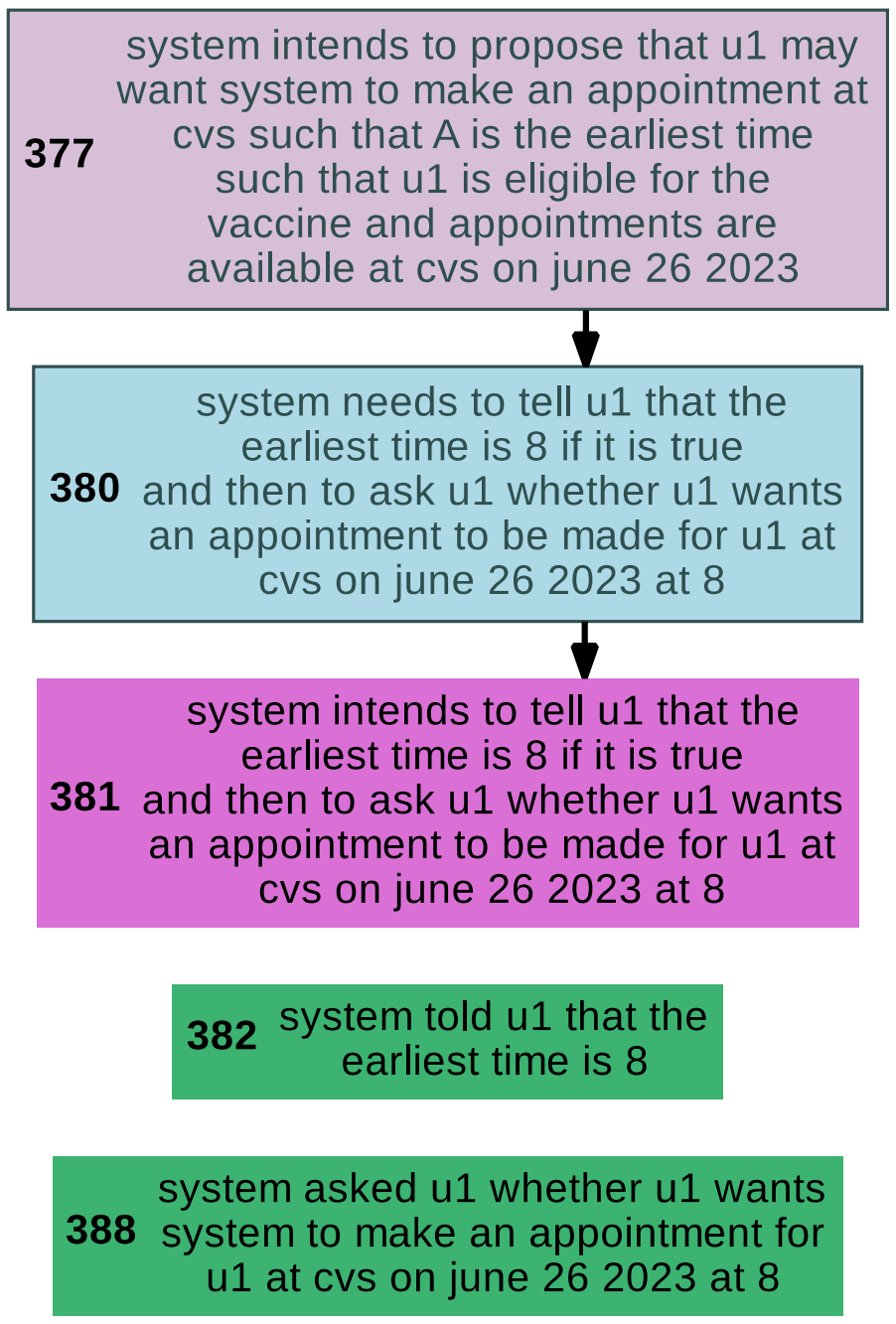

**U:** *sure*

**S:** *Ok I made an appointment for you at CVS on June 26 2023 at 8am.*

User accepts; hence system adopts the U's goal to make the appointment on Monday at 8am. It creates the intention to do so, and then to tell the user that the appointment was made.

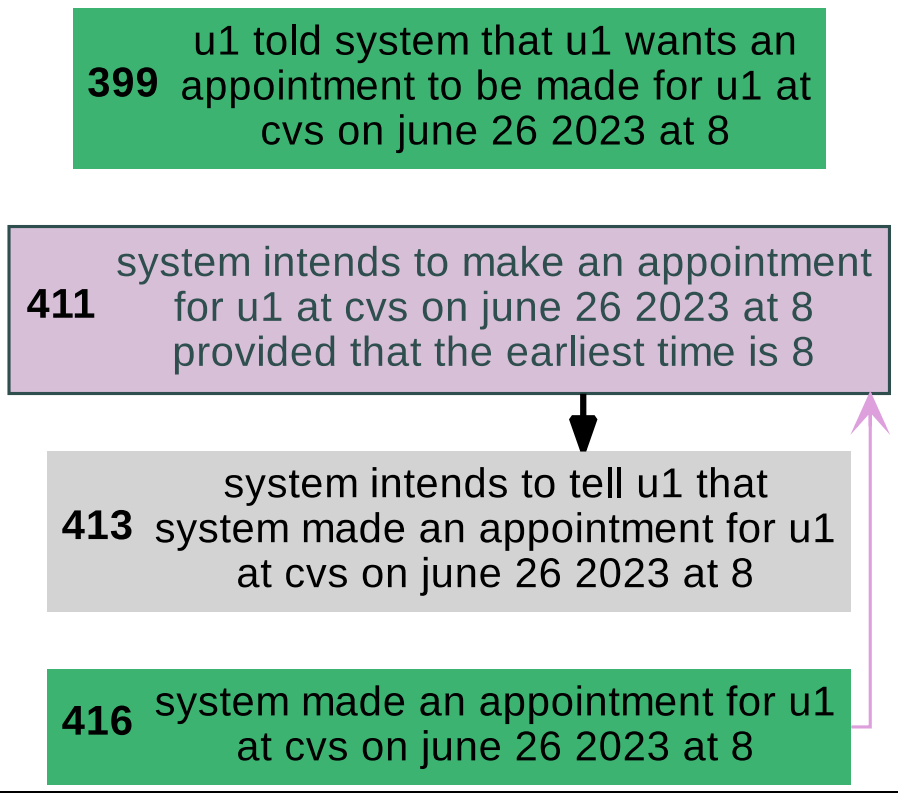

**S:** *Is there anything else I can help you with?*

At this point the system has no more actions ready to execute, so it checks if the user has any additional goals.

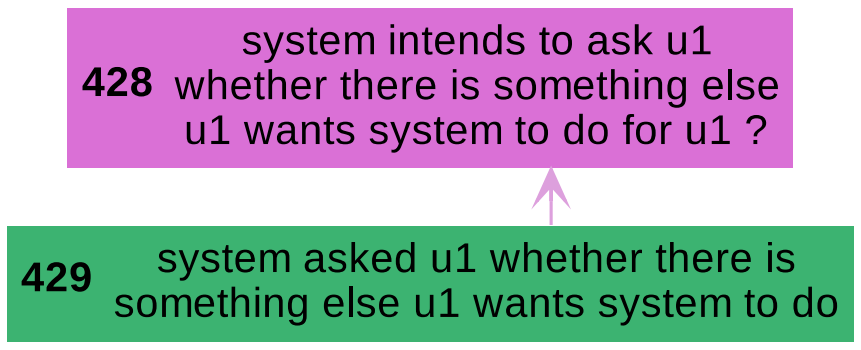

**U:** *yes, i actually cannot make it on monday at 8*

**S:** *Would you like me to reschedule your appointment at CVS that is currently scheduled on June 26 2023 at 8?*

System interprets U's utterance to mean they don't want the previously made appointment anymore. This triggers an intention revision process: if U still wants to get vaccinated, they would need another appointment, provided any are available. Therefore, system asks U if they would like to reschedule.

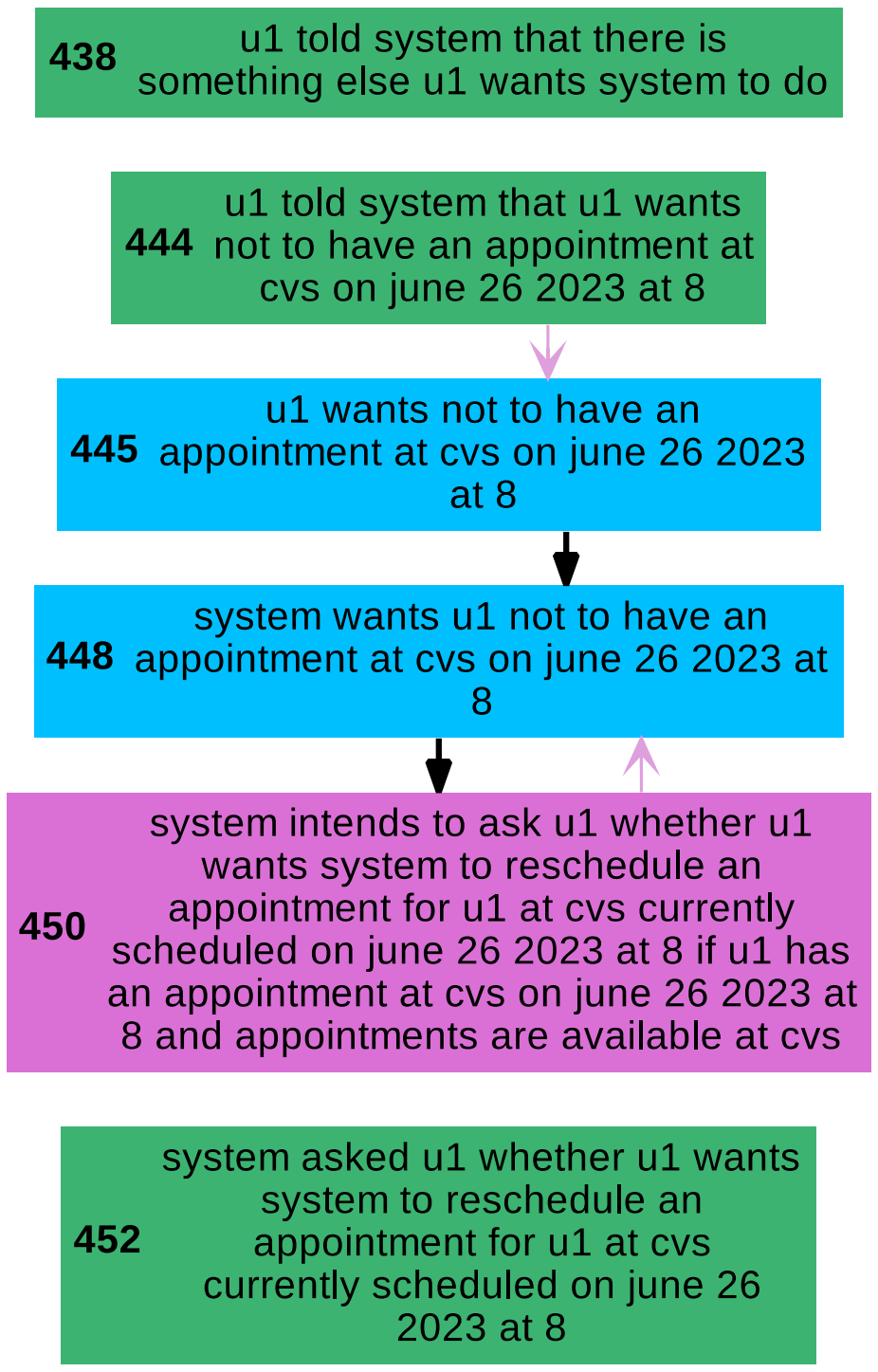

**U:** *yes*

**S:** ***What date would you like me to reschedule your appointment at CVS to?***

Since U expressed the desire to have the appointment rescheduled, system plans to obtain from U the date and time for the new appointment.

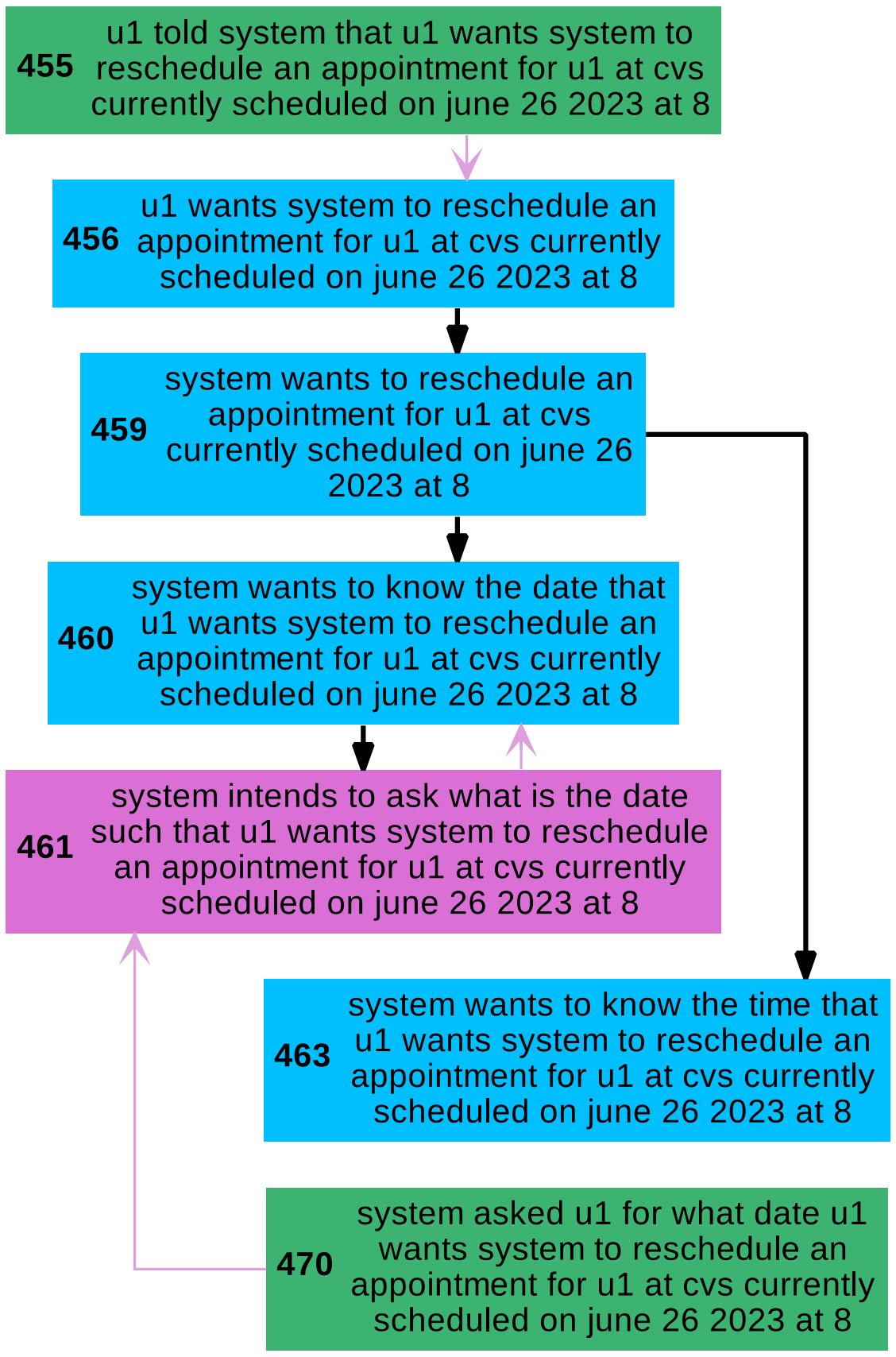

**U:** ***monday at 10***

User again over-answers, but this time provides values for both the date and the time “slots”. The system adopts the user’s goal to reschedule the appointment to the new date and time.

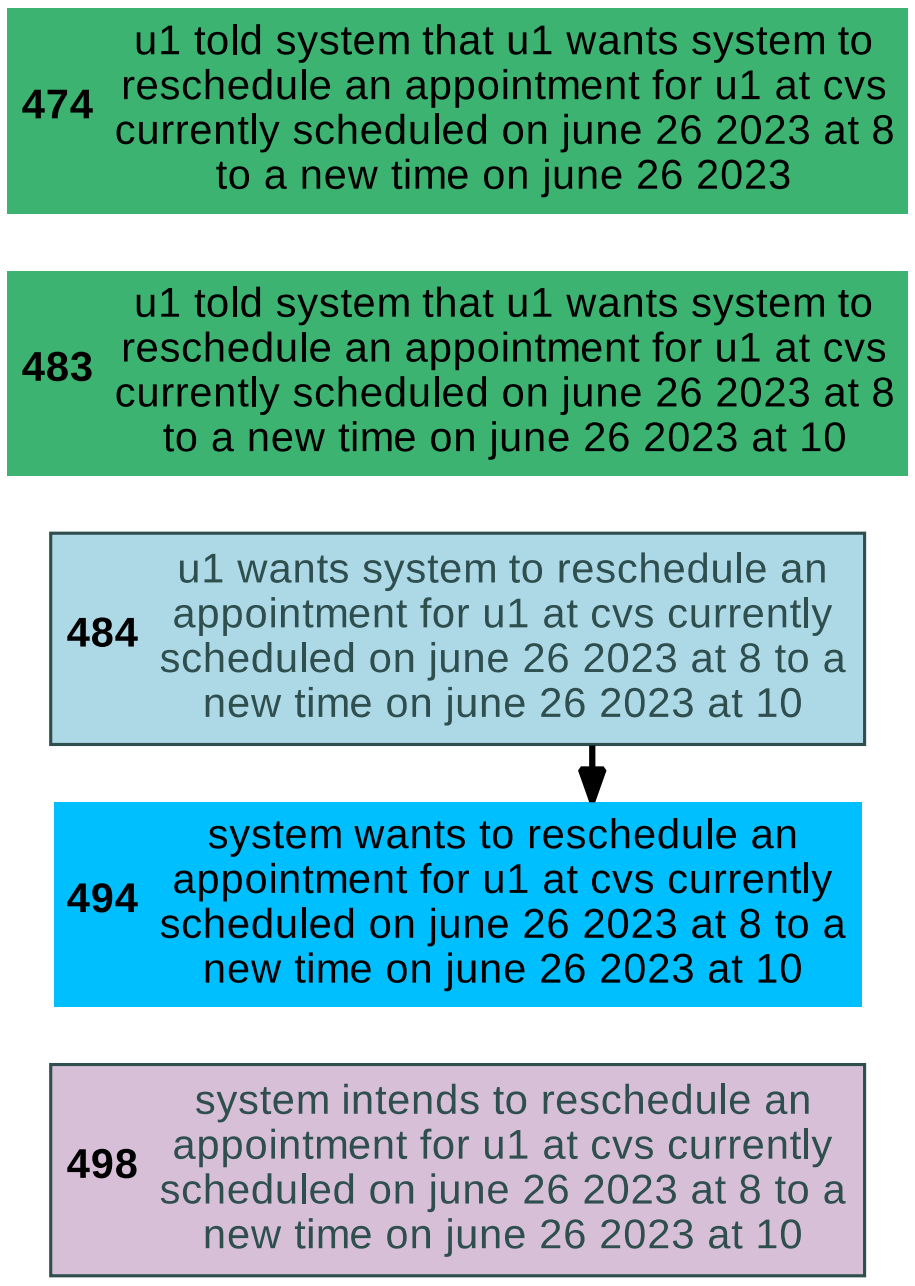

**S:** ***Ok, I’ve rescheduled your appointment at CVS to a new time on June 26 2023 at 10am.***

The action of rescheduling an appointment is hierarchically decomposed into a sequence of two (sub-) actions: first, canceling the old appointment, and second making the new appointment. The system executes each of the sub-actions and informs the user that the appointment was rescheduled, and the system updates its beliefs about U’s beliefs.

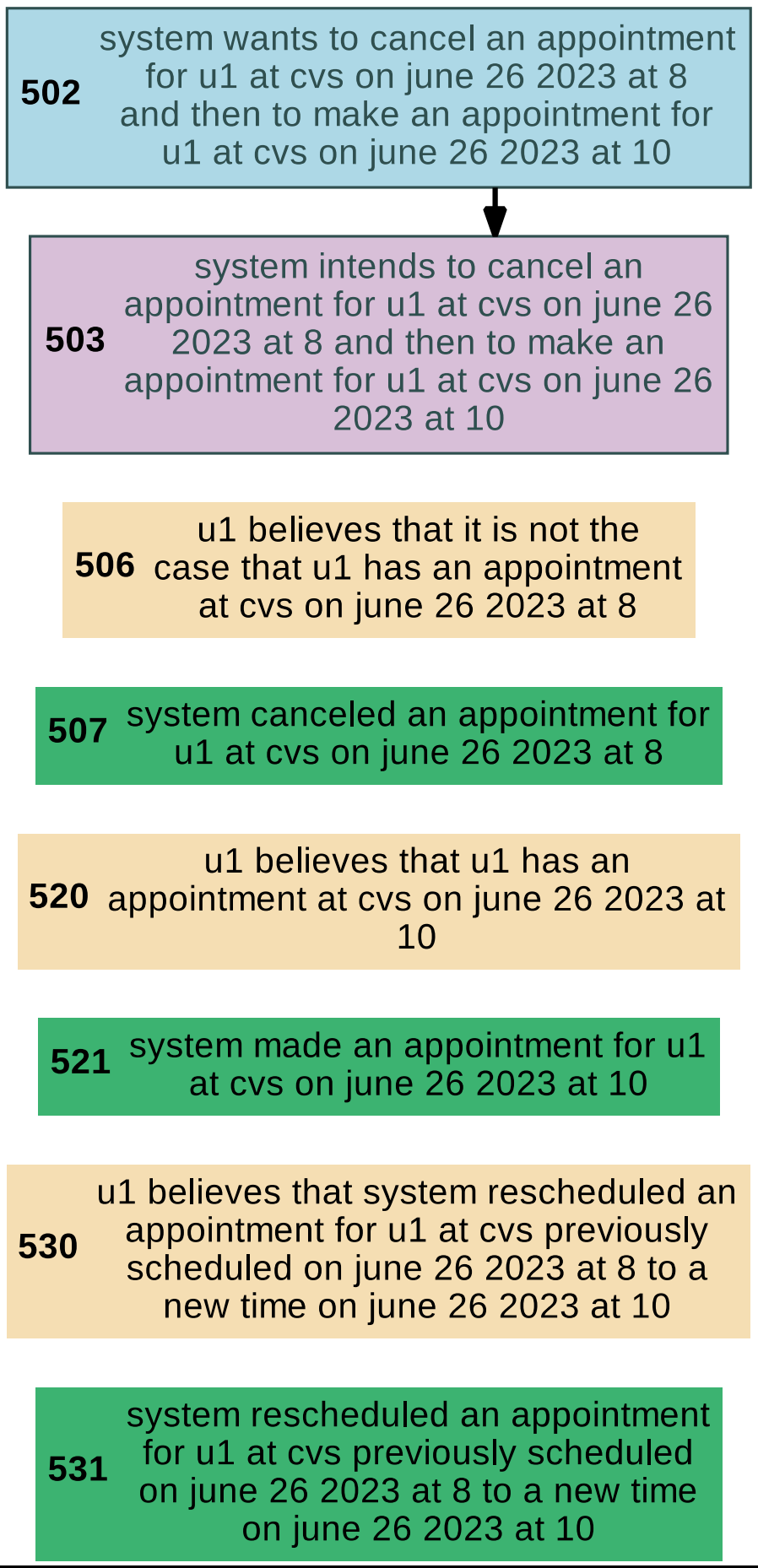

| | |
|---|---|
| **S:** ***Can I help you with anything else?***<br>**U:** ***No, thanks!***<br>**S:** ***Glad to be of service. Goodbye!*** | |
| Having again exhausted all the actions ready to be executed, system asks U if they have additional goals. Since the user responds in the negative, the system closes the conversation. | |